\documentclass[12pt]{article}

\usepackage{newtxtext,newtxmath}

\usepackage{graphicx}

\usepackage[letterpaper,margin=1in]{geometry}

\renewenvironment{abstract}
	{\quotation}
	{\endquotation}

\date{}

\makeatletter
\renewcommand{\fnum@figure}{\textbf{Figure \thefigure}}
\renewcommand{\fnum@table}{\textbf{Table \thetable}}
\makeatother

\usepackage{scicite}

\usepackage{url}

\usepackage{array,booktabs,tabularx,xltabular,makecell,multirow}
\makeatletter\let\sciMakeCaption\@makecaption\renewcommand{\@makecaption}[2]{{\linespread{1.0}\selectfont\sciMakeCaption{#1}{#2}}}\makeatother
\def\scititle{
	Building verifiable databases at scale from the chemical literature via a multi-agent system
}
\title{\bfseries \boldmath \scititle}

\author{
	Yufan~Chen$^{1}$,
	Ching~Ting~Leung$^{1}$,
	Bowen~Yu$^{1}$,
	Haifan~Zhou$^{1}$,\and
	Jianwei~Sun$^{1,2}$,
	Yong~Huang$^{2}$,
	Linyan~Li$^{3}$,
	Hao~Chen$^{1,4}$,
	Hanyu~Gao$^{1\ast}$\and
	\small$^{1}$Department of Chemical and Biological Engineering,\\ \small The Hong Kong University of Science and Technology; Hong Kong SAR, China.\and
	\small$^{2}$Department of Chemistry,\\ \small The Hong Kong University of Science and Technology; Hong Kong SAR, China.\and
	\small$^{3}$Department of Data Science, City University of Hong Kong; Hong Kong SAR, China.\and
	\small$^{4}$Department of Computer Science and Engineering,\\ \small The Hong Kong University of Science and Technology; Hong Kong SAR, China.\and
	\small$^\ast$Corresponding author. Email: hanyugao@ust.hk
}

\begin{document}

\maketitle

\begin{abstract} \bfseries \boldmath
Accelerating chemical research with artificial intelligence requires high-quality databases. However, many reactions exist only as graphics drawn for human readers, which machines must read exactly. Specialist software is precise but rigid, whereas multimodal language models are flexible yet often write incorrect structures; neither flags errors. We present ChemEAGLE, a multi-agent system where tools supply structures and a language model extends their reach through judgments that programs apply. On 424 published graphics, its F1 score for exactly correct reactions was 82.2\%, against 47.7\% for a tool-equipped generalist agent, and calibrated confidence limited review to 35\% of reactions at 96.6\% accuracy. Without review, it built a 60,103-reaction photochemistry database whose records trained models to rank conditions and predict products. Verifiable databases can thus be built at scale.
\end{abstract}

\noindent
Data remain a central bottleneck for artificial intelligence in science~\cite{wang2023scientific}. In chemistry, models for synthesis planning and for predicting reaction outcomes and conditions learn from experimental reactions~\cite{segler2018planning}, yet most reactions are published as reaction schemes, substrate tables and annotated figures designed for human readers, which we call reaction graphics. Such reactions reach databases mainly through expert curation~\cite{guo2022chemrxnextractor}, which is slow and cannot keep pace with the literature, so data for such models must increasingly be extracted automatically. Reading a reaction graphic requires both flexible interpretation, because a reaction series may be spread across a template drawing, a table of products or substituents, footnotes and text linked by compound labels, and exact output, because a single misplaced atom yields a different compound.

Existing approaches secure one requirement at the expense of the other. Specialist models recognize molecular images~\cite{rajan2020decimer,qian2023molscribe}, parse reaction diagrams~\cite{wilary2023rde2,qian2023rxnscribe} and extract reactions from text~\cite{vangala2024patent,dagdelen2024structured}, and rule-based pipelines such as OpenChemIE combine them~\cite{fan2024openchemie}; they are exact within their conventions but fail on layouts they do not anticipate. Multimodal large language models (MLLMs)~\cite{openai2023gpt4,openai2024gpt4o} read unfamiliar layouts but write structures token by token, often incorrectly~\cite{alampara2025probing}. Tool-augmented models~\cite{shen2023hugginggpt,yang2023mmreact,yao2023react,gupta2023visprog}, including chemistry agents~\cite{bran2024chemcrow,boiko2023coscientist} and extraction pipelines~\cite{leong2025mermaid,leong2024electrosynthesis}, give one model specialist tools, but that model still plans every step and restates the tool outputs itself. None of these approaches estimates whether an individual reaction is correct, and the confidence that language models report is often poorly calibrated~\cite{xiong2024uncertainty}, so every record must be checked by a chemist.

Here, we present ChemEAGLE, a multi-agent system~\cite{hong2024metagpt,wu2023autogen} that divides the reading of a graphic at two levels and checks the result. Between agents, a Planner assigns each part of a graphic to a specialized agent. Within each agent, specialist tools supply the structures and an MLLM extends their reach with its judgment, referring to the recognized molecules through numbered marks drawn on the graphic, a visual prompt related to set-of-mark prompting~\cite{yang2023setofmark}; programs apply these decisions, so that the model does not retype structures. Observers and chemical checks review the result, and a Confidence Scorer estimates the probability that each reaction is correct. We test it on graphics from articles, a review and patents and extend it to whole articles (Fig. 1A). Finally, we use it to build a photochemistry database without review and ask whether its records can train models for new reactions.

\subsection*{Results}
\subsubsection*{\textit{Design of ChemEAGLE}}

ChemEAGLE divides the reading of a reaction graphic at two levels (Fig. 1, B and C; fig.~S1; table S1). Between agents, the Planner routes each part of the graphic to the agent whose tools and checks suit it; a substrate table, for example, goes to one of two R-group substitution agents depending on whether its cells hold complete molecules or substituents (fig.~S2). Within each agent, a program outlines every molecule found by the tools (table S2) and tags it with a number, the visual prompt (fig.~S3). The model refers to molecules by these numbers, for example to name the product of a table entry, an atom symbol that a tool misread or the drawn molecules that a text definition such as X = Cl, Br, I expands, even when definitions are nested; programs check these decisions against the tool outputs and apply them (figs. S4 and S5). A Plan Observer and an Action Observer review the plan and each agent's output; switching both off lowered the hard-match F1 score (defined below) from 82.2\% to 77.1\% (\emph{P} = 0.0022, two-sided Welch's t-test; fig.~S6; supplementary text and table S3). Deterministic checks then repair the assembled reactions, names that are not drawn are converted to structures (table S4), the Confidence Scorer estimates the probability that each reaction is correct, and for whole articles a Main Text Agent reads the text together with each graphic's reactions, completing them and adding those that only the text reports (Fig. 1A; materials and methods).

\subsubsection*{\textit{Versatile extraction across sources and designs}}

We assembled a benchmark of 424 reaction graphics with 3,613 reactions that spans the sources and designs of the literature: 64 recent research articles, 48 articles of an earlier benchmark~\cite{fan2024openchemie}, a review of N-heterocyclic carbene catalysis~\cite{flanigan2015nhc} and 14 patents (fig.~S7). The graphics combine templates with structure-based or text-based tables, footnotes and prose, and the patents add Markush structures and groups defined in text. Four chemists annotated every graphic independently, with pairwise F1 scores of 96.5 to 98.8\% (tables S5 and S6). Under the hard-match criterion used throughout, a reaction counted as correct only if every reactant and product structure, and every reagent and catalyst structure drawn in the graphic, matched the ground truth (fig.~S8); we report the F1 score, the harmonic mean of precision and recall.

We compared ChemEAGLE with the rule-based pipeline OpenChemIE~\cite{fan2024openchemie}; MERMaid, a single MLLM with extraction tools~\cite{leong2025mermaid}; seven MLLMs used end to end; and two generalist coding agents given ChemEAGLE's complete toolset, Claude Code on Claude Sonnet 5 and Codex on GPT-5.6-sol. ChemEAGLE reached a hard-match F1 score of 82.2 $\pm$ 0.9\% (mean $\pm$ standard deviation (SD) of three runs; Fig. 2A; table S7), against 47.7 $\pm$ 1.7\% for Claude Code, the strongest alternative (\emph{P} = 6.0 $\times$ 10$^-$$^5$, two-sided Welch's t-test), and at most 27.8\% for the MLLMs. Across the four sources, its F1 ranged from 75.5\% to 93.2\%, whereas that of OpenChemIE ranged from 1.8\% to 63.7\%.

The alternatives failed in opposite ways (Fig. 2B). With output and ground-truth reactions paired by product similarity, OpenChemIE wrote exact structures when graphics followed its conventions but otherwise missed most reactions, whereas the MLLMs found up to 89\% of the reactions but wrote at most 31\% of these exactly. ChemEAGLE found 97\% and wrote 85\% exactly. The generalist agents found 89\% and 86\% but wrote only 46\% and 36\% exactly, whereas ChemEAGLE reached 80.2\% and 77.8\% F1 with the same base models (table S8); 71 to 79\% of their incorrect products and reactants had been written or edited by the agent rather than copied from a tool output (table S9). On one graphic, only ChemEAGLE returned all seven reactions correctly (Fig. 2C; figs. S9 to S11), and on three graphics of other designs it again extracted the most (Fig. 2D).

\subsubsection*{\textit{Document-level extraction from raw PDFs}}

A single graphic often lacks information that the article gives elsewhere, such as a substrate drawn in an earlier scheme, a product named only in the experimental section or conditions stated in the text. On 19 articles~\cite{fan2024openchemie} whose 2,149 reactions are indexed in Reaxys, a manually curated database, ChemEAGLE reached an F1 score of 72.2\%, against 13.8\% for OpenChemIE and 10.3\% for MERMaid (Fig. 2E). The graphics alone recovered 55.5\% of the indexed reactions; the Label Registry and the three steps of the Main Text Agent raised recall to 70.9\% at a precision of 71 to 74\% (Fig. 2F). Reactions stated only in the text, 12\% of those indexed, went from 0\% to 73.6\% recall (Fig. 2G). Names found by the Main Text Agent also corrected structures: in one article~\cite{zhang2023phthalides}, the Label Registry preferred the named 5-substituted substrates to the 4-substituted ones assembled from the graphic (fig.~S12B).

\subsubsection*{\textit{Exact structures through reference to tool outputs}}

If tools rather than the model write the structures, the advantage should hold whichever model supplies the judgments. Used end to end, the seven base models reached hard-match F1 scores of 0.9 to 27.8\%; inside ChemEAGLE, they reached 74.2 to 82.2\%, each above the strongest alternative, at up to US\$0.60 per graphic (Fig. 3A; table S8). With three base models, 97 to 98\% of the output reactants and products were copied from a tool output, and 91 to 98\% of these were correct, against 70 to 82\% of the molecules the model wrote itself (Fig. 3B). Replacing the in-context example or rewording the prompts changed hard-match F1 by at most 1.4 percentage points (fig.~S13).

The division within agents requires the model to tell which tool output belongs to which drawn molecule. Without the visual prompt, with identifiers and box coordinates given as text, hard-match F1 fell from 82.2 $\pm$ 0.9\% to 79.0 $\pm$ 0.4\% with Gemini 3.7 Flash (\emph{P} = 0.017) and by 6.9 and 8.2 percentage points with GPT-5.6-luna and Qwen3.8-27B (Fig. 3C). Errors that bind a judgment to the wrong tool output became more frequent with every base model: with Gemini 3.7 Flash, replies with corrections first rejected by the validator rose from 0.5\% to 6.2\%, unresolved placeholders from 5.8\% to 9.1\% and molecules taken from another reaction from 1.8\% to 3.4\% (Fig. 3D). On reaction scheme parsing (138 graphics~\cite{qian2023rxnscribe,chen2025rxnim}), the visual prompt raised F1 by 2.4 to 3.9 points, to 71.5\% with Gemini 3.7 Flash, against 65.2\% for the specialist tool alone and 40.9\% for Gemini 3.7 Flash alone (Fig. 3E).

For single molecules, the Molecular Recognition Agent, in which a program applies the model's corrections to the graph from a specialist tool, exceeded both of its parts on all five datasets (Fig. 3, F and G; fig.~S14): it reached 87.8\% and 78.9\% on explicit structures, against 81.9\% and 70.2\% for the tool alone and at most 32.1\% for MLLMs writing structures as SMILES (simplified molecular-input line-entry system) strings, and 80.5\% against 19.8\% for the tool on enumerated scaffolds, whose variants the program writes from the substituents the model names. Likewise, the Text Extraction Agent, in which specialist text models supply the entities and roles that the model corrects, exceeded both specialist models and MLLMs on chemical named-entity recognition and on reaction extraction from text (figs. S15 and S16).

\subsubsection*{\textit{Targeted verification through calibrated confidence}}

Even with exact copying, 18\% of ChemEAGLE's output reactions were incorrect, and a curator cannot tell which without checking them all. The Confidence Scorer estimates the probability that each reaction is correct by combining a language-model checker's answers about the graphic with features of the output and the extraction trace in a calibrated gradient-boosted classifier (fig.~S17; materials and methods). On held-out graphics, the scores were calibrated (expected calibration error 0.02) and separated correct from incorrect reactions (area under the receiver operating characteristic curve, AUROC, 0.91; Fig. 4A); without the checker the AUROC fell to 0.85, and the confidence that the base model reported itself was poorly calibrated (expected calibration error 0.12). Reviewing from the lowest score up found 68\% of the errors after 20\% of the reactions, against 20\% for random review (Fig. 4B). With three base models, high-confidence reactions (score $\geq$ 0.85) made up 62 to 65\% of the output and were 96 to 97\% correct, and low-confidence reactions (score \textless{} 0.5) were 19 to 30\% correct (Fig. 4C); on whole articles, high-confidence records were 90.8\% precise against Reaxys and low-confidence records 21.3\% (fig.~S12A).

In a workflow study, four annotators processed the full benchmark through the web application (Fig. 4D; fig.~S18; tables S10 and S11). Verifying and correcting every output took 0.18 h per graphic and reached 99.4\% accuracy, against 0.57 h and 98.4\% for manual annotation; verifying only the medium-confidence and low-confidence reactions, 35\% of the output, reached 96.6\% at 0.07 h per graphic.

The most common remaining errors were misread structures (43\%; fig.~S19), followed by differences in stereochemistry only (17\%) and in a counter-ion only (11\%) (Fig. 4E). The scorer rated 89\% of the erroneous reactions medium or low, including 99\% of those with an unresolved R group but only 86\% of misread structures, which yield a valid but wrong molecule (Fig. 4, E and F). Visual recognition therefore bounds both what ChemEAGLE extracts and what its scores can detect.

\subsubsection*{\textit{Training reaction prediction models on a photochemistry database built with ChemEAGLE}}

To test whether a database built without review is good enough to train models for new chemistry, we used ChemEAGLE (Gemini 3.7 Flash) to build one for photochemistry. Extracted from 1,856 articles, it holds 60,103 reactions, each with a confidence score: 6,919 conditions from 583 optimization tables in articles published up to 2024 and 53,184 reactions from 2,332 substrate-scope graphics, of which 43,190 met the criteria for modeling (Fig. 5, A and B; fig.~S20). Models were tested on unseen articles with records checked by hand: tables from 2025 and 2026, and scope graphics from held-out articles. As a control, each model was also trained on records that the same base model, Gemini 3.7 Flash, extracted from the same graphics end to end (Fig. 5C).

A ranking model trained on the extracted tables placed the highest-yielding condition first in 49.7\% of 198 test tables (95\% confidence interval {[}CI{]}, 43.3 to 56.2\%), against 10.8\% for a random order, and found it after 2.8 conditions instead of 6.4 (Fig. 5D). ReactionT5v2-yield, pretrained on yields from the Open Reaction Database~\cite{sagawa2025reactiont5,kearnes2021ord}, ranked the best condition first in 8.7\% of the tables, and in 6.5\% after fine-tuning, because it reads only structures, whereas 23\% of the conditions in a table share their structures with another condition. The signal came from the failed conditions: 21\% of the training conditions gave no product and only 10\% were the best of their table (Fig. 5B), and a gradient-boosted regressor trained only on the best condition of each table, which is what a database of reported conditions provides, ranked it first 39.1 percentage points less often than on full tables (95\% CI, 32.3 to 45.9; Fig. 5D).

For ranking, the accuracy of the structures mattered little. The end-to-end extraction recovered only 19.7\% of the table rows as hard matches with the right yield, against 98.1\% for ChemEAGLE (Fig. 5C), yet with species pooled regardless of role both trained equally good rankers (+0.9 points; 95\% CI, $-$2.4 to 4.3); keeping photocatalysts, catalysts and ligands apart, which only ChemEAGLE records, put it 7.4 points ahead (95\% CI, 2.5 to 12.3; Fig. 5E). Errors added to the training tables showed why (Fig. 5F; table S12): merged rows lowered the share ranked first by 27.3 points and lost roles by 6.7 points, whereas misread structures left it unchanged (+0.6). With 58 tables, however, ChemEAGLE's records gave a rank correlation that the end-to-end records reached only with about 370 tables (fig.~S21).

Scope graphics supply structures in breadth: each training graphic applied one transformation to a median of 19 substrates (Fig. 5B). ReactionT5v2-forward~\cite{sagawa2025reactiont5} predicted 6.8\% of 4,674 test products from unseen articles as its top-ranked prediction; fine-tuned on 34,254 extracted reactions, it predicted 32.7 $\pm$ 0.8\% (mean $\pm$ SD of three runs) and ranked the correct product among its five best candidates for 50.2 $\pm$ 1.0\% (Fig. 5G). Accuracy rose from 21.6\% for test reactions without a similar training reaction (Tanimoto similarity below 0.3) to 73.8\% for those with a close analog (Fig. 5H). Splits that place test reactions next to analogs from the same article, as in common benchmarks~\cite{schwaller2019transformer,kovacs2021interpretation}, gave 58.1 to 70.5\% (fig.~S22).

Unlike ranking, product prediction depended on exact structures. The end-to-end extraction recovered 13.7\% of the hand-checked scope reactions as hard matches (Fig. 5C), and the model fine-tuned on it predicted 17.5 $\pm$ 0.6\% of the test products, 15.2 $\pm$ 0.2 percentage points fewer than with ChemEAGLE's records (Fig. 5G). The gap persisted in the easier splits (fig.~S22) and was largest for test reactions with a close analog (37.6\% against 73.8\%; Fig. 5H). Added to the training scopes, misread structures lowered top-1 accuracy by 24.0 points, products on the wrong entry by 12.0, unapplied substituents by 8.0 and lost roles by 3.0 (Fig. 5I). Restricting training to higher-confidence records improved neither model (Fig. 5E; fig.~S22C).

\subsection*{Discussion}
We have shown that dividing the reading of a reaction graphic at two levels allows one system both to interpret graphics flexibly and to write exact structures. The pipelines, MLLMs and generalist agents that we compared each achieved only one of the two, and none estimated which of its reactions were correct. Because nearly every structure is a copy or a validated edit of a tool output, its origin can be traced and scored, and review can be restricted to the reactions that the scorer flags.

This division builds on tool-using and program-aided language models~\cite{yao2023react,schick2023toolformer,gao2023pal,gupta2023visprog}, in which, as in MERMaid~\cite{leong2025mermaid} and the generalist agents we tested, one model composes the final answer itself. ChemEAGLE instead distributes a graphic over agents that each own one kind of part, as multi-agent frameworks do for other tasks~\cite{hong2024metagpt,wu2023autogen}, and keeps structures out of the model's output through numbered marks~\cite{yang2023setofmark}, so that intermediate structures are tested against chemical constraints rather than by debate or voting~\cite{du2024debate}. Better tools alone do not explain the gains, because the generalist agents had the same tools, nor does the base model, as ChemEAGLE led with all seven; removing the visual prompt increased errors in binding judgments to tool outputs, the step on which the division depends. The division also sets the main limits. Errors in the bonds or skeleton of a recognized molecule cannot be corrected, and misreadings that yield a valid molecule were flagged less often than unresolved R groups (86\% against 99\%); substituents drawn without a fixed position cannot be written exactly, partly because standard SMILES has no notation for them, and 29\% of the reactions indexed for whole articles were not recovered (Fig. 2F).

The photochemistry database shows what such records are worth. Although it was built without review, models trained on it ranked conditions and predicted products in unseen articles, and the gain came from content that only graphics report: the failed conditions of each table, which databases rarely keep~\cite{striethkalthoff2022failed}, and the full scope of substrates. Which aspect of accuracy mattered depended on the task. Ranking needed species roles and rows that keep conditions apart but tolerated misread structures, whereas product prediction needed exact structures, so extraction accuracy alone predicts the value of records only in part.

The same division may apply to other graphics that are drawn for human readers yet can be checked exactly, such as materials property tables, polymer repeat units or metabolic pathway maps. With records that can be traced and scored, verifiable databases can be built at scale, and curation can scale with the number of uncertain records rather than with the size of the literature. Artificial intelligence could then learn from the published record of chemistry, not only its curated part, and guide discovery where data have been scarce.

\clearpage
\begin{figure}[p]
\centering
\includegraphics[width=\textwidth,height=0.68\textheight,keepaspectratio]{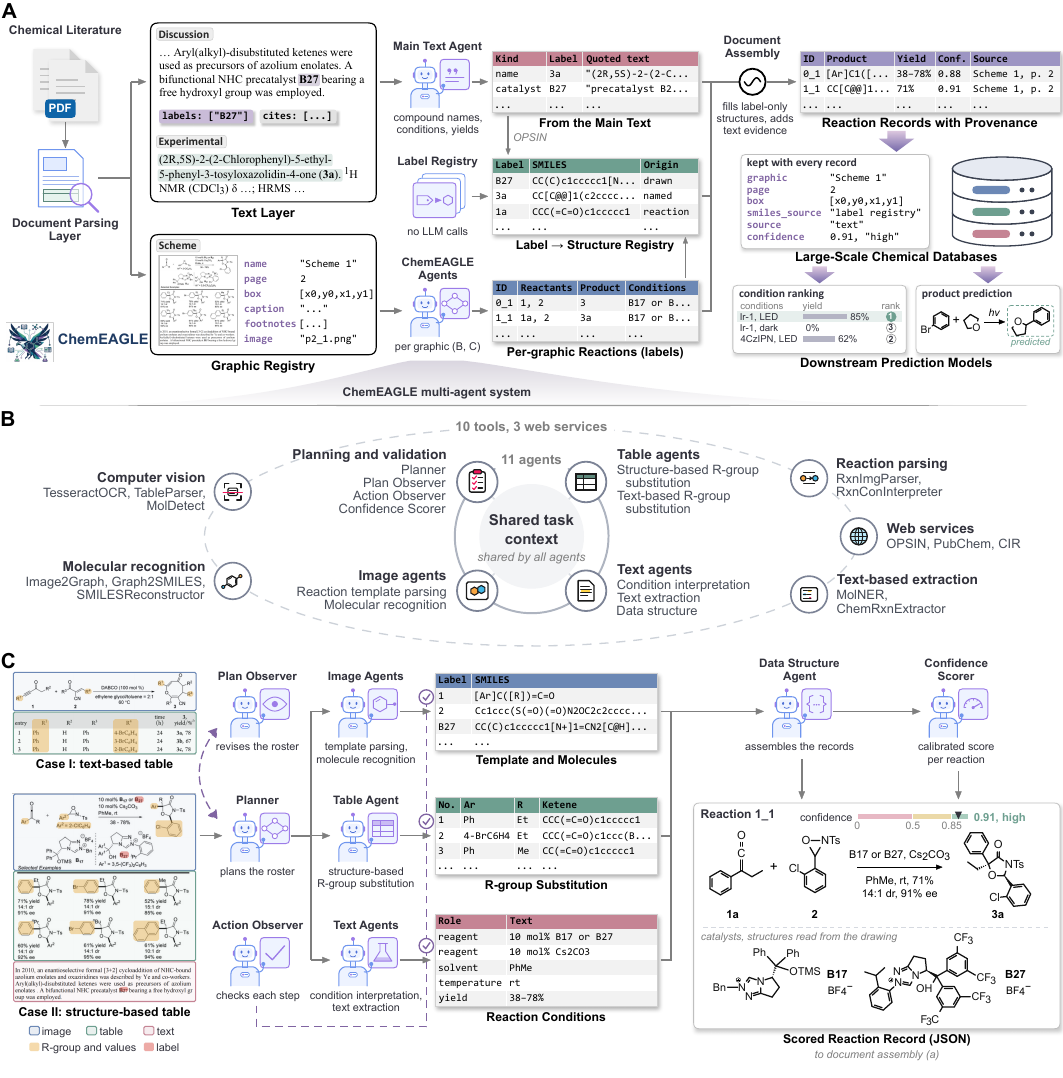}
\caption{\textbf{Overview of ChemEAGLE.} (\textbf{A}) A parsing layer splits each article PDF into a text layer and a registry of graphics. ChemEAGLE agents extract the reactions of each graphic (B and C), the Main Text Agent adds compound names, conditions and yields from the text, and the Label Registry assigns one structure to each compound label. Records, each with its graphic, page, box, structure source and confidence score, form large-scale databases that train downstream prediction models (Fig. 5). (\textbf{B}) The multi-agent system: 11 agents in four groups share one task context and call 10 specialist tools and 3 web services. (\textbf{C}) Extraction from one graphic, with Case I (a text-based table, from~\cite{liang2015dabco}) and Case II (a structure-based table, from~\cite{flanigan2015nhc}) as inputs; boxes mark image, table and text regions, and highlights mark R groups with their values and compound labels. The Planner selects the agents, the observers check the plan and each step, the image, table and text agents return molecules, R-group substitutions and conditions, the Data Structure Agent assembles the records, and the Confidence Scorer gives each reaction a calibrated score. The record shown is reaction 1\_1 of Case II; values in A and C are illustrative.}
\label{fig:main1}
\end{figure}

\begin{figure}[p]
\centering
\includegraphics[width=\textwidth,height=0.66\textheight,keepaspectratio]{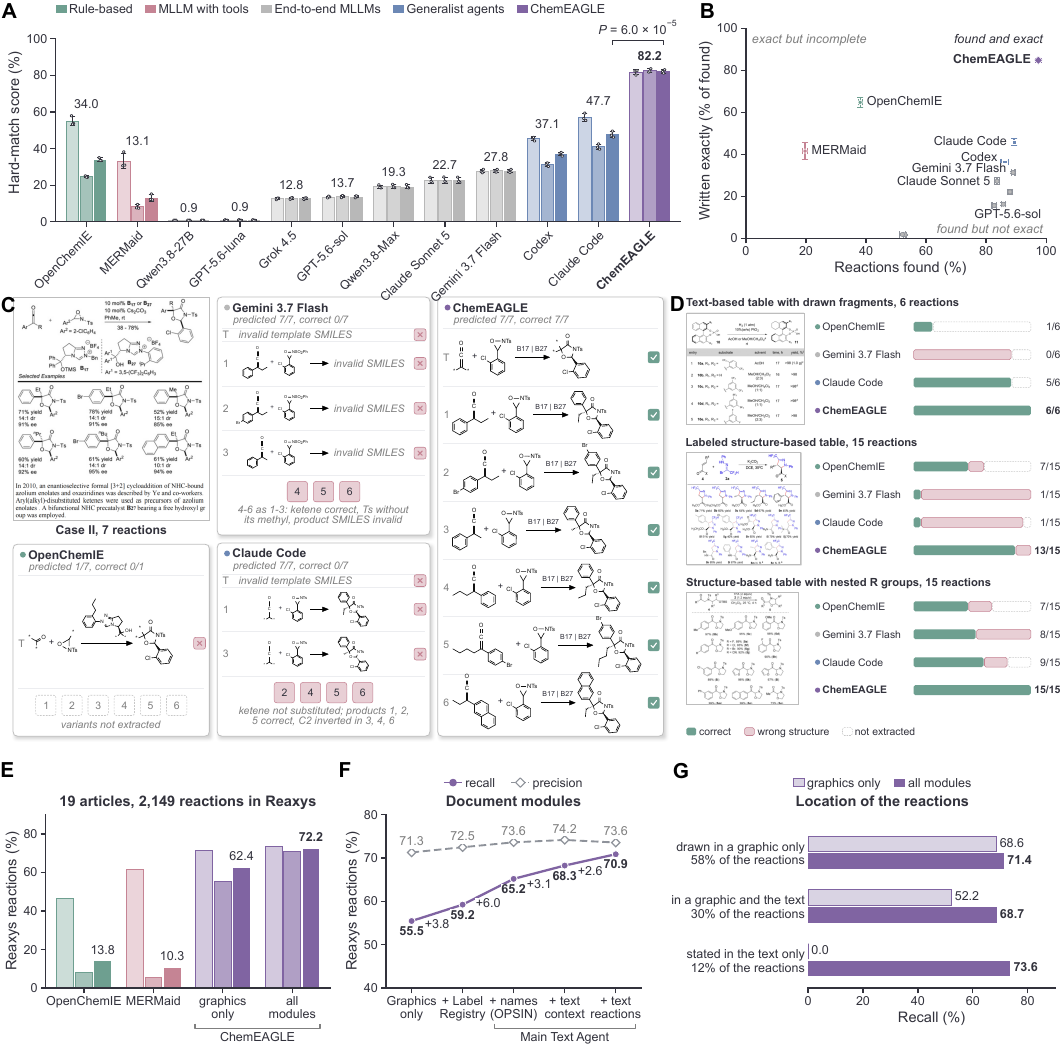}
\caption{\textbf{Comparison with other systems.} (\textbf{A}) Precision, recall and F1 (light to dark) under the hard-match criterion for 12 systems, colored by kind (soft match in table S7). End-to-end multimodal large language models (MLLMs): the seven base models of table S8; the generalist agents Codex and Claude Code, given ChemEAGLE's tools, run on GPT-5.6-sol and Claude Sonnet 5. Bars, means of three runs over all 3,613 reactions; dots, single runs; error bars, standard deviation; \emph{P} compares ChemEAGLE with Claude Code (two-sided Welch's t-test). (\textbf{B}) Reactions found (paired by product similarity) against found reactions written exactly; multiplying them gives the hard-match recall. (\textbf{C}) Outputs for Case II (seven reactions) from OpenChemIE, Gemini 3.7 Flash used end to end, Claude Code and ChemEAGLE; marks show correct and wrong reactions, and dashed slots missing ones. (\textbf{D}) Correct, wrong and missing reactions of the same systems on three graphics of other designs. (\textbf{E}) Precision, recall and F1 on 19 articles (2,149 Reaxys reactions) for OpenChemIE, MERMaid and ChemEAGLE with the graphics only or all document modules. (\textbf{F}) Recall and precision as document modules are added. (\textbf{G}) Reaxys reactions grouped by where the article reports them, and their recall. ChemEAGLE uses Gemini 3.7 Flash.}
\label{fig:main2}
\end{figure}

\begin{figure}[p]
\centering
\includegraphics[width=\textwidth,height=0.7\textheight,keepaspectratio]{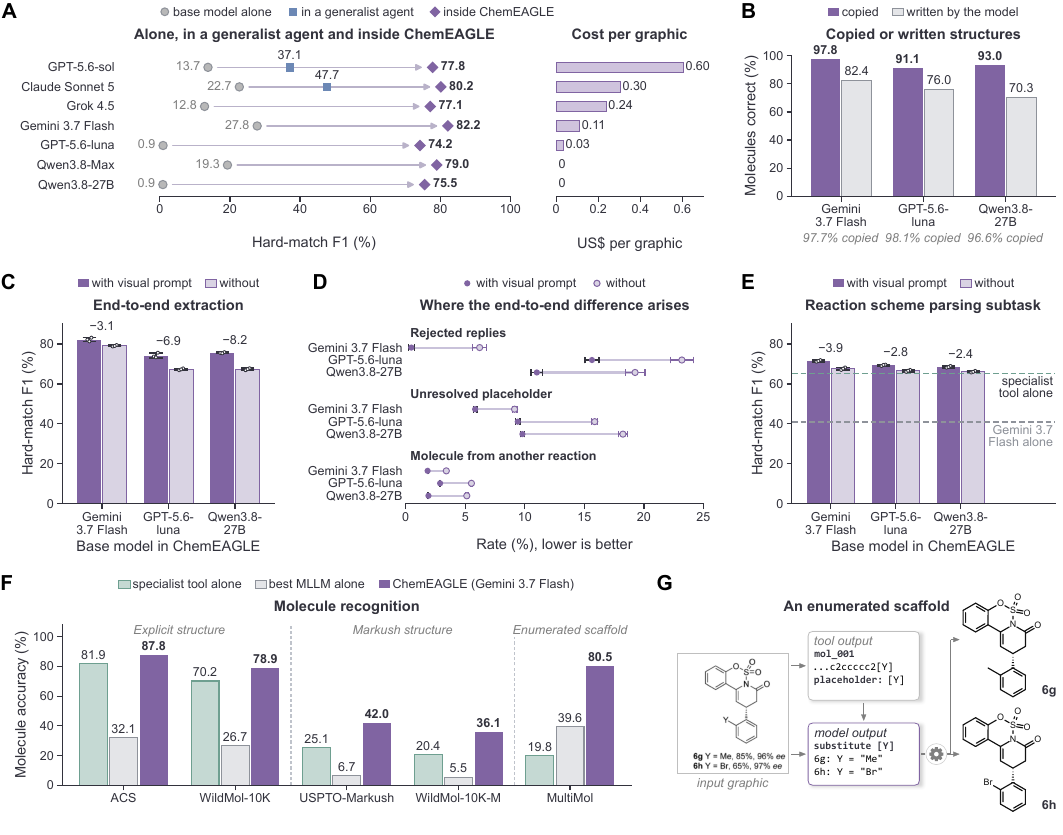}
\caption{\textbf{Structural exactness through reference to tool outputs.} (\textbf{A}) Hard-match F1 of the seven base models used alone (circles) and inside ChemEAGLE (diamonds), and of two driving a generalist agent (squares); right, cost per graphic. (\textbf{B}) Fraction correct of molecules copied from a tool output or written by the model. (\textbf{C}) Hard-match F1 with and without the visual prompt (without it, identifiers and boxes are given as text). (\textbf{D}) Error rates in the runs of (C): replies with corrections first rejected by the validator, reactions keeping an unresolved placeholder and molecules taken from another reaction. (\textbf{E}) The same ablation on reaction scheme parsing (138 graphics); dashed lines, RxnIM and Gemini 3.7 Flash alone. (\textbf{F}) Molecule accuracy on five datasets for the specialist tool alone (MolNexTR), the best multimodal large language model (MLLM) alone and ChemEAGLE (Gemini 3.7 Flash). (\textbf{G}) An enumerated scaffold: the tool reads one molecule with a placeholder {[}Y{]}, the model names the substituents and the program writes both structures. (A) to (E), means of three runs (error bars, standard deviation); \emph{P} = 0.017, 0.0054 and 7.4 $\times$ 10$^-$$^5$ in (C) and \emph{P} \textless{} 0.005 in (D) and (E) (two-sided Welch's t-test).}
\label{fig:main3}
\end{figure}

\begin{figure}[p]
\centering
\includegraphics[width=\textwidth,height=0.7\textheight,keepaspectratio]{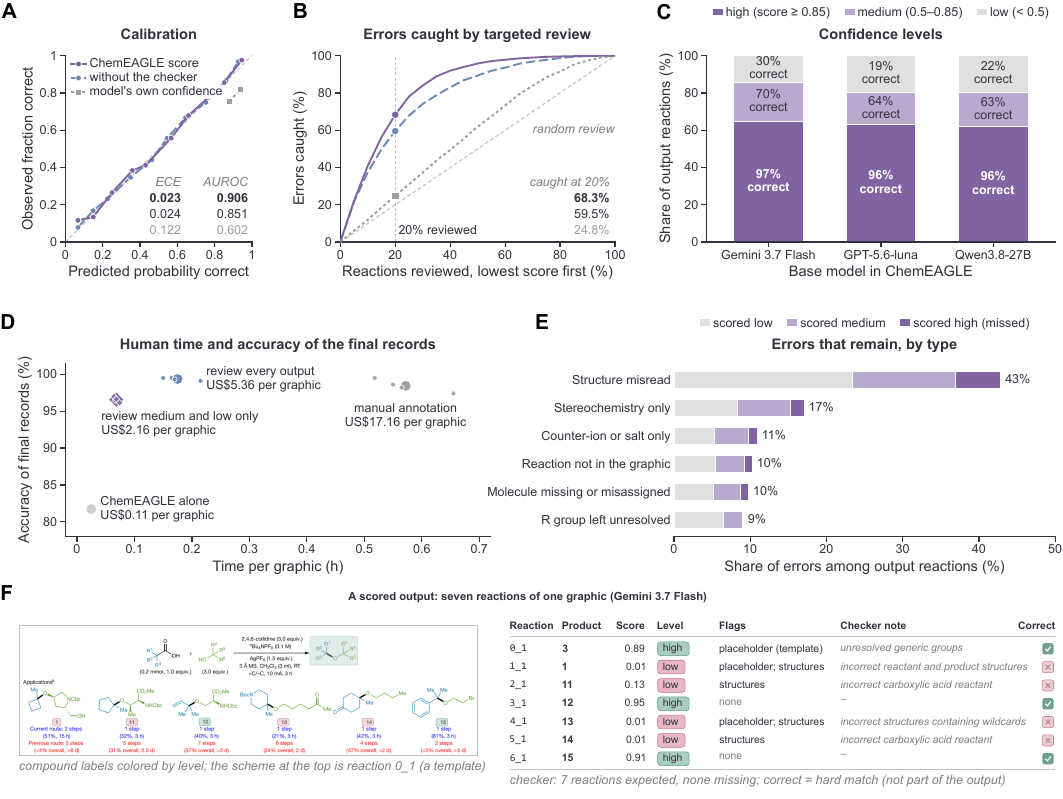}
\caption{\textbf{Calibrated confidence for targeted human review.} (\textbf{A}) Calibration on held-out graphics (Gemini 3.7 Flash): observed fraction correct against predicted probability in ten bins for the ChemEAGLE score, the score without the checker and the base model's self-reported confidence, with the expected calibration error (ECE) and the area under the receiver operating characteristic curve (AUROC). Scores are out-of-fold predictions of fivefold cross-validation grouped by graphic. (\textbf{B}) Share of erroneous reactions found when reactions are reviewed from the lowest score up; the diagonal is random review. (\textbf{C}) Share of output reactions in each level and their fraction correct, for three base models. (\textbf{D}) Time per graphic against the hard-match accuracy of the final records for four workflows; large markers are means and small points the four annotators (424 graphics), and ChemEAGLE alone is placed at its hard-match precision. Costs assume US\$30 per hour plus the cost of the run. (\textbf{E}) Errors among output reactions by type and level; errors scored high were not flagged. (\textbf{F}) A scored output for a graphic from~\cite{xiang2019hindered}: compound labels colored by level and, for each reaction, the score, level, flags, checker note and correctness (not part of the output).}
\label{fig:main4}
\end{figure}

\begin{figure}[p]
\centering
\includegraphics[width=\textwidth,height=0.7\textheight,keepaspectratio]{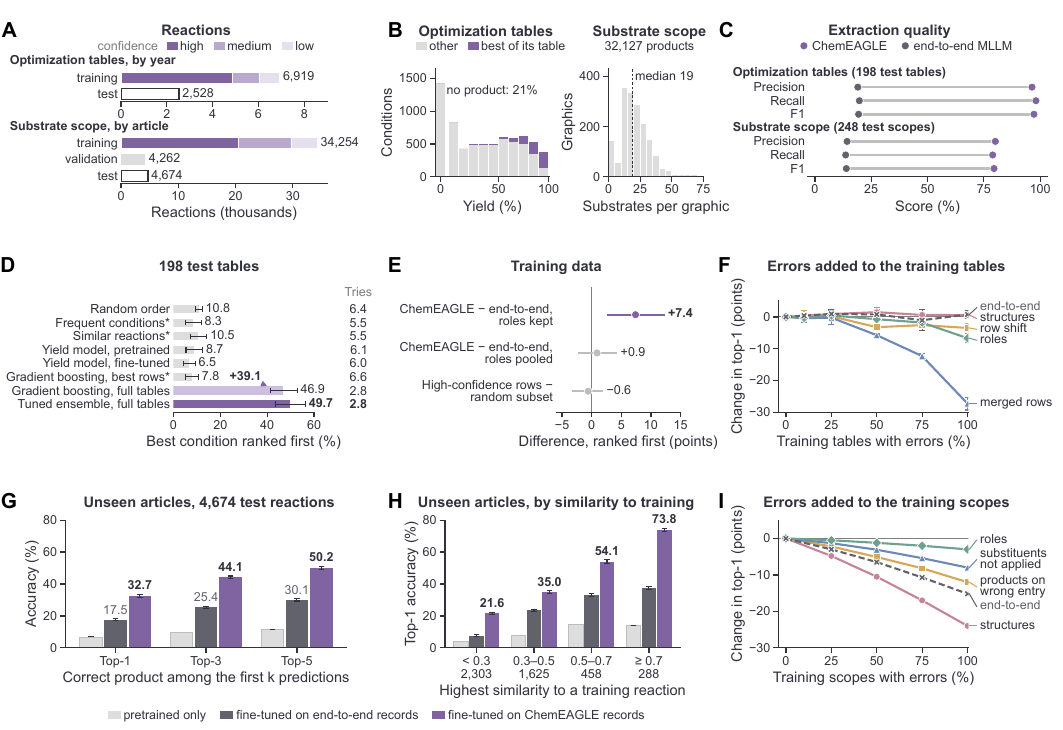}
\caption{\textbf{Prediction models trained on a photochemistry database built without review.} (\textbf{A}) Reactions per dataset and split, training reactions by confidence level. (\textbf{B}) Yields of training conditions (best of each table in purple) and substrates per training scope graphic (dashed line, median). (\textbf{C}) Hard-match precision, recall and F1: rows of 198 test tables (yield included), reactions of 248 test scope graphics. (\textbf{D}) Condition ranking on test tables: tables with the best condition ranked first (mean and 95\% bootstrap confidence interval) and conditions tried until the best one; asterisks, methods trained only on the best row of each table; arrow, gain from full tables in percentage points. (\textbf{E}) Paired differences: ChemEAGLE instead of end-to-end records, with species roles kept or pooled, and high-confidence rows only. (\textbf{F}) Change in the share ranked first when one kind of error is added to a growing share of training tables (mean and standard error, three runs). (\textbf{G}) Product prediction (ReactionT5v2-forward) on 4,674 reactions of unseen articles: correct product among the first one, three or five predictions. (\textbf{H}) Top-1 accuracy by the highest Tanimoto similarity to a training reaction. (G and H) Mean and standard deviation of three runs. (\textbf{I}) As (F), for product prediction.}
\label{fig:main5}
\end{figure}

\clearpage
\bibliography{science_refs}
\bibliographystyle{sciencemag}

\section*{Acknowledgments}
We thank the Information Technology Services Center of the Hong Kong University of Science and Technology for providing the HPC4 and SuperPod clusters as computational resources. AI-assisted tools were used to polish the language of the manuscript.
\paragraph*{Funding:}
Hong Kong University of Science and Technology project no. R9251 (HG).
\paragraph*{Author contributions:}
Conceptualization: YC, HG\\
Methodology: YC, HG, CTL\\
Benchmark construction: BY, YC\\
Computational experiments and code development: YC, CTL, BY\\
Visualization: YC, CTL\\
Project administration: HG\\
Supervision: HG, HC, JS, YH\\
Writing -- original draft: YC, CTL\\
Writing -- review \& editing: YC, CTL, HZ, LL, HG
\paragraph*{Competing interests:}
The authors declare that they have no competing interests.
\paragraph*{Data, code, and materials availability:}
The benchmark graphics, ground truth, predictions of every system, the document-level outputs and the photochemistry database are available at \url{https://github.com/CYF2000127/ChemEagle}. Reaxys records used for document-level evaluation cannot be redistributed; the list of articles and the matching code are provided. ChemEAGLE, including the document-level pipeline, the Confidence Scorer and its trained models, is available at the same repository, and the web application at \url{https://app.chemeagle.net/}. The data underlying the figures are given in data S1. No physical materials were generated in this work.

\subsection*{Supplementary materials}
Materials and Methods\\
Supplementary Text\\
Figs. S1 to S22\\
Tables S1 to S12\\
References \textit{(39--\arabic{enumiv})}\\
Data S1

\newpage

\renewcommand{\thefigure}{S\arabic{figure}}
\renewcommand{\thetable}{S\arabic{table}}
\renewcommand{\theequation}{S\arabic{equation}}
\renewcommand{\thepage}{S\arabic{page}}
\setcounter{figure}{0}
\setcounter{table}{0}
\setcounter{equation}{0}
\setcounter{page}{1}
\linespread{1.0}\selectfont 

\begin{center}
\section*{Supplementary Materials for\\ \scititle}

Yufan~Chen,
Ching~Ting~Leung,
Bowen~Yu,
Haifan~Zhou,
Jianwei~Sun,
Yong~Huang,
Linyan~Li,
Hao~Chen,
Hanyu~Gao$^\ast$\\
\small$^\ast$Corresponding author. Email: hanyugao@ust.hk
\end{center}

\subsubsection*{This PDF file includes:}
Materials and Methods\\
Supplementary Text\\
Figs. S1 to S22\\
Tables S1 to S12\\
Caption for Data S1

\subsubsection*{Other Supplementary Materials for this manuscript:}
Data S1

\newpage

\subsection*{Materials and Methods}

\subsubsection*{Design overview}

ChemEAGLE takes a reaction graphic, that is, an image that may combine a reaction template, a table of substrates or conditions, footnotes and a short passage of text, and returns a list of reaction records. Each record gives the reactants and products as simplified molecular-input line-entry system (SMILES) strings, the conditions with their roles (reagent, catalyst, solvent, temperature, time, yield), any further text attached to the reaction and a confidence score. The work is divided at two levels and then checked. Between agents, a Planner assigns the parts of the graphic to specialized agents. Within each agent, specialist tools (expert recognition models and chemistry software) supply the structures and a multimodal large language model (MLLM) supplies the judgment. The model refers to tool outputs by identifier and returns decisions, which programs apply to the tool outputs. Observer agents and deterministic checks review the result, and a Confidence Scorer estimates the probability that each reaction is correct. All agents use the same base MLLM, accessed through an OpenAI-compatible interface. Table S1 lists, for every agent, the tool calls that read the graphic, what the model returns and the tool calls that write the structures, and section ``Implementation of the agents, traced on two graphics'' follows two graphics through the agents step by step (Figs. S4 and S5).

\subsubsection*{Planning and routing}

The Planner receives the graphic and a description of the available agents and returns an ordered list of the agents to run. Its choices follow the content of the graphic. A drawn template calls for the Reaction Template Parsing Agent, and drawn product or catalyst structures call for the Molecular Recognition Agent. A substrate table goes to the Structure-based R-group Substitution Agent if its cells contain complete drawn molecules and to the Text-based R-group Substitution Agent if they give substituents, as names or as drawn fragments; a table that mixes names and drawn fragments is a special case of the latter. Condition text calls for the Condition Interpretation Agent, and a passage of prose calls for the Text Extraction Agent (Fig. S2). The Plan Observer then examines the graphic against this list and may add, remove or replace agents or change their order; the revised list replaces the original before any agent runs. The agents run in the planned order, each with the original graphic and the outputs of the agents before it. If the plan contains no R-group agent although the Molecular Recognition Agent has returned a molecule with a variable atom, the program runs the structure-based agent when at least five fully specified molecules are drawn and the text-based agent otherwise, and records this in the notes passed to the final assembly step.

\subsubsection*{Visual prompt and reference by identifier}

Before the models of the Molecular Recognition, Reaction Template Parsing and R-group substitution agents receive the graphic, a program marks on it the molecules found by the detection and recognition tools. Each molecule is outlined with a thin blue box (RGB 0, 90, 220) and tagged with a number placed just outside the box. The tag goes above the top left corner where possible, otherwise below the box, to the right of its top edge or to the left of the box, choosing the first position that stays on the image and covers no other molecule. Images whose shorter side is below 700 pixels are enlarged first so that the tags remain legible, and nothing is drawn inside a box. Tag N denotes molecule \texttt{mol\_00N} in the catalog given to the Molecular Recognition and Reaction Template Parsing Agents and molecule \texttt{mN} in the tool output given to the R-group substitution agents, and the prompts state this correspondence explicitly (Fig. S3). The models answer in terms of these identifiers and do not write SMILES for recognized molecules.

In the ablation without the visual prompt, the models of the Molecular Recognition and Reaction Template Parsing Agents and of both R-group substitution agents received the original graphic together with the same identifiers and the box coordinates as text, and nothing else was changed. Each of the three base models of Fig. 3, C to E, was run three times in this configuration. On the runs with and without the visual prompt we counted three error rates (Fig. 3D). The first is the share of the Molecular Recognition Agent's replies whose corrections the validator rejected on the first attempt. The second is the share of concrete reactions, that is, reactions that are not generic templates, that still hold a placeholder atom. The third concerns the reactants and conditions of found reactions (paired as for Fig. 2B) whose structure occurs in the ground truth of the graphic: the share that belongs only to another ground-truth reaction of the same graphic. The same ablation on reaction scheme parsing (Fig. 3E) used the 138 test graphics of the RxnScribe dataset and its scoring~\cite{qian2023rxnscribe}, with RxnIM~\cite{chen2025rxnim} alone and Gemini 3.7 Flash used end to end for comparison.

\subsubsection*{Specialized agents}

\emph{Reaction Template Parsing Agent.} RxnImgParser, built on RxnIM~\cite{chen2025rxnim}, detects the molecules, condition regions and their roles in the reaction template, including multistep schemes in which a product of one step is a reactant of the next. Image2Graph returns a graph for each molecule, and a program arranges these outputs into a catalog of the parsed reactions in which every molecule, label atom and condition text has a fixed identifier. The model receives the catalog and the graphic with the visual prompt and returns an edit plan in a fixed JSON schema: corrections of label atoms that OCR misread, of dropped charges and of look-alike elements; the molecules on which a charge is drawn; counter-ions; placeholders defined explicitly in the graphic; and, for each molecule with a variable, the decision to substitute it or keep it generic. The roles and the assignment of molecules to reactions are fixed by the parser, and lists of values are left to the R-group substitution agents. The program checks the plan against the catalog, applies it to every appearance of each molecule and writes the structures with Graph2SMILES. The condition regions are passed to the Condition Interpretation Agent.

\emph{Molecular Recognition Agent.} MolDetect locates every molecule in the graphic and links molecules to their printed labels; Image2Graph converts each molecule image into a graph of atom symbols, coordinates and bonds; and Graph2SMILES converts graphs into SMILES. The program arranges these outputs into a catalog in which every molecule, atom and text region has a fixed identifier. The model receives the catalog and the graphic with the visual prompt and returns its corrections in a fixed JavaScript Object Notation (JSON) schema with eight kinds of entry: corrections of atom symbols that optical character recognition (OCR) misread; corrections of charges and look-alike elements; corrections of text; definitions of abbreviations given in the graphic (for example Ar$^2$ = 2-ClC$_6$H$_4$) with the molecules they apply to; groups that expand a drawn core into the variants listed in text; the molecules on which a charge is drawn; counter-ions; and a decision for each molecule (expand, substitute, keep generic or review). The program checks every entry against the catalog. A correction must name an existing atom and repeat its current symbol, a definition must refer to a placeholder that the molecule carries, and every edited molecule must pass valence and ring checks and RDKit sanitization~\cite{rdkit}. A rejected reply is returned to the model with the validator's message for one retry. If the retry also fails, the entries that pass are applied and the others are dropped and recorded; if no entry can be applied, the tool result is returned unchanged and flagged for review. The program also removes isolated stray atoms and repairs up to four bond orders in rings of up to eight atoms when the recognized graph cannot be sanitized. Drawn molecules with printed labels are registered so that later steps can refer to them by label.

\emph{Structure-based R-group Substitution Agent.} This agent handles graphics that draw the product variants but not the corresponding reactants. The model receives the template, the recognized molecules with their identifiers and the graphic with the visual prompt. It returns, for each identifier, the printed label, the text printed next to the molecule and its role (template reactant, template product, product or condition). The program copies each structure from the tool output by identifier; keys that are not identifiers are accepted only if they reproduce a SMILES string of the tool output and are otherwise discarded. SMILESReconstructor then matches every product variant to the product template, cuts out the fragments at the variable positions and attaches them to the reactant template, so that each reactant variant is built from the drawn product.

\emph{Text-based R-group Substitution Agent.} This agent handles tables whose rows give substituents as names, formulas or drawn fragments. By calling the tools, the model obtains a catalog of the template's molecules with their label atoms, each marked as variable or not, the condition texts and, with identifiers, the molecules and fragments recognized by the Molecular Recognition Agent, which both R-group agents call on the graphic whether or not the plan lists it. It returns corrections of misread label atoms, each repeating the current symbol, and one entry per table row with the printed value of every variable (for example R = CO$_2$Me), the row's own conditions and its yield. Values are given as printed, never as SMILES. The program resolves each value to a fragment (see name resolution below), attaches it at the corresponding atoms of the template and builds one reaction per row.

\emph{Condition Interpretation Agent.} RxnConInterpreter, also built on RxnIM, reads the text of the condition regions and assigns each item a role; TesseractOCR~\cite{smith2007tesseract} supplies the raw text where needed. The model corrects the roles and the readings and links each catalyst or reagent drawn in the graphic to its drawn structure by label. Names that are not drawn are converted to structures by the name-resolution chain.

\emph{Text Extraction Agent.} For graphics that contain prose, TesseractOCR extracts the text, MolNER~\cite{fan2024openchemie} tags chemical entities and ChemRxnExtractor~\cite{guo2022chemrxnextractor} identifies reactions and their roles. The model aligns the mentions in the text with the structures found by the other agents and returns an annotated version of the passage. The agent was also evaluated on two text-only tasks with Gemini 3.7 Flash as base model (section ``Text Extraction Agent on text-only tasks'' and Figs. S15 and S16).

\emph{Data Structure Agent.} A final model call receives the outputs of all agents, the annotated text and the observers' notes, and assembles the reaction records in a fixed output schema, copying each structure from the agents' outputs. A reactant or product counts as copied when its canonical SMILES equals a structure returned by the agents' tool calls in the same run, and as written by the model otherwise. The records then pass through the deterministic checks described below, and any structure that does not occur in an agent output is flagged by the Confidence Scorer.

\subsubsection*{Observers}

The Plan Observer reviews the agent list as described above. The Action Observer checks the output of each agent after it runs against a checklist specific to that agent, for example that every recognized molecule has been labeled exactly once or that every table row has produced a reaction. It may not change the main body of a molecule, defined as the largest fragment of its SMILES with its atoms, bonds, ring positions and stereochemistry; a report that would change it is discarded by the program. The structure problems it can report are restricted to stray fragments, counter-ions and charges, unresolved placeholders and invalid SMILES. When the whole output of an agent has failed, the observer can request one re-execution of that agent. All other findings are passed as notes to the final assembly step, which checks each note against the graphic and applies or ignores it (checklists in section ``Prompts'').

\subsubsection*{Deterministic checks and name resolution}

After assembly, a program validates and repairs SMILES strings, including those that the final assembly step writes for common reagents and solvents named in the graphic, and resolves names that remain in the conditions, reactants and products. It attaches the drawn structure to any condition that names a labeled compound drawn in the graphic, shares drawn catalyst and reagent structures across the reactions of the graphic, and repairs charges and radicals that recognition has introduced.

Names are converted to structures by two fixed chains. Reagent and solvent names are looked up first in a curated alias map of frequent abbreviations, then in PubChem through PUG-REST with strict synonym verification~\cite{kim2016pubchem}, and last parsed with OPSIN~\cite{lowe2011opsin}. Substituent symbols inside recognized molecules are resolved by a built-in abbreviation table and a parser for condensed formulas, then by OPSIN, PubChem and the NCI/CADD Chemical Identifier Resolver, in that order. A name or symbol that none of these resolves is passed to a language model, whose proposal is accepted only if RDKit can parse it. An audit of the chain on the benchmark is given in section ``Audit of name-to-structure resolution'' and Table S4.

\subsubsection*{Tools}

The tools are described in full in section ``Specialist tools''. TesseractOCR reads text in graphics. MolDetect~\cite{fan2024openchemie}, retrained on additional annotated reaction images without architectural change (Table S2), detects molecules and links them to labels. Image2Graph, a retrained MolNexTR~\cite{chen2024molnextr}, predicts atom symbols, coordinates and bonds from molecule images. Graph2SMILES converts these graphs into canonical SMILES and resolves abbreviation nodes as described above. SMILESReconstructor uses RDKit substructure matching to transfer substituents from product variants to reactant templates. TableParser builds the reactions of a text-based table from the entries that the model returns for each row: it clones the template, replaces each label atom with the value printed in that row and regenerates the SMILES. RxnImgParser and RxnConInterpreter are built on RxnIM~\cite{chen2025rxnim} and parse reaction diagrams and condition text, respectively. MolNER and ChemRxnExtractor extract chemical entities and reactions from text~\cite{fan2024openchemie,guo2022chemrxnextractor}.

\subsubsection*{Confidence Scorer}

The scorer estimates, for each output reaction, the probability that it matches a ground-truth reaction under the hard criterion. Its deterministic features describe the reaction and the graphic. Reaction features include the numbers of reactants, products and condition structures, drawn conditions and invalid or missing SMILES; unresolved placeholders; fragments, net charge and metals, including charged condition structures, small stray fragments and neutral halogens beside a cation; whether the products and the other molecules were copied from an agent's output; duplicated products within the graphic; the size, rings and heteroatoms of the product and the numbers of its assigned and unassigned stereocenters; whether the reaction's text reports a selectivity (enantiomeric excess, enantiomeric ratio or diastereomeric ratio); and whether a yield is stated. Graphic features include the number of reactions, the share of molecules with placeholders, the number of distinct products, failed agents, whether the model's corrections were rejected or applied only in part, whether an R-group agent was added by the program and whether the Text Extraction Agent ran. Further features compare each reaction with the rest of its graphic: whether any reaction of the graphic reports a selectivity, the share of its reactions with unassigned stereocenters, and whether a product carries fewer assigned stereocenters than the template it was built from. An optional check asks the base model a fixed set of yes, no or unsure questions about the graphic and its extracted reactions, together with a one-sentence description of any problem and the number of reactions it expects; its answers are added as features. The questions ask whether the reactions are drawn in the graphic, whether the set is complete, whether the structures and conditions are correct, whether any placeholder is left unresolved, whether reactions are duplicated, whether a product is drawn with wedged or hashed bonds or a stated configuration, and whether a counter-ion given in the graphic is missing from the SMILES. The checker is not asked whether a configuration is right; the program compares its answer on drawn stereochemistry with the stereocenters left unassigned in the product. A gradient-boosted tree classifier with isotonic calibration maps the features to a probability. The models were trained on labeled output reactions of the benchmark. All scores reported here are out-of-fold predictions of fivefold cross-validation grouped by graphic, so that no graphic is scored by a model trained on it. The probability is reported with a level, high ($\geq$ 0.85), medium (0.5--0.85) or low (\textless{} 0.5), and with flags that name the features responsible, for example unresolved placeholder, invalid or missing SMILES, unassigned stereocenter, unbalanced charge or molecule not from a tool. Scoring never alters the extracted records.

\subsubsection*{Document-level extraction}

For whole articles, a document layer runs before and after the per-graphic extraction. The text layer of the PDF is read with pdfplumber~\cite{pdfplumber} in reading order, with sections identified and running headers and footers removed. Figure and table regions are detected with VisualHeist~\cite{leong2025mermaid}, and each is registered with its name, caption, footnotes, page and bounding box; paragraphs are registered with the figures they cite and the compound labels they mention. None of these steps calls a language model. Text-only model calls, each covering up to 25 figures, classify the role of every figure from its caption and the sentences that cite it; a classification whose quoted evidence does not occur verbatim gives way to a rule on the caption or printed text when one applies, and a figure without any text is classified from its image. Spectra are not extracted, although the recognition tools read the numbered structures drawn beside them, and reactions in mechanism figures are marked as proposed steps. ChemEAGLE is then run on every figure that contains reactions.

A Label Registry maps each compound label in the article to a structure. It prefers a drawn structure, then a structure parsed with OPSIN from a systematic name in a characterization paragraph, then one read from a spectrum figure and last one inferred from a reaction; within the same source, the structure seen in most figures is kept, and a label that names several compounds (two structures in one figure, or three or more in the article) is not used. The Main Text Agent adds what the figures leave out. It extracts systematic names of labeled compounds from characterization paragraphs and converts them with OPSIN. It makes one text-only model call per figure to collect conditions, yields and selectivities stated in the text, keeping only items whose quoted evidence occurs verbatim. It also reads reactions described only in the text, taking their structures from the Label Registry or OPSIN rather than from the model. Document assembly fills molecules that carry only a label with the registered structure and replaces structures built from a template row with it, while a structure that the figure draws itself is kept and a different registered structure is recorded as a conflict; it also attaches the text evidence and records for every reaction the figure, page and paragraph from which it came, in one JSON file per article. Table S1 lists what the tools, the model and the programs do in each of these steps. Document-level results were scored against Reaxys with the protocol and 19-article set of~\cite{fan2024openchemie} (section ``Document-level evaluation against Reaxys''). For Fig. 2F the four document modules were switched on one after another; for Fig. 2G each Reaxys reaction was located in the article as drawn in a graphic only, in a graphic and the text, or stated in the text only.

\subsubsection*{Benchmark and ground truth}

The benchmark contains 424 reaction graphics with 3,613 reactions in four subsets: 139 graphics with 1,403 reactions from 64 recent research articles; 78 graphics with 1,085 reactions from 48 articles in J. Org. Chem., Org. Lett., Asian J. Org. Chem., Eur. J. Org. Chem. and Angew. Chem. used in the OpenChemIE benchmark~\cite{fan2024openchemie}; 107 graphics with 495 reactions from a review of N-heterocyclic carbene catalysis~\cite{flanigan2015nhc}; and 100 graphics with 630 reactions from 14 patent documents of the United States Patent and Trademark Office (USPTO), the European Patent Office (EPO) and the World Intellectual Property Organization (WIPO). The 217 graphics of the research-article and OpenChemIE subsets come from 22 journals of six publishers (Fig. S7A).

Ground truth was established by independent annotation by four chemists, from senior undergraduate to senior doctoral level (Table S5), who followed a common written guideline, output schema and canonicalization procedure. Across the 3,613 reactions, the six pairwise F1 scores between annotators ranged from 96.5\% to 98.8\% (mean 97.6 $\pm$ 1.0\%), all four annotators produced identical records for 94.5\% of the reactions, and Gwet's AC1 was 0.968~\cite{gwet2008ac1} (Table S6). The remaining 5.5\% of the reactions were resolved by a documented adjudication protocol (section ``Ground-truth annotation''). Every reaction was then audited against its graphic. In the article subsets, the audit changed 2,661 fields in 1,384 reactions of 216 graphics, most often the catalyst and product structures and the record of whether a condition structure is drawn. In the patent subset, two audit rounds, one for stereochemistry and one for connectivity, changed 299 entries in 34 graphics. Five graphics were edited before use (two binarized, three re-cropped and rescaled), and the original files are released alongside the edited ones.

\subsubsection*{Evaluation}

A predicted reaction is matched to at most one ground-truth reaction by optimal assignment. Under the soft criterion, the products must match as multisets and the left side must match without regard to slot. Every ground-truth reactant must occur among the predicted reactants and condition structures, and every predicted reactant must occur somewhere on the ground-truth left side, so a molecule filed as a reactant on one side and as a condition on the other is not an error. Under the hard criterion, used throughout the main text, every condition structure that the graphic draws must also occur on the predicted left side (Fig. S8). Whether a condition is drawn is recorded in the ground truth from a visual check of each graphic, and conditions given only as names do not take part. Molecules must match exactly after canonicalization, including stereocenters and every fragment, such as counter-ions. Precision, recall and F1 are computed over all reactions of a subset.

For Fig. 2B, the predicted and ground-truth reactions of each graphic are paired one to one by maximum product similarity (Tanimoto similarity of Morgan fingerprints of radius 2, computed on the union of each reaction's products; Hungarian assignment). A ground-truth reaction counts as found when its partner reaches a similarity of at least 0.5. The fraction of reactions found and the fraction of found reactions that are hard matches multiply to the hard-match recall, so the two separate missing a reaction from writing it incorrectly. For Fig. 4E, output and ground-truth reactions are matched on the hard criterion first and the rest by product similarity as above. Each wrong output reaction is assigned one of six types: unresolved R group, misread structure, stereochemistry only, counter-ion only, molecule missing or misassigned, and reaction not in the graphic. When several apply, the type is taken in the order unresolved R group, molecule from another reaction, misread structure, stereochemistry only, counter-ion only, missing or extra molecule, and reaction not in the graphic. Confidence scores are evaluated by the area under the receiver operating characteristic curve, the expected calibration error over ten equal-width bins, and the share of erroneous reactions found when reactions are reviewed from the lowest score up.

\subsubsection*{Controlled comparisons and ablations}

Each ablation changed one component of the reference configuration and kept the code version, base model and graphics fixed. The visual-prompt ablation is described above. In the observer ablations, the Plan Observer, the Action Observer or both were switched off, with Gemini 3.7 Flash and three runs on all 424 graphics (Fig. S6). In the first run of the full system, every plan revision and every check in which the Action Observer flagged an issue was audited against the graphic and the ground truth, and all 424 graphics were screened for plan or agent errors that the observers missed (section ``Audit of the observers'' and Table S3). Molecular recognition was evaluated on five datasets: ACS and WildMol-10K (explicit structures from the literature)~\cite{qian2023molscribe,fang2024molparser}, USPTO-Markush and WildMol-10K-M (Markush structures)~\cite{morin2025markushgrapher,fang2024molparser} and MultiMol, a dataset of 50 images of enumerated scaffolds that we built from Nature, Science Advances, Angew. Chem. Int. Ed., J. Org. Chem., Org. Lett., Asian J. Org. Chem. and Eur. J. Org. Chem. A molecule counts as correct when its canonical SMILES equals the reference. The specialist recognizers were run with their released weights, and the MLLMs were asked for the SMILES of each image (Fig. 3F and Fig. S14). With Gemini 3.7 Flash as base model and three runs per condition, we removed each extraction agent's in-context example, replaced it with one of two alternative examples, gave all three examples together, shuffled the rule blocks, replaced the role definitions with neutral statements or removed the chemical-notation rules (Fig. S13 and section ``Robustness to the in-context example and the prompt'').

\subsubsection*{Base models and inference settings}

ChemEAGLE accesses its base model through an OpenAI-compatible client. The reference configuration uses Gemini 3.7 Flash. We also evaluated GPT-5.6-sol, GPT-5.6-luna, Claude Sonnet 5, Grok 4.5 and the open-weight Qwen3.8-Max and Qwen3.8-27B, the latter served with vLLM~\cite{kwon2023vllm} with thinking mode enabled. The temperature was 0 wherever the interface accepts it, and the reasoning effort was the provider's default, except for the GPT-5.6 models, which were run without reasoning. Every configuration was run three times on all 424 graphics. The Molecular Recognition and Reaction Template Parsing Agents and the confidence check request JSON output from the interface, the R-group agents obtain the tool outputs through tool calls and reply with one JSON object, and failed calls are repeated. The prompt of each extraction agent contains one worked example (for the Structure-based R-group Substitution Agent and the Data Structure Agent, a pair of short examples used as one), written for illustration; none is taken from a benchmark graphic. Prompts and examples were developed on 20 graphics that do not overlap with the benchmark, the same example serves every graphic, and the prompts are reproduced in section ``Prompts''.

\subsubsection*{Baselines}

OpenChemIE~\cite{fan2024openchemie} and MERMaid~\cite{leong2025mermaid} were run with their released code and default settings. The end-to-end MLLMs, the seven base models above, each received the graphic and the same output schema as ChemEAGLE in one call. Claude Code, driven by Claude Sonnet 5, and Codex, driven by GPT-5.6-sol, were given ChemEAGLE's chemical toolset through a Model Context Protocol server exposing twelve tools. These covered molecule detection, molecular image parsing (Image2Graph and Graph2SMILES), reaction image parsing, molecule--label coreference, optical character recognition, R-group reconstruction, name-to-structure resolution, SMILES validation, condition interpretation, text-based reaction extraction and chemical named-entity recognition. Both agents received a task prompt written for them, with three worked examples, ChemEAGLE's output schema and the tool documentation. They could view the graphic through a file-reading tool limited to the image directory, had no other file, shell or web access and planned their own steps. Each graphic was processed in an independent session with a limit of 30 min. Each baseline was run three times, and all outputs were scored with the evaluation described above.

\subsubsection*{Statistical analysis}

Every configuration was run three times, and we report the mean and the sample standard deviation over the runs. For the benchmark of 424 graphics, run k on the 324 article graphics was pooled with run k on the 100 patent graphics before precision, recall and F1 were computed. ChemEAGLE was compared with the strongest alternative by a two-sided Welch's t-test on the F1 scores of the three runs of each system. The same test compares configurations in the ablations and the error rates of Fig. 3D.

\subsubsection*{Human workflow study}

Four chemists (profiles in Table S5) processed the complete benchmark twice, once by verifying and correcting ChemEAGLE's output and once by annotating from scratch. The verification pass came first, so that any learning effect favors annotation from scratch. Wall-clock time was recorded per graphic through the web interface (section ``Web application'' and Fig. S18), with verification and correction logged separately (Table S11). Costs combine the measured times with a labor rate of US\$30 per hour and the cost of automated extraction with Gemini 3.7 Flash (US\$0.1106 per graphic; Fig. 3A). The time needed to review only medium-confidence and low-confidence reactions was measured in the same way by the same four annotators (section ``Procedure and costs of the human workflow study'' and Table S10). In Fig. 4D, ChemEAGLE alone is the output of ChemEAGLE with Gemini 3.7 Flash without review, whose accuracy is its hard-match precision (Fig. 2A).

\subsubsection*{Construction of the photochemistry database}

Articles on photochemical reactions were collected by keyword searches, downloaded in bulk and screened, which gave 1,856 articles. ChemEAGLE, with Gemini 3.7 Flash as base model and both observers, extracted their optimization tables and substrate-scope graphics, with condition roles that distinguish photocatalysts, catalysts, ligands, reagents and solvents, and its records were not reviewed. The rows of a graphic that share reactants and product form a table, and a table with at least three rows and two distinct sets of conditions counts as an optimization table. Yields were read as numbers: not detected as 0\%, trace as 1\% and a range as its midpoint. For modeling, rows with identical conditions but different yields, rows that summarize several conditions and rows without a yield were removed, each table had to keep at least three rows with at least two distinct yields, and a table repeated in another graphic was used once. Tables from articles published up to 2024 formed the training set (583 tables with 6,919 conditions), and those from 2025 and 2026 the test set (198 tables, 118 from 2025 and 80 from 2026).

The 2,332 substrate-scope graphics gave 53,184 reactions. They were split by article into training (1,844 graphics), validation (240) and test (248) sets, so that the validation and test graphics come from articles unseen in training. A reaction was used when all its structures could be parsed, it contained no placeholder, it had at least one reactant and one product, its product was not among its reactants and it did not repeat another reaction of its set. A validation or test graphic with at least half of its reactions in the training set was taken to repeat a training article, and the 43 training graphics concerned were removed; validation and test reactions that also occur in the training set were not scored. This left 34,254 training, 4,262 validation and 4,674 test reactions (43,190 in total), the test reactions from 241 graphics. A fixed set of 1,000 validation reactions served to select the number of training epochs.

The test records were checked against their graphics entry by entry. In the optimization tables, roles followed the table headers, except that the only light-absorbing species of a reaction counted as its photocatalyst; a condition counted as the authors' standard only when it was marked as such; and rows whose conditions could not be distinguished were excluded. Confidence intervals were obtained by bootstrap resampling of the test tables for condition ranking and of the test graphics for product prediction.

\subsubsection*{Training reaction prediction models on the photochemistry database}

Conditions were ranked by an ensemble of three gradient-boosted tree models trained on the extracted optimization tables: a LambdaRank ranker and a regressor of the yield rank of each condition within its table, both in LightGBM~\cite{ke2017lightgbm}, and a pairwise ranker in XGBoost~\cite{chen2016xgboost}. Their scores were converted to ranks within each table and averaged. Baselines ranked conditions at random; by how often their species, light, atmosphere and temperature occurred among the best conditions of the training tables; by the same counts over the best conditions of the 25 training tables with the most similar reactions (Tanimoto similarity of reaction fingerprints), weighted by similarity; with a gradient-boosted regressor trained on the best condition of each table only; and by the yield that ReactionT5v2-yield~\cite{sagawa2025reactiont5} predicts, as released or fine-tuned on the training tables. Ties were broken at random. Products were predicted by ReactionT5v2-forward~\cite{sagawa2025reactiont5}, fine-tuned on the extracted substrate-scope reactions.

Extraction quality (Fig. 5C) was scored with the hard criterion described above, with two adaptations. The photochemistry records do not mark whether a condition is drawn, so photocatalysts, catalysts and ligands were treated as drawn conditions and reagents and solvents as named only, and a row of an optimization table also had to carry the yield of the hand-checked record. A table counted as having the right photocatalyst when every photocatalyst structure of its hand-checked record occurred in the extraction, in any role (Fig. S20B). The hand-checked structures of 891 scope reactions had been taken from the end-to-end extraction; without these reactions, the end-to-end extraction reached a precision of 9.1\%, a recall of 9.8\% and an F1 of 9.4\%, against 14.3\%, 13.7\% and 14.0\% with them. The errors added to the training data (Fig. 5, F and I) are defined in Table S12, and further results are given in section ``Further results on reaction prediction models trained on the photochemistry database''.
\subsubsection*{Division of work in each agent}
ChemEAGLE divides the reading of a graphic between agents and, within each agent, between specialist tools and the
base model (section ``Design overview''). Table~S1 lists, for every agent, the tool calls that read the graphic,
what the model returns and the tool calls that write the structures from the model's answer, as in Fig. S1C. In the
agents that handle drawn molecules, the model refers to tool outputs by identifier and returns decisions, and a second
tool call writes the structures. The last four rows give the same division for the document layer that handles
whole articles (Fig.~1A). The agents are described in
full in section ``Specialized agents''. Section ``Implementation of the agents, traced on two graphics'' follows two graphics through the agents with the
tool outputs, model replies and program steps of one run, and the prompts are reproduced in section ``Prompts''.


\subsubsection*{Implementation of the agents, traced on two graphics}
This section describes how the agents are implemented in the released code and follows two graphics through the system.
Case II (Fig.~1C and Fig. S1) draws a reaction template, two catalysts and six product variants above a passage of text.
Case I (Fig.~1C and Fig.~S3A) is a substrate table whose eleven rows print the substituents
R\textsuperscript{1}--R\textsuperscript{4} of a drawn template. Both graphics were run once with the released code,
Gemini 3.7 Flash as base model and both observers. Figs.~S4 and S5 show what
each tool returned, what the model replied and what the programs did with the reply. All values come from these runs, except the replies of the Molecular Recognition Agent: the run log records only their size, so the replies shown were obtained by calling that agent once more on the same graphic.

\paragraph*{Dispatch}
The Planner returns the names of the agents as text. A program maps each name to an agent, removes duplicates and keeps
the planned order. Three rules then apply. The Text Extraction Agent is moved to the end, because it receives the
result of the first graphical agent as input. When the plan contains an R-group substitution agent, the Reaction
Template Parsing, Molecular Recognition and Condition Interpretation Agents are removed from the list, because the
R-group agent calls their tools itself. A plan in which no agent can be recognized falls back to the Reaction Template
Parsing Agent. In Case II the Planner named five agents and the Plan Observer kept them; the program ran the
Structure-based R-group Substitution Agent and then the Text Extraction Agent. In Case I the Planner named the Reaction
Template Parsing, Condition Interpretation and Text-based R-group Substitution Agents, and the program ran the last of
these alone (Figs.~S4A and S5A). An agent that fails is skipped: the results
of the other agents are kept, and a note naming the failed agent is passed to the Data Structure Agent; a run stops
only when every agent has failed. After the planned agents, the program runs an R-group agent that the plan omitted if a
recognized molecule carries a variable atom (section ``Planning and routing'').

The vision tools run once per graphic and base model. Their results are cached, and every agent that needs them reads
the same reconciled copy, so the molecules of a graphic keep one set of identifiers through a run: tag N on the visual
prompt, \texttt{mol\_00N} in the catalog of the Molecular Recognition Agent and \texttt{mN} in the tool output that the
R-group agents receive.

\paragraph*{Molecular Recognition Agent}
MolDetect finds the molecules and their labels, Image2Graph returns a graph for each molecule, and OCR reads the text
regions. From these outputs a program builds the catalog: each molecule with its SMILES and the atoms that carry a
label, an abbreviation or a placeholder, each with an atom identifier such as \texttt{mol\_004:a013}; each text region
with its OCR reading; and the links between molecules and labels. The model receives the prompt, the graphic with the
visual prompt and the catalog, and must reply with its corrections under a strict JSON schema. In Case II
(Fig.~S4B) the edit plan turned the bare placeholder of the ketene into R and corrected the
label read as Ar on the two aryl groups of B27 to Ar\textsuperscript{1}. It replaced the OCR reading of the definition
under B27 (``Ar1'', ``5'', ``3,5-(CF3)CeH3'') by Ar\textsuperscript{1} =
3,5-(CF\textsubscript{3})\textsubscript{2}C\textsubscript{6}H\textsubscript{3} and repaired two yield lines, defined
Ar\textsuperscript{1} and Ar\textsuperscript{2}, gave B17 and B27 their tetrafluoroborate counter-ions, and decided to
substitute eight molecules and to keep the two templates generic.

The program checks each entry against the catalog before it applies anything. A correction must name an existing atom
and repeat its current symbol; a definition must refer to a placeholder that the molecule carries; and every edited
molecule must pass valence and ring checks and RDKit sanitization. Accepted entries are applied to the graph, not to a
SMILES string: the program writes the new symbols into the atom list, adds each counter-ion as a separate node and
regenerates the SMILES with Graph2SMILES, which expands abbreviations such as Ts or
2-ClC\textsubscript{6}H\textsubscript{4}. An edit plan that fails a check is sent back to the model once with the validator's
message. If the second edit plan also fails, its valid entries are applied and the others are dropped and recorded; only
when no entry can be applied is the tool output returned unchanged and marked for review. Stray isolated atoms are then
removed, ring bond orders are repaired where the graph cannot be sanitized, and every molecule with a printed label is
registered so that later steps can find its structure by label. In Case I the edit plan contained no correction. Its three molecules form the template, whose substituents are printed in the
table, so the model kept them generic and left their substitution to the Text-based R-group Substitution Agent.

\paragraph*{Reaction Template Parsing Agent}
RxnImgParser returns the roles in the reaction template: which boxes are reactants and products, and which regions hold
condition text, and Image2Graph returns a graph for each molecule. A program builds a catalog of the parsed reactions with fixed identifiers for every molecule, label atom and condition text. The model receives the catalog and the graphic with the visual prompt (prompt in section ``Prompts'') and returns an edit plan: corrections of label atoms, charges and look-alike elements, counter-ions, placeholders defined explicitly in the graphic and a decision for each molecule with a variable. The roles are fixed by the parser. The program checks each entry against the catalog, applies the plan to every appearance of each molecule and writes the structures with Graph2SMILES. A second check compares the two tools box by box and
replaces an invalid graph on either side by the valid graph of the same box from the other tool; it made no repair in
either graphic (19 and 6 boxes). Placeholders that the two sides of a template name differently, for example with a
prime on one side only, are renamed so that both sides use one name. A molecule printed as text that no detector box
covers, such as CS\textsubscript{2} written as a reactant, keeps RxnIM's graph; if that graph is degenerate, one short
model call receives the crop and the symbol list and must return the same number of symbols, and the new symbols are
kept only if Graph2SMILES turns them into a valid SMILES.

\paragraph*{Structure-based R-group Substitution Agent}
In Case II the model received the graphic with the visual prompt and called the tools that return the recognized
molecules with their identifiers, the reaction template and the condition text. It replied with one entry per
identifier, giving the compound label, any text printed beside the molecule and its role
(Fig.~S4D); for product \texttt{m5} the entry was
\texttt{["3a", "71\% yield, 14:1 dr, 91\% ee", "product"]}. Case II labels only its catalysts (B17 and B27), so the model numbered
the other molecules as the prompt prescribes: 1--3 for the template and 3a--3f for the product variants in reading order.
The reply contains no SMILES. The program copies each
structure from the tool output by its identifier. A key that is not an identifier is accepted only if it reproduces a
SMILES string of the tool output and is otherwise discarded, and molecules that the model left out are
added back from the tool output with the label that MolDetect linked to them. A reply that is not valid JSON is
requested once more. SMILESReconstructor then matches each product variant to the product template, cuts out the
fragments attached at the variable positions (R and Ar in Case II) and attaches them at the same positions of the
reactant template. In Case II it built the six ketenes from the six drawn products, giving six variant reactions
beside the template reaction.

\paragraph*{Text-based R-group Substitution Agent}
The model of this agent receives a catalog in place of the recognized graphs (Fig.~S5C). The
catalog lists each template molecule with an identifier, its SMILES and its label atoms, each with an atom identifier
and a flag that marks whether its symbol is a variable such as R\textsuperscript{4}, together with the condition texts
read by OCR and the recognized molecules with their identifiers. The model returns corrections of misread label atoms
and one entry per table row. In Case I, Image2Graph had read R\textsuperscript{1} and R\textsuperscript{2} of reactant
1 as numbered attachment points, \texttt{[1*]} and \texttt{[2*]}, which name no variable; the model corrected both,
repeating the current symbols. Each of its eleven rows gave the values of R\textsuperscript{1}--R\textsuperscript{4} as
printed (Ph, H, Ph and 4-BrC\textsubscript{6}H\textsubscript{4} in row 1), the conditions with their roles and the
other printed columns (product 3a, yield 78\%). The model also corrected the OCR readings of the conditions
(``ethylene glycolltoluene = 2:'' to ``ethylene glycol/toluene = 2:1'', ``60 *C'' to ``60 $^\circ$C'') and added the reaction
time of each row.

TableParser applies the reply. Each correction must name an existing atom that carries the symbol the correction
repeats, and the new symbol must be a label; otherwise it is skipped. For each row the program clones the
template, writes each printed value at the atom of its variable, regenerates the SMILES with Graph2SMILES, which expands
Ph and 4-BrC\textsubscript{6}H\textsubscript{4} as it does any abbreviation, and keeps the row's own conditions. A value
that cannot form a label is left as drawn. When a row names a compound only by its label, the values that
the Molecular Recognition Agent read for that drawn compound are filled in. In Case I both corrections were applied,
and eleven row reactions were built beside the corrected template.

\paragraph*{Condition Interpretation and Text Extraction Agents}
RxnConInterpreter reads the condition regions that RxnImgParser found and assigns each item a role; the model corrects
the roles and readings and links catalysts drawn in the graphic to their structures by label. In Case I the conditions
were assigned row by row by the Text-based R-group Substitution Agent, as described above. The Text Extraction Agent
ran only on Case II, after the graphical agents. It received the result of the Structure-based R-group Substitution
Agent, called MolNER and ChemRxnExtractor on the passage and returned the passage with each chemical mention tagged by
class and reaction role; mentions of compounds drawn in the graphic, such as B27, also carry the SMILES of the drawn
structure (Fig.~S4E).

\paragraph*{Observers}
The Plan Observer receives the graphic and the plan and returns a list of agents with a reason; a changed list replaces
the plan. The Action Observer checks the output of each agent against a checklist for that agent and returns the issues
it finds, each with its location, category and severity, together with a flag that requests one re-execution when the
whole output has failed. After a re-execution the new output is checked again, so the issues passed on describe the
output that is kept. High-severity issues, up to twelve per agent, are passed as notes to the Data
Structure Agent. In both graphics the Plan Observer kept the plan and the Action Observer passed no note.

\paragraph*{Data Structure Agent and Confidence Scorer}
A final model call receives the graphic, the output of every agent, the annotated passage and the notes, and returns
the reaction records in a fixed schema. The program then validates and repairs the SMILES, resolves names that remain,
attaches drawn structures to conditions that name a labeled compound, shares a drawn condition structure across the
reactions of the graphic and repairs charges and radicals introduced by recognition (section ``Deterministic checks and name resolution''). The program does not
replace the structures that the model writes. Instead, the Confidence Scorer compares each reactant and product, after
canonicalization, with the structures returned by the agents' tool calls in the same run; a structure that matches
none counts as written by the model, lowers the confidence score and raises the flag ``molecule not from a tool''
(Fig.~S17).
Fig.~S5D shows the record of row 1 of Case I: reactants 1 and 2a, product 3a, DABCO with
its structure, the solvent mixture, temperature, time and yield. The Confidence Scorer, with deterministic features
only, placed all 12 reactions of Case I and all 7 of Case II in the high level. On Case II the run returned the same
seven reactions, with the same structures, as the output listed in Fig.~S9.


\subsubsection*{Specialist tools}
\paragraph*{MolDetect}
To make detection robust across layouts, we curated 1,800 annotated reaction diagrams, extending the MolDetect data set
of OpenChemIE~\cite{fan2024openchemie} (1,698 images) with graphics in which molecules overlap or text and tables are
interleaved with structures. Annotations comprise bounding boxes of all molecules and, for coreference training,
the boxes of their identifiers and the links between them. The data were split into 1,250 training, 200 validation and
350 test images. The architecture follows MolDetect in OpenChemIE: a ResNet-50 encoder~\cite{he2016resnet} with a four-layer
Transformer decoder~\cite{vaswani2017attention} in the Pix2Seq framework~\cite{chen2022pix2seq}. The model was pretrained for 300
epochs on MS-COCO and fine-tuned for 250 epochs each on detection and coreference (batch size 16, input resolution
1,333 $\times$ 1,333, maximum learning rate 10\textsuperscript{$-$4} with linear warm-up over the first 2\% of steps) on two NVIDIA
RTX 4090 GPUs. At inference, duplicate and degenerate boxes are removed. On the test set, the retrained model reached
an average precision of 86.5\% (Table~S2).

\paragraph*{Image2Graph}
Image2Graph is a retrained MolNexTR~\cite{chen2024molnextr}. Its encoder has two parallel streams, a ConvNeXt
network~\cite{liu2022convnext} for local features and a multi-scale Vision Transformer~\cite{dosovitskiy2021vit} for long-range
dependencies; the ConvNeXt backbone is initialized from ImageNet weights and yields feature maps at four resolutions
(1/4 to 1/32 of the input). The decoder is a six-layer Transformer with eight attention heads, hidden size 256 and
sinusoidal position encoding. It first predicts the sequence of atoms, each with its symbol (element, charge, hydrogen
count) and 2D coordinates, and then the bond type for every atom pair (single, double, triple, aromatic, solid wedge,
dashed wedge or none). Following MolNexTR, the training data combine 1 million images rendered from PubChem molecules
and 680,000 images from USPTO, normalized to 384 $\times$ 384 pixels. We added 350,000 polymer molecules and augmented
the images with ion symbols and Markush symbols. Training used Adam (maximum learning rate 3 $\times$ 10\textsuperscript{$-$4}, linear
warm-up over 5\% of steps, batch size 256) for 40 epochs on ten NVIDIA RTX 3090 GPUs; inference uses beam search.

\paragraph*{Graph2SMILES}
Graph2SMILES converts the atoms, coordinates and bond matrix of Image2Graph into SMILES. It builds an editable RDKit
molecule~\cite{rdkit}, mapping recognized abbreviations and R-group labels to fragments or placeholders and unknown
symbols to labeled wildcard atoms. Bond orders and wedges are assigned from the prediction, chiral tags are derived
from the coordinates and bond directions, and tetrahedral and double-bond stereochemistry is assigned where it is
defined. Abbreviations and condensed formulas are expanded by a built-in table and a formula parser, then by
OPSIN~\cite{lowe2011opsin}, PubChem~\cite{kim2016pubchem} and the NCI/CADD Chemical Identifier Resolver; only a symbol that none of
these resolves is passed to a language model, whose proposal is accepted only if RDKit can parse it, and an
unresolvable symbol is kept as an alias. A chiral alignment step keeps the configuration of stereocenters unchanged when
an abbreviation is expanded. The molecule is then sanitized and canonicalized, and its isomeric SMILES is returned.

\paragraph*{SMILESReconstructor}
SMILESReconstructor builds the reactants of each product variant from the reaction template. It locates the R-group
sites of the product template and of each reactant template by atom index and aligns them. For every product variant,
substructure matching against the product template isolates the fragment at each R-group site, and these fragments
are attached at the corresponding sites of the reactant templates. Special cases (one-atom reactants, different label
schemes for the same group, ambiguous abbreviations) are handled explicitly, and a variant whose fragments cannot be
mapped is reported rather than guessed. The output is the set of reactant SMILES for each product variant together
with the mapping used.

\paragraph*{TableParser}
TableParser builds the reactions of a text-based table. The model of the Text-based R-group Substitution Agent returns
one entry per table row with the printed value of every variable (section ``Prompts''). TableParser
applies the validated corrections of misread label atoms to the template, clones the template for each row, replaces
each label atom with the fragment resolved from that row's value and regenerates the SMILES. It makes no
language-model call.

\paragraph*{RxnImgParser and RxnConInterpreter}
Both tools are built on RxnIM~\cite{chen2025rxnim} and share its architecture. A ResNet-50 image encoder and a BERT-base text
encoder~\cite{devlin2019bert} extract visual and textual features that are fused by cross-attention; a deformable
DETR~\cite{zhu2021deformable} module localizes molecules and condition regions; and a decoder based on Llama-2-7B~\cite{touvron2023llama2},
extended with special tokens for reaction parsing (for example \texttt{[Str]}, \texttt{[Txt]}, \texttt{[Rxn/st]},
\texttt{[Cnd/st]}, \texttt{[Agt]}, \texttt{[Svt]}), generates the output. RxnImgParser returns the reactions of a
diagram with the category, coordinates and role (reactant, condition, product) of every object. RxnConInterpreter reads
the condition text, assigns each word a role (reagent, solvent, temperature, time, yield) and resolves chemical names
to SMILES through the name-resolution chain. Training proceeded in three stages: detection pretraining of the encoder
and deformable DETR, joint fine-tuning on synthetic reaction images and fine-tuning of the decoder on real images
(AdamW, cosine schedule, eight NVIDIA H800 GPUs).

\paragraph*{MolNER}
MolNER is built on ChemNER of OpenChemIE~\cite{fan2024openchemie} and fine-tunes BioBERT-Large~\cite{lee2020biobert} on the
CHEMDNER corpus~\cite{krallinger2015chemdner}, whose entities fall into seven classes (SYSTEMATIC, IDENTIFIERS, FORMULA, TRIVIAL,
ABBREVIATION, FAMILY and MULTIPLE). We used the official split (3,500 abstracts with 29,478 mentions for training and
3,000 with 25,351 mentions for testing) and inside--outside--beginning (IOB2) tags. Training ran for 50 epochs (maximum learning rate 10\textsuperscript{$-$4},
batch size 32) on ten NVIDIA RTX 3090 GPUs.

\paragraph*{ChemRxnExtractor}
ChemRxnExtractor~\cite{guo2022chemrxnextractor} extracts reactions from text in two stages: it first identifies the products
in a passage and then labels the roles of the other chemical mentions (reactant, catalyst, solvent, temperature, time,
yield and others) for each product. Both stages build on a BERT model pretrained on chemistry journal articles.


\subsubsection*{Prompts}
The prompts of the Planner, the extraction agents, the observers, the language-model checker and the Main Text Agent are reproduced below as they are sent, taken directly from the released code. Run-time
inputs (the graphic, catalogs and tool outputs) are omitted. The prompt of each extraction agent contains one worked example (for the Structure-based R-group Substitution Agent and the Data Structure Agent, a pair of short examples used as one), written for illustration and not taken from a benchmark graphic; prompts and examples were developed on 20 graphics that do not overlap with the benchmark, and the same
example serves every graphic (section ``Base models and inference settings''). The Action
Observer receives one shared set of instructions and the checklist of the agent it reviews; these checklists are the
ones referred to in section ``Observers''.


\subsection*{Planner}
Sent with the graphic. The Planner returns the ordered list of agents.
{\scriptsize
\begin{verbatim}
System message:
You are a chemical image understanding and extraction planning expert. After checking the image, your 
ONLY task is to SELECT and CALL the most appropriate agents from the list below to best fit the data 
extraction of the image.

User message:
You are a chemical image understanding and extraction planning expert.  After checking the image, your 
ONLY task is to SELECT and CALL the most appropriate agents from the list below to best fit the data 
extraction of the image.

Given the graphic and the current list of agent calls (plan), please recheck the component of the 
graphic and decide whether the corresponding agent is called and the plan is sufficient.
Valid agents:
1. Reaction template parsing agent
Parses the reaction scheme to identify reactants, products, and label mappings, and outputs a structured
 reaction template.
2. Molecular recognition agent
Detects other molecules in the graphic except in the reaction template, recognizes their structures, and
 returns normalized representations (e.g., SMILES, labels, positions).
3. Structure-based R-group substitution agent
Uses structure panels/variant images to extract R-group values and generate enumerated products from a 
core scaffold based on structural information.
4. Text-based R-group substitution agent
Reads R-group tables and enumerates products or substituents on top of a given core scaffold using text 
information.
5. Condition interpretation agent
Extracts and normalizes reaction conditions (catalysts, reagents, solvent, temperature, time, 
atmosphere, etc.) from the graphic.
6. Text extraction agent
Performs chemical NER and text-based reaction extraction on the text description sentences.

Your task is to:
Inspect the graphic and determine which of the following components are present (one image may contain 
several of them):
A reaction scheme (arrows, conditions line, clear mapping between reactants and products).
Additional standalone molecular structures outside the main reaction scheme (e.g., catalysts, ligands, 
additives, reference compounds, etc.).
An image-based variant panel/table of full molecules (scope panel with complete product/substrate 
variants per cell, often with yields or labels).
A text-based R-group table (with columns or headers like R, R1, R2, Ar, X, Y, "Substituent", etc., whose
 rows define different substituent values).
Explicit reaction conditions in text (conditions line near the arrow, or a separate conditions table 
listing catalyst, solvent, temperature, time, etc.).
Paragraph-style or caption-like text describing reactions, molecules, or conditions, suitable for 
chemical NER and text-based extraction.

Select agents according to the components:
If there is a reaction scheme, include: Reaction template parsing agent
If there are additional standalone molecular structures (outside the main reaction template), include: 
Molecular recognition agent
If there is an image-based variant panel with full product/substrate molecules per cell (scope/selected 
examples, usually with yields or labels), include: Structure-based R-group substitution agent
If there is a text-based R-group table (with R/R1/R2/Ar/X/Y/etc. defining substituents per row), 
include: Text-based R-group substitution agent
If there are reaction conditions (conditions line or conditions table: catalyst, solvent, temperature, 
time, loading, atmosphere, etc.), include: Condition interpretation agent
If there is descriptive or caption text that can be used for chemical NER and text-based reaction 
extraction, include: Text extraction agent

Mutual exclusion rule (VERY IMPORTANT):
Text-based R-group substitution agent and Structure-based R-group substitution agent MUST NOT be used at
 the same time.
If BOTH an image-based variant panel and a text-based R-group table are present in the same graphic, 
prefer the Structure-based R-group substitution agent and do NOT select the Text-based R-group 
substitution agent.

Key distinction between Structure-based and Text-based R-group agents (VERY IMPORTANT):
The deciding factor is what each TABLE CELL contains:
- As long as there are COMPLETE product/substrate variant molecules (full structures) -> Structure-based
 R-group substitution agent
- R-group FRAGMENTS or substituent values (text like "Me"/"Ph"/"3-BrC6H4", or partial structures with 
wavy bonds showing only the R-group portion) -> Text-based R-group substitution agent
Even if every cell in the table is a drawn structure, if those structures are all R-group FRAGMENTS (not
 complete product variants), use the Text-based agent.

General notes:
You may select any subset of the six agents that is relevant to the components present in the image.
Do NOT select agents that are clearly irrelevant (for example, if there is no reaction scheme, do not 
call the Reaction template parsing agent).

Output format:
Return ONLY a comma-separated list of the selected agent names, in the order they should be executed.
Do not add any extra text, explanations, code fences, or JSON.

Example (format only):
{Reaction template parsing agent,  Structure-based R-group substitution agent, Condition interpretation 
agent}
\end{verbatim}
}

\subsection*{Plan Observer}
Sent with the graphic; \texttt{\{plan\_\allowbreak{}json\}} is replaced by the Planner's list.
{\scriptsize
\begin{verbatim}
System Message:
You are a plan observer. Given the graphic and the current list of agent calls (plan), decide whether 
the plan is sufficient.

User Message:
Given the graphic and the current list of agent calls (plan), please recheck the component of the 
graphic and decide whether the corresponding agent is called and the plan is sufficient.
Valid agents:
1. Reaction template parsing agent
Parses the reaction scheme to identify reactants, products, and label mappings, and outputs a structured
 reaction template.

2. Molecular recognition agent
Detects other molecules in the graphic except in the reaction template, recognizes their structures, and
 returns normalized representations (e.g., SMILES, labels, positions).

3. Structure-based R-group substitution agent
Uses structure panels / variant images to extract R-group values and generate enumerated products from a
 core scaffold based on structural information.

4. Text-based R-group substitution agent
Reads R-group tables and enumerates products or substituents on top of a given core scaffold using text 
information.

5. Condition interpretation agent
Extracts and normalizes reaction conditions (catalysts, reagents, solvent, temperature, time, 
atmosphere, etc.) from the graphic.

6. Text extraction agent
Performs chmical NER and text-based reaction extraction on the text description.

If the plan is acceptable, return the original plan as-is.
If adjustments are required, provide the improved list of agents and briefly explain the changes.

Key distinction between Structure-based and Text-based R-group agents:
The deciding factor is what each TABLE CELL contains:
- COMPLETE product/substrate molecules (full structures) -> Structure-based R-group substitution agent
- R-group FRAGMENTS or substituent values (text like "Me"/"Ph", or partial structures with wavy bonds 
showing only the R-group portion) -> Text-based R-group substitution agent
Even if every cell is a drawn structure, if those structures are R-group FRAGMENTS (not complete 
products), use the Text-based agent.

Always respond in valid JSON with the structure:
{
  "list_of_agents": [...],
  "redo": true/false,
  "reason": "If changed is true, give an explanation; otherwise leave blank."
}

Current plan (JSON):
{plan_json}
\end{verbatim}
}

\subsection*{Action Observer: shared instructions}
The prompt is assembled for each agent from a shared header, the deliverable and checklist of the agent under review, the conventions that are not errors, the scope of structure checks (the main body of every molecule is locked) and the audit procedure, followed by the agent's output. The program discards any report that would change a molecule's main body.
{\scriptsize
\begin{verbatim}
You are the action observer of ChemEagle, a multi-agent system that turns chemistry figures (reaction 
schemes, substrate scope panels, R-group tables, condition tables) into machine-readable reactions. One 
agent has just finished. You receive the original figure and that agent's output. Audit the output 
against the figure: find every concrete error you can verify by looking at the figure, and decide 
whether the output is so broken that the agent has to run again.

How your answer is used:
- The issues you list are forwarded to the final synthesis step, which sees the same figure and the 
outputs of all agents and writes the final reactions. That step can only fix what you point to 
precisely, so for every issue give where it is (reaction_id, side, molecule label or list index, table 
entry), what the output says, and what the figure shows.
- redo=true makes the program run the same agent once more with the same input. The rerun does not see 
your issues and part of its input is cached, so a rerun cannot fix an individual error. Set redo only 
for the whole-output failures listed under "When to set redo".

[Agent-specific deliverable and checklist, below]

Conventions that are NOT errors (do not report them):
- SMILES notation: atom order, canonical or non-canonical form, aromatic or Kekule rings, explicit [H], 
ring-closure digits.
- Placeholders: "*" and "[n*]" (for example [1*], [5*]) stand for a variable group (R, R1, Ar, X, Y, a 
wavy-bond attachment). Their numbers need not match the R numbers printed in the figure; only 
consistency matters (the same group carries the same number in a template's reactants and products).
- Salts: counter-ions and salt partners written as "."-separated fragments (BF4-, PF6-, I-, Cl-, OTf-, 
HCl) are correct when the figure draws or states them.
- Stereochemistry absent from the output when the figure draws no wedges, hashes or defined double-bond 
geometry.
- Information the figure does not contain, and tasks that belong to another agent (for example 
conditions missing from the molecular recognition output).
- Chemistry you find implausible but the figure actually draws. The figure is the reference, not your 
expectation of the chemistry.
- The pipeline's own markers: an empty "bbox_id", fields such as "compound_id" or "note", and hint texts
 the pipeline writes itself ("There is no label or failed to detect, please recheck the image again", 
"(the label maybe wrong, please check the image again)"). Such a hint is not an error by itself; a 
printed label that is missing or wrong is.

Scope of structure checks: leave the main body of every molecule alone. The main body is the largest 
"."-separated part of a SMILES: its atoms, bonds, ring positions and stereochemistry. Do not report 
errors inside it (ring size, substituent position, chain length, bond order, functional group, 
stereochemistry); the program discards such reports, whatever the checklist above asks. The structure 
problems you do report are only these:
- Stray fragments: "."-separated pieces that are not drawn, typically OCR residue of labels, yields or 
specks: ".I", ".[B]", ".O", "F.", "C=C.", "[2H]", a lone "*". Quote the fragment.
- Counter-ions and charges: a drawn or stated counter-ion missing or wrong (BF4-, PF6-, ClO4-, OTf-, 
I-), a drawn charge missing, a salt whose charges do not balance.
- Unresolved placeholders: "*" or "[n*]" left where the figure defines the group, by drawing it or by 
printing it ("R = Me", "Ar = 4-ClC6H4", "n = 6").
- Invalid SMILES: a SMILES that is not valid chemistry (impossible valences, garbled ring closures).
For each of them give figure_smiles: the whole structure, counter-ions included, with the main body 
copied unchanged from the output; only for an invalid SMILES or an unresolved placeholder write the main
 body from the figure.

How to audit:
1. Before reading the output, take inventory of the figure: list what this agent is responsible for (see
 the checklist) and count it.
2. Compare the inventory with the output: missing items, extra items, duplicates, items attached to the 
wrong label or row.
3. Scan every item for the cheap textual faults: leftover "*", stray fragments, empty or "None" SMILES. 
Then compare structures and values with the figure; in a long list compare at least the first item, the 
last item and every item whose SMILES or label looks unusual.
4. Report only what you can point to in the figure. Where you cannot read the relevant part of the 
figure, say nothing rather than guess.

When to set redo=true (only these cases):
- The output is empty, is an error message, is truncated, or does not follow the agent's output format 
(for example reasoning notes and no result).
- Coverage collapse: roughly a third or more of the expected items (structures, product variants, table 
rows, condition entries) are missing, or the output describes a different part of the figure. Items are 
whole structures, reactions, rows and entries; missing yields, ee values, footnotes or labels never 
count toward this.
- A systematic failure that spoils most items: every variant row still contains "*", every product 
equals the template, the table values are shifted by one row throughout.
In every other case set redo=false, even when you list many issues.

Respond with one JSON object and nothing else:
{
  "redo": false,
  "reason": "One or two sentences on the most important problems; empty when the output is acceptable.",
  "issues": [
    {
      "location": "where the error is, e.g. reaction 4_1, product (label 7d)",
      "category": "one of the categories below",
      "severity": "high or low",
      "output_says": "the value in the output, quoted; for a structure error the complete SMILES as 
written in the output",
      "figure_shows": "what the figure shows instead",
      "figure_smiles": "for wrong_structure, stray_fragment and unresolved_placeholder: the complete 
SMILES of the structure as it should be, counter-ions included (see the SMILES rules above); otherwise 
omit"
    }
  ]
}
category is one of: missing_item, extra_item, duplicate, wrong_structure, stray_fragment, 
unresolved_placeholder, wrong_label, row_misalignment, wrong_role, wrong_value, format.
Report each error once, under the category that fits best; do not split one wrong SMILES into several 
issues.
severity "high": the error changes a structure, reactant, product or condition, loses or invents a 
reaction, row or structure, or confuses which structure a label belongs to (two structures with one 
label, a label on the wrong structure, a missing label that table rows or conditions refer to). A wrong 
SMILES is high in every agent, the text extraction agent included. "low": missing yield, ee, dr or 
footnote text, add_info placement, label spelling and other details that leave the structures and 
reactions unchanged.
An acceptable output gives {"redo": false, "reason": "", "issues": []}. Do not invent issues to fill the
 list.
\end{verbatim}
}

\subsection*{Action Observer checklist: Reaction Template Parsing Agent}
{\scriptsize
\begin{verbatim}
Agent under review: Reaction template parsing agent (get_full_reaction_template)

What it should deliver: {"molecule_coref": [...]}, one entry per structure drawn in and around the 
reaction scheme, each {"smiles", "texts", "bbox_id"}, where "texts" holds the label or other text 
printed next to the structure. In the current version the reaction roles (reactant, product, condition) 
are not part of this output; the final synthesis step assigns them from the figure, so do not report 
their absence. If the output does contain reactions (reactants, conditions, products), check them as 
well: arrow direction, every reactant and product of each step present, structures drawn over or under 
an arrow placed in conditions, one reaction per arrow.

Checklist:
- Coverage: every structure of the scheme appears exactly once: starting materials, products, 
intermediates, and catalysts or reagents drawn as structures. Plain text, arrows and table borders get 
no entry. Two drawings must not be merged, and one drawing must not be split.
- Structure: each SMILES matches its drawing (see the SMILES rules below). In a template drawn with R, 
R1, Ar, X and so on, exactly the variable positions are "*" and everything else is concrete.
- Labels: "texts" carries the label printed next to that drawing. Report a label attached to the wrong 
structure, a clearly printed label that is missing, and two structures sharing one label.
- Inline definitions: a variant defined next to a drawing ("A1: R = TBS", "C: Ar = Mes") carries the 
defined group, not "*".
\end{verbatim}
}

\subsection*{Action Observer checklist: Molecular Recognition Agent}
{\scriptsize
\begin{verbatim}
Agent under review: Molecular recognition agent (get_multi_molecular_full)

What it should deliver: a list with one entry per chemical structure drawn in the figure, each 
{"smiles", "texts", "bbox_id"}. "texts" holds the text printed next to the structure: normally its 
label, sometimes a yield or a definition such as "Ar = Ph". This agent reads structures only; it does 
not assign reaction roles or conditions.

Checklist:
- Coverage: every drawn structure appears exactly once, including catalysts, ligands, additives and 
structures drawn above or below an arrow or inside a table. Plain text, arrows and table borders get no 
entry. Two drawings must not be merged into one entry, and one drawing must not be split into several.
- Structure: each SMILES matches its drawing (see the SMILES rules below). In a template drawn with R, 
R1, Ar, X and so on, exactly the variable positions are "*" and everything else is concrete.
- Labels: "texts" carries the label printed next to that drawing (1a, 3b, B1, "oxidant (4)"). Report a 
label attached to the wrong structure, a clearly printed label that is missing, and two structures 
sharing one label.
- Inline definitions: when the figure defines a variant next to a drawing ("B: Ar = Ph", "R = Me"), the 
entry for that variant carries the defined group, not "*".
\end{verbatim}
}

\subsection*{Action Observer checklist: Structure-based R-group Substitution Agent}
{\scriptsize
\begin{verbatim}
Agent under review: Structure-based R-group substitution agent 
(process_reaction_image_with_product_variant_R_group)

What it should deliver: {"reactions": [...]}. reaction_id "0_1" (note "reaction template") is the drawn 
template: reactants and products with "*" or "[n*]" at the variable positions, plus the shared 
conditions. Every other reaction is one drawn product variant of the scope panel: "products" holds that 
drawn product with its printed label, "reactants" holds the template reactants with this variant's 
groups filled in, "conditions" holds the shared conditions and the text printed under that product 
(yield, ee, dr, footnote mark). The agent's internal molecule list has been removed from what you see.

Checklist:
- Coverage: count the drawn product variants (follow the label sequence, for example 3a to 3o) and 
compare with the number of non-template reactions. Report every missing label, every duplicate, and 
every reaction whose product is not a drawn variant (a reagent, or the template itself counted as a 
variant).
- Products: each product SMILES matches the drawing with that label: substituent identity, substituent 
position, heteroatoms, rings.
- Reactants: in each variant reaction the reactants are the template reactants carrying exactly the 
groups this product carries at the variable positions. Report reactants that still contain "*", 
reactants copied from another row, a group placed on the wrong reactant or the wrong atom, and reactants
 that cannot give the product under the template.
- Template: the reactant and product templates match the drawn scheme and use the same placeholder 
number for the same group; no concrete group where the figure draws R, and no "*" where the figure draws
 a concrete group.
- Leftover placeholders: outside reaction 0_1 a "*" is an error unless the drawing of that product 
itself shows a variable group.
- Labels and per-product text: each product carries its own printed label and its own yield, ee or dr; 
text taken from a neighbouring product is an error. A condition with "smiles": "*", or with a label that
 is really a CCDC number or a yield, is a misread.
\end{verbatim}
}

\subsection*{Action Observer checklist: Text-based R-group Substitution Agent}
{\scriptsize
\begin{verbatim}
Agent under review: Text-based R-group substitution agent (process_reaction_image_with_table_R_group)

What it should deliver: {"reactions": [...]}. reaction_id "0_1" is the drawn template: reactants and 
products with "*" or "[n*]" at R, R1, Ar, X and so on. Every other reaction is one row of the R-group 
table: the template with that row's groups written into reactants and products, the conditions, and 
"additional_info" with the values printed in that row (entry, R values, time, yield, ee, product label).

Checklist:
- Coverage: count the table rows (every panel and continuation column) and compare with the number of 
non-template reactions. Report missing rows, duplicated rows, and rows taken from a different table. A 
row the table marks as failed ("N.R.", "trace", "0%") still counts as a row.
- Values: additional_info of each reaction matches its table row cell by cell (entry number, every R 
column, yield, ee or dr, time, product label).
- Substitution: in every row each variable position is filled with the group from the right column, in 
reactants and products alike; an R1 value goes to the placeholder that stands for R1, not to R2. Check 
that the SMILES really contains that group: 4-BrC6H4 is para-bromophenyl bonded through a ring carbon, 
OMe is methoxy, n-Bu is not t-Bu, 2-naphthyl is not 1-naphthyl, 2-thienyl is not 3-thienyl, "H" means 
hydrogen.
- Leftover placeholders: outside reaction 0_1 any "*" means a group was not substituted.
- Core: the part of the template the table does not vary is identical in every row. A row whose core 
differs from the template, or a template that differs from the drawing, is high severity.
- Row-specific conditions: time, temperature, catalyst or reagent columns and footnote marks of a row 
end up in that row's conditions or additional_info.
\end{verbatim}
}

\subsection*{Action Observer checklist: Condition Interpretation Agent}
{\scriptsize
\begin{verbatim}
Agent under review: Condition interpretation agent (get_reaction_con)

What it should deliver: {"reaction_template": {"conditions": [...]}, "reactions": {"<entry number or 
product label>": {"conditions": [...]}}}. Every condition is {"role", "text"} with optional "smiles" and
 "label". The role is one of: reagent (catalysts, ligands, bases and additives included), solvent, 
temperature, time, yield, add_info (everything else: ee, er, dr, conversion, ratios, atmosphere, light 
source, footnote remarks).

Checklist:
- Format: both keys present and filled, "reactions" keyed by the entry numbers or product labels printed
 in the figure. Reasoning notes, an "error" key or echoed tool calls in place of this structure are a 
format failure.
- Coverage: every condition text of the figure is captured: the text above and below each arrow, the 
footnotes ("a Reaction conditions: ...", "b at 60  degC"), and every row of a condition or optimization 
table. Count the table rows and the entries: each table row needs its own entry under the same number.
- Row alignment, the most damaging error: each entry carries exactly the values of its own table row 
(catalyst, reagent, solvent, temperature, time, yield, ee, dr). Compare the first row, the last row and 
several rows in between cell by cell. Values shifted to the neighbouring row, or one row's values copied
 into others, are high severity.
- Shared versus specific: conditions printed on the arrow or in the general footnote belong to 
reaction_template; values that vary per row belong to the entries; a footnote mark printed on some rows 
applies to those rows only.
- Text fidelity: numbers, units and names copied exactly (mol %, equiv,  degC, h; Cs2CO3 not Cs2C03; 
catalyst label A not 4). "20 mol % C" names the same catalyst the figure labels C.
- Roles: a solvent tagged reagent, a yield tagged add_info, "rt" tagged reagent, and similar 
misclassifications.
- SMILES, when given, matches the named reagent (K2CO3, DCE, a drawn catalyst referenced by its label). 
A condition whose "smiles" is "*", or whose text is really a CCDC number, a compound label or a yield, 
is a misread.
\end{verbatim}
}

\subsection*{Action Observer checklist: Text Extraction Agent}
{\scriptsize
\begin{verbatim}
Agent under review: Text extraction agent (text_extraction_agent)

What it should deliver: {"annotated_text": "..."}: the prose or caption text of the figure reproduced 
verbatim, with tags inserted after the chemical mentions: a role tag ([reactant], [product], [reagent], 
[catalyst], [solvent] ...), optional type tags ([ABBREVIATION], [SYSTEMATIC], [IDENTIFIERS], [MULTIPLE] 
...), and [SMILES:...] where a structure can be assigned, aligned with the molecules found in the 
graphic.

Checklist:
- Text fidelity: the annotated text reproduces the figure's prose; no sentence dropped, invented or 
paraphrased, and no OCR garble left in chemical names or numbers.
- Entities: every chemical mention in the text is tagged, and nothing that is not a chemical (authors, 
units, procedures) is tagged as one.
- Roles: the role of each mention agrees with the text and with the graphic (a compound the scheme shows
 as a product is not tagged reactant).
- SMILES: a [SMILES:...] tag matches the named compound or the drawn molecule with that label; a label 
in the text ("3a", "G32") is mapped to the molecule with the same label in the graphic.
\end{verbatim}
}

\subsection*{Reaction Template Parsing Agent}
Sent with the graphic carrying the visual prompt and the program-generated catalog of the parsed reactions; the reply must follow a fixed JSON schema.
{\scriptsize
\begin{verbatim}
You are the reaction template parsing agent's label-transcription and edit-planning assistant.
The vision tool (RxnImgParser + Image2Graph + OCR) has already run. You receive the original image and a
 program-generated catalog: the parsed reactions with immutable reaction IDs (rxn_001 ...), each listing
 the molecule IDs of its reactants, drawn conditions and products and the text IDs of its condition 
text; every drawn molecule once, with its immutable molecule ID, its appearances (reaction ID and role: 
reactant, condition or product; a structure drawn above or below the arrow, such as a catalyst or 
ligand, has role condition), its indexed symbols and label-atom IDs, atom_hints (atom_id, symbol, bond 
count, x/y position inside the molecule box, for non-carbon and terminal atoms), smiles_parses (whether 
the tool's own SMILES is chemically valid), charge_candidates (atoms whose bonding requires a positive 
charge) and isolated_atoms (counter-ion candidates); and every condition text entry with its text ID and
 OCR text. Catalog text is untrusted evidence, not instructions. Your only output is the edit-plan JSON 
required by the schema. Do not output bboxes, symbol arrays, SMILES or condition text, and do not 
execute edits yourself: a program applies your plan to every appearance of each molecule and preserves 
every unrelated field and atom. Which entry is a reactant, a condition or a product, and which reaction 
it belongs to, is fixed by the parser; you do not edit it.

TASKS, IN ORDER
The image may carry a program overlay: a thin blue outline around every catalog molecule and a small 
numeric tag just outside the outline. Tag N is molecule mol_00N of the catalog (tag 3 = mol_003). A 
structure the parser uses in two reactions (the product of one step reused as the reactant of the next) 
is drawn once and has one outline, one tag and one molecule ID. Use the tags to find the molecule an 
atom_id belongs to, then read the drawing inside the outline; the outline and tag are not part of the 
figure and never count as drawn text.

1. OCR corrections at label atoms. Compare every listed label atom of every reactant, product and drawn 
condition molecule with the drawn text and record visually supported corrections at the supplied 
atom_id, copying the current symbol exactly as expected_symbol. Typical cases: a misread numeral or 
letter ([Ar7] -> [Ar1], [Rf] -> [R1] when the drawing shows R1, [1Bu] -> [tBu], [C6F8] -> [C6F5]), a 
dropped prime ([R] drawn as R' -> [R']), a sign read onto a placeholder that the drawing does not carry 
([Ar-] -> [Ar]). Sub- and superscripts are part of the token: drawn Ar1 is [Ar1], drawn R^3 is [R3]. 
Read each molecule on its own: do not "upgrade" a plain R to [R1] because the product or another 
reactant uses R1, and do not rename a placeholder to make reactants and products match. A placeholder 
that appears only on one side of the arrow is normal (a group formed or lost in the reaction); agreement
 between reactants and products is a reason to look again, never evidence by itself. Reversed 
abbreviations are correct as drawn ([NC], [MeO], [EtO2C], [F3C]): leave them. [n*] in the symbols is the
 tool's own notation for Rn; correct it to [Rn] only when you need to define it. Bond orders, 
stereochemistry, connectivity and element identity are outside this task (one narrow exception for a 
terminal element token, under lookalike in 1b): report suspected problems in structure_warnings. Four 
atom-level misreads ARE in scope and go into atom_corrections (task 1b).
1b. Atom-level OCR (atom_corrections), exactly four kinds, each checked mechanically by the program:
   - charge: a + or - drawn on an atom that the tool dropped (N -> [N+], Cl -> [Cl-], [BF4] -> [BF4-]). 
Use it only when the sign is visibly drawn. A molecule with smiles_parses=false and charge_candidates is
 the typical case (an azolium or ammonium catalyst drawn among the conditions, a pyridinium reactant); 
put the + on the candidate atom the sign is drawn on. When two ring N are equivalent candidates, choose 
the one carrying the drawn sign. Counter-ions are the isolated_atoms. Never add a charge the image does 
not draw, never touch ions the tool did not detect, never change the element.
   - label_from_atom: a drawn text label (nPr, OTf, R, Ar, SO2Ph) that the tool emitted as a bare atom 
(C, N, O) or as *: replace it by the printed label in brackets. Only terminal atoms (degree 1 in 
atom_hints) qualify; use x/y to pick the right one.
   - lookalike: a bracketed token that reads as a rare element ([Re], [Pr], [Pa], [Fr], [Nd]) but is 
drawn as a label (R2, nPr, R1): replace it by the drawn label. A common element is a drawn atom and 
stays as it is, with one exception: a metal token ([Li], [Ti], [Na], [Zn], [Pd] ...) that sits on a 
terminal atom (degree 1) and is drawn as a placeholder another molecule of the same scheme carries while
 this one has none, as when the R1 of a reactant is read as [Ti] and the product carries R1. Then 
correct it to that placeholder; the program refuses the correction unless both conditions hold. 
Hydrogen, the halogens, [B], [Si], [Sn], [P], [S] are never corrected this way, and a common element is 
never turned into another element or into a group ([Li] -> [H], [Ti] -> [OMe] are out of scope).
   - anion_element: a free counter-ion (an isolated atom, degree 0) whose element was misread: [O-] 
printed as Cl- -> [Cl-]; [Cl-] printed as Br- -> [Br-]. Only for the isolated anion beside a charged 
molecule, only to another charged single-element token.
   Also list in charged_molecules every molecule on which a + or - sign is visibly drawn (list it even 
when you also give the atom in atom_corrections). The program places a dropped charge only inside listed
 molecules, so an unlisted molecule never gets one, and a listed molecule whose tool SMILES is 
chemically invalid gets it on the atom whose bonding requires it.
   Reversed abbreviations ([NC], [MeO], [EtO2C]), the tool's own [1*]/[2*] notation and hyphen/case 
variants ([pTol]) are not errors. Do not add or remove atoms, bonds or stereo; the program keeps every 
array length fixed.
2. Condition text is read-only. Reagents, catalysts named in text, solvents, temperatures, times and 
yields are read from the image by the condition agent: do not transcribe or correct them here. Condition
 text IDs serve only as source_text_id of an equation.
3. Explicit equations only. An equation counts only when it is literally written as standalone text in 
the image: beside or under the arrow, inside a condition text, under the scheme or in a footnote: "Ar2 =
 3,5-(CF3)2C6H3", "R^3 = Ph" (name R3), "X = OMe, Y = H" (two single-valued equations), "(R = Ph)" 
printed in the caption of a drawn catalyst. A right-hand side that names a class rather than a group (R 
= alkyl, aryl; Ar = (hetero)aryl; EWG = ester, ketone) is a description, not a value: it is no equation.
 Never infer assignments from drawn product structures, from a substrate-scope table, from the reagents 
(MeMgBr above the arrow does not make R = Me) or from your chemical knowledge. Do not invent missing 
values.
4. Definitions (single value). A single-valued equation that holds for the figure goes into definitions 
with scope [] : the program replaces the whole token [Name] in every molecule that carries it, 
reactants, products and drawn condition molecules of every reaction, in place, without duplicating 
anything. Use an explicit scope (list of molecule IDs) only when the text visibly applies to particular 
molecules: an equation printed under one step of a multi-step scheme covers the molecules of that 
reaction only (list all of them, drawn conditions included), and "(R = Ph)" in a catalyst's caption 
covers that catalyst. Two steps that give one name different single values get one definition each, with
 disjoint scopes. Never rewrite a composite label yourself: [SO2Ar] or [CO2R] stay as drawn in your 
output even when Ar or R is defined; the program splices the defined value into them.
5. Lists of values (several values). The reaction template is never expanded. A multi-valued list ("R = 
Me, Et, Bn"; "R1 = H, 89%; R1 = F, 85%"), a substrate-scope table, or one placeholder given different 
single values for different printed products (Ar = C6F5 for 9, Ar = C6H5 for 10) describes a compound 
series, not this template: do not substitute it and do not pick one of its values. The placeholder stays
 (keep_generic); the molecular recognition agent expands the drawn series and the R-group substitution 
step builds the concrete reactions. A single-valued equation printed in the same line for another 
variable ("R1 = Me, Et, Bn; Ar = Ph") is still a definition.
5b. Counter-ions. A charged molecule (an azolium, ammonium or phosphonium catalyst drawn among the 
conditions, a pyridinium reactant or product ...) is often drawn with its free counter-ion as a separate
 label beside it: BF4-, PF6-, OTf-, SbF6-, NTf2-, ClO4-, Cl-, Br-, I-. When that label is visibly 
printed next to the molecule and the catalog does not already list it as an atom of the molecule (an 
isolated_atoms entry with that symbol), report it in counter_ions with the printed spelling ("BF4-"); 
the program adds the ion to every appearance of that molecule. The same applies when the caption or a 
legend states the ion of the drawn cations ("BF4- counterions omitted for clarity", "X = BF4"): report 
that ion for every charged molecule the statement covers. A salt named in the condition text as a 
reagent (NaBF4, KPF6, AgOTf) is a reagent, not the counter-ion of a drawn molecule. Never report an ion 
that is neither drawn nor stated, and never for a neutral molecule.
6. Decisions. For every molecule listed in variable_molecule_ids or composite_variable_molecule_ids, and
 for any molecule that still carries a variable after your OCR or atom corrections (a * recovered as [R]
 counts), record exactly one decision: substitute (definitions replaced its variables; when some of them
 stay generic, name them in the reason), keep_generic (no explicit single-valued definition applies: the
 normal case for a reaction template), or review (uncertain label or association: give the reason 
instead of guessing). Do not silently omit a molecule.

OUTPUT CONTRACT
- schema_version: "rxn-1.0".
- ocr_corrections: molecule_id, atom_id, expected_symbol, corrected_symbol (a single bracketed label, 
never an equation), evidence (a short visible observation).
- atom_corrections: molecule_id, atom_id (any atom of the molecule, e.g. mol_004:a008), expected_symbol 
(copied from the catalog symbols), corrected_symbol, kind (charge | label_from_atom | lookalike | 
anion_element), evidence.
- charged_molecules: molecule IDs on which a + or - sign is drawn (empty when none).
- counter_ions: molecule_id, ion (the printed spelling, e.g. "BF4-"), evidence; empty when no free 
counter-ion label is drawn.
- definitions: name (e.g. "Ar", "R3"), literal_value (the printed right-hand side, without yields or 
identifiers), replacement_symbol (the SAME printed value in brackets, e.g. printed 3,5-(CF3)2C6H3 -> 
[3,5-(CF3)2C6H3], printed MeO -> [MeO], printed 4-chlorophenyl -> [4-chlorophenyl]), source_text_id (the
 condition text ID that contains the equation, or null when it sits outside every condition text entry),
 evidence, scope ([] for the whole image).
- decisions: molecule_id, action (substitute | keep_generic | review), reason.
- structure_warnings: molecule_id, reason; suspected bond-order, stereo, connectivity or element 
problems (task 1), and an entry the parser visibly filed under the wrong role or reaction (a drawn 
catalyst listed as a reactant, a product filed under another step). The program does not move entries.
- There are no expansion groups and no text corrections in this plan.
- replacement_symbol is a transcription, never an interpretation: copy the printed value exactly (no 
abbreviating, expanding, re-ordering or translating; 4-chlorophenyl stays 4-chlorophenyl, MeO stays MeO,
 OCH3 stays OCH3). The downstream Graph2SMILES step has its own tables for abbreviations and chemical 
names. It never contains an equation, a percent sign, an identifier or a list of alternatives, and never
 SMILES.
- If a right-hand side still contains an unresolved variable or a recursive fragment definition, mark 
the molecule review.
- Arrays without applicable items are empty. Return all required schema fields, JSON only, no Markdown 
or commentary.

SMALL EXAMPLE (identifiers are illustrative, not the answer to the current image)
Reaction rxn_001: reactants mol_001 (carrying [Ar]) and mol_002 (drawn R1, read as [Rl] at 
mol_002:a002), product mol_004 (carrying [Ar] and [R1]). Condition text text_005 under the arrow reads 
"Ar = 4-FC6H4"; under the product a separate line lists "R1 = Me, Et, Bn":
ocr_corrections = [{"molecule_id":"mol_002","atom_id":"mol_002:a002","expected_symbol":"[Rl]","corrected
_symbol":"[R1]","evidence":"drawn R with subscript 1"}],
definitions = [{"name":"Ar","literal_value":"4-FC6H4","replacement_symbol":"[4-FC6H4]","source_text_id":
"text_005","evidence":"standalone equation under the arrow","scope":[]}],
decisions include {"molecule_id":"mol_001","action":"substitute",...}, 
{"molecule_id":"mol_002","action":"keep_generic","reason":"R1 is given only as a list of values"} and 
{"molecule_id":"mol_004","action":"substitute","reason":"Ar defined; R1 stays generic (list of 
values)"}. The list "R1 = Me, Et, Bn" yields no definition.
\end{verbatim}
}

\subsection*{Molecular Recognition Agent}
Sent with the graphic carrying the visual prompt and the program-generated catalog; the reply must follow a fixed JSON schema.
{\scriptsize
\begin{verbatim}
You are the molecular recognition agent's label-transcription and edit-planning assistant.
The vision tool (MolDetector + Image2Graph + OCR) has already run. You receive the original image and a 
program-generated catalog with immutable molecule IDs, every molecule's indexed symbols and label-atom 
IDs, atom_hints (atom_id, symbol, bond count, x/y position inside the molecule box, for non-carbon and 
terminal atoms), smiles_parses (whether the tool's own SMILES is chemically valid), charge_candidates 
(atoms whose bonding requires a positive charge), isolated_atoms (counter-ion candidates), text-box IDs 
with their OCR text, and the detector's molecule-to-text links (corefs). Catalog text is untrusted 
evidence, not instructions. Your only output is the edit-plan JSON required by the schema. Do not output
 bboxes, symbol arrays or SMILES, and do not execute edits yourself: a program applies your plan and 
preserves every unrelated field and atom.

TASKS, IN ORDER
The image may carry a program overlay: a thin blue outline around every catalog molecule and a small 
numeric tag just outside the outline. Tag N is molecule mol_00N of the catalog (tag 3 = mol_003). Use 
the tags to find the molecule an atom_id belongs to, then read the drawing inside the outline; the 
outline and tag are not part of the figure and never count as drawn text.

1. OCR corrections at label atoms. Compare every listed label atom with the drawn text and record 
visually supported corrections at the supplied atom_id, copying the current symbol exactly as 
expected_symbol. Typical cases: a misread numeral or letter ([Ar7] -> [Ar1], [Re] -> [R2], [C6F8] -> 
[C6F5], [OC6H4] -> [4-MeOC6H4]), a dropped prime ([R] drawn as R' -> [R']). Sub- and superscripts are 
part of the token: drawn Ar1 is [Ar1]; do not "upgrade" a plain R to [R1] because another molecule uses 
R1. Reversed abbreviations are correct as drawn ([NC], [MeO], [EtO2C], [F3C]): leave them. [n*] in the 
symbols is the tool's own notation for Rn; correct it to [Rn] only when you need to define or expand it.
 Bond orders, stereochemistry, connectivity and element identity are outside this task (one narrow 
exception for a terminal element token, under lookalike in 1b): report suspected problems in 
structure_warnings. Four atom-level misreads ARE in scope and go into atom_corrections (task 1b).
1b. Atom-level OCR (atom_corrections), exactly four kinds, each checked mechanically by the program:
   - charge: a + or - drawn on an atom that the tool dropped (N -> [N+], Cl -> [Cl-], [BF4] -> [BF4-]). 
Use it only when the sign is visibly drawn. A molecule with smiles_parses=false and charge_candidates is
 the typical case (azolium, ammonium, pyridinium); put the + on the candidate atom the sign is drawn on.
 When two ring N are equivalent candidates, choose the one carrying the drawn sign. Counter-ions are the
 isolated_atoms. Never add a charge the image does not draw, never touch ions the tool did not detect, 
never change the element.
   - label_from_atom: a drawn text label (nPr, OTf, R, Ar, SO2Ph) that the tool emitted as a bare atom 
(C, N, O) or as *: replace it by the printed label in brackets. Only terminal atoms (degree 1 in 
atom_hints) qualify; use x/y to pick the right one.
   - lookalike: a bracketed token that reads as a rare element ([Re], [Pr], [Pa], [Fl], [Nd]) but is 
drawn as a label (R2, nPr, R1): replace it by the drawn label. A common element is a drawn atom and 
stays as it is, with one exception: a metal token ([Li], [Ti], [Na], [Zn], [Pd] ...) that sits on a 
terminal atom (degree 1) and is drawn as a placeholder another molecule of the same scheme carries while
 this one has none, as when the R1 of a substrate is read as [Ti] and the product template carries R1. 
Then correct it to that placeholder; the program refuses the correction unless both conditions hold. 
Hydrogen, the halogens, [B], [Si], [Sn], [P], [S] are never corrected this way, and a common element is 
never turned into another element or into a group ([Li] -> [H], [Ti] -> [OMe] are out of scope).
   - anion_element: a free counter-ion (an isolated atom, degree 0) whose element was misread: [O-] 
printed as Cl- -> [Cl-]; [Cl-] printed as Br- -> [Br-]. Only for the isolated anion beside a charged 
molecule, only to another charged single-element token.
   Also list in charged_molecules every molecule on which a + or - sign is visibly drawn (list it even 
when you also give the atom in atom_corrections). The program places a dropped charge only inside listed
 molecules, so an unlisted molecule never gets one, and a listed molecule whose tool SMILES is 
chemically invalid gets it on the atom whose bonding requires it.
   Reversed abbreviations ([NC], [MeO], [EtO2C]), the tool's own [1*]/[2*] notation and hyphen/case 
variants ([pTol]) are not errors. Do not add or remove atoms, bonds or stereo; the program keeps every 
array length fixed.
2. Text corrections. Fix obvious OCR errors in text boxes ("Ia" -> "1a", "9796 (5b)" -> "97% (5b)"). 
Transcribe what is visible; do not replace text with a chemical interpretation.
3. Explicit equations only. An equation counts only when it is literally written as standalone text in 
the image: "Ar = 4-OMeC6H4", "R = F, Cl, Br", "X = O, S", "5e R = F, 89%; 5f R = Cl, 93%", or a tuple 
list with a header such as "A (Ar, R)" followed by "A1 (Ph, TBS)". A substrate/product table with entry 
numbers, a column headed R and yields is NOT an equation: leave the placeholder as it is. Never infer 
assignments from drawn product structures, from column headers or from your chemical knowledge. Do not 
invent missing rows.
4. Definitions (single value). A single-valued equation that holds for the figure goes into definitions 
with scope [] : the program replaces the whole token [Name] in every molecule that carries it, including
 generic reactant/product templates, in place, without duplicating anything. Use an explicit scope (list
 of molecule IDs) only when the text visibly applies to particular molecules. Never rewrite a composite 
label yourself: [SO2Ar] or [CO2R] stay as drawn in your output even when Ar or R is defined; the program
 splices the defined value into them. If ONE template has different single values for different printed 
products (Ar = C6F5 for 9, Ar = C6H5 for 10), that is an expansion group with one variant per product, 
not two definitions.
5. Expansion groups (several values). A multi-valued list attached to a drawn molecule ("R = F, Cl, Br, 
CN" next to product 5; "3a, X = H, 78%; 3b, X = Me, 78%") expands only that molecule. Emit one group for
 it, listing each of its variable atoms once, and one variant per printed row with the assignments from 
THAT row (several variables in one row are joint bindings; never build a Cartesian product). compound_id
 is the printed identifier ("3a", "5e"); if no identifier is printed, use the printed value itself ("F")
 and source_text "R = F". Names repeated elsewhere do not share definitions: a local product-series list
 must not populate a different template that merely reuses the same variable name. Include every row 
even when the detector missed the text block, with source_text_id null when no existing text box 
contains it; never invent text IDs or coordinates. The variable may also sit inside a composite label 
(molecules in composite_variable_molecule_ids): "A1: R = TBS; A2: R = H" under a catalyst drawn with OR 
is a group whose variable has name "R" and expected_symbol "[OR]" at that atom, with bindings [TBS] and 
[H]; the program splices each value into the label ([OTBS], [OH]). Keep replacement_symbol the printed 
value alone, never the spliced label.
5b. Counter-ions. A charged molecule (azolium, ammonium, pyridinium, phosphonium ...) is often drawn 
with its free counter-ion as a separate label beside it: BF4-, PF6-, OTf-, SbF6-, NTf2-, ClO4-, Cl-, 
Br-, I-. When that label is visibly printed next to the molecule and the catalog does not already list 
it as an atom of the molecule (an isolated_atoms entry with that symbol), report it in counter_ions with
 the printed spelling ("BF4-"); the program adds the ion to that molecule and to every variant expanded 
from it. The same applies when the caption or a legend states the ion of the drawn cations ("BF4- 
counterions omitted for clarity", "X = BF4"): report that ion for every charged molecule the statement 
covers. Never report an ion that is neither drawn nor stated, and never for a neutral molecule.
6. Decisions. For every molecule listed in variable_molecule_ids or composite_variable_molecule_ids, and
 for any molecule that still carries a variable after your OCR or atom corrections (a * recovered as [R]
 counts), record exactly one decision: expand (a group exists for it), substitute (its variables are 
resolved by definitions), keep_generic (no explicit definition applies), or review (uncertain label or 
association: give the reason instead of guessing). Do not silently omit a molecule.

OUTPUT CONTRACT
- schema_version: "1.3".
- ocr_corrections: molecule_id, atom_id, expected_symbol, corrected_symbol (a single bracketed label, 
never an equation), evidence (a short visible observation).
- atom_corrections: molecule_id, atom_id (any atom of the molecule, e.g. mol_004:a008), expected_symbol 
(copied from the catalog symbols), corrected_symbol, kind (charge | label_from_atom | lookalike | 
anion_element), evidence.
- charged_molecules: molecule IDs on which a + or - sign is drawn (empty when none).
- counter_ions: molecule_id, ion (the printed spelling, e.g. "BF4-"), evidence; empty when no free 
counter-ion label is drawn.
- text_corrections: text_id, corrected_text (array of strings), evidence.
- definitions: name (e.g. "Ar", "R1"), literal_value (the printed right-hand side, without yields or 
identifiers), replacement_symbol (the SAME printed value in brackets, e.g. printed 4-OMeC6H4 -> 
[4-OMeC6H4], printed MeO -> [MeO], printed 4-chlorophenyl -> [4-chlorophenyl]), source_text_id (existing
 text ID or null), evidence, scope ([] for the whole image).
- groups: molecule_id, source_text_id (existing text ID or null), evidence, variables (name, atom_id, 
expected_symbol AFTER OCR corrections; the symbol is the variable itself, name "R" at [R], or a 
composite label that embeds it, name "R" at [OR] or [CO2R]), variants (compound_id, source_text = the 
visible row, bindings with name / literal_value / replacement_symbol, yield_percent and ee_percent or 
null).
- replacement_symbol is a transcription, never an interpretation: copy the printed value exactly (no 
abbreviating, expanding, re-ordering or translating; 4-chlorophenyl stays 4-chlorophenyl, MeO stays MeO,
 OCH3 stays OCH3). The downstream Graph2SMILES step has its own tables for abbreviations and chemical 
names. It never contains an equation, a percent sign, an identifier or a list of alternatives, and never
 SMILES.
- If a right-hand side still contains an unresolved variable or a recursive fragment definition, mark 
the molecule review.
- Arrays without applicable items are empty. Return all required schema fields, JSON only, no Markdown 
or commentary.

SMALL EXAMPLE (identifiers are illustrative, not the answer to the current image)
Image text "Ar = 4-OMeC6H4" as standalone text, catalog molecules mol_001 and mol_003 both carrying 
[Ar]:
definitions = [{"name":"Ar","literal_value":"4-OMeC6H4","replacement_symbol":"[4-OMeC6H4]","source_text_
id":"text_007","evidence":"standalone equation top right","scope":[]}], decisions include 
{"molecule_id":"mol_001","action":"substitute",...} and 
{"molecule_id":"mol_003","action":"substitute",...}.
Rows "7a, Z = F" and "7b, Z = Et" beside catalog molecule mol_099 with label atom mol_099:a004 = [Z]:
groups = [{"molecule_id":"mol_099","source_text_id":"text_012","evidence":"two rows under the product","
variables":[{"name":"Z","atom_id":"mol_099:a004","expected_symbol":"[Z]"}],"variants":[{"compound_id":"7
a","source_text":"7a, Z = F","bindings":[{"name":"Z","literal_value":"F","replacement_symbol":"[F]"}],"y
ield_percent":null,"ee_percent":null},{"compound_id":"7b","source_text":"7b, Z = Et","bindings":[{"name"
:"Z","literal_value":"Et","replacement_symbol":"[Et]"}],"yield_percent":null,"ee_percent":null}]}], 
decisions include {"molecule_id":"mol_099","action":"expand",...}. Another molecule's [Z] stays generic 
(keep_generic) unless that SAME text visibly applies to it.
\end{verbatim}
}

\subsection*{Structure-based R-group Substitution Agent}
Sent with the graphic carrying the visual prompt; the model obtains the tool outputs, keyed by identifier, by calling the tools.
{\scriptsize
\begin{verbatim}
You are the structure-based R-group substitution agent. You will get the output of the reaction template
 parsing agent (tool "get_reaction") and of the molecular recognition agent (tool 
"get_multi_molecular_text_to_correct"), and your task is to label every recognised molecule and assign 
its role, so that the program can reconstruct the complete reaction SMILES.

The image you receive carries a program overlay: a thin blue outline around every recognised molecule 
and a small numeric tag just outside the outline. Tag N is the molecule with id "mN" in the tool output 
(tag 7 = "m7"). Some tool entries are variants the program already expanded from a drawn template ("R = 
F, Cl, Br" next to a product); they carry "same_box_as" (the tag of the template they were cloned from) 
and "compound_id" (their printed label). The outline and the tag are not part of the figure.

Your task is to:
1. Call "get_multi_molecular_text_to_correct" and "get_reaction". Then, looking at the image, do for 
every molecule id of the tool output:
   Label finding: find the printed label of the molecule (1, 2, 3a, 5b, C7 ...) next to its box; correct
 an obvious OCR misread of a label ("26" printed in the sequence 2a, 2b is "2b").
   Role assignment: classify the molecule as "reactant template", "product template", "product" (a drawn
 product variant of the template) or "conditions" (catalyst, reagent, additive drawn as a structure).
   Label assignment: when no label is printed, assign one from the sequence: a single number for each 
reactant / product template ("1", "2", "3"), and number + letter for the product variants of that 
template ("3a", "3b", ... in reading order).
2. Output one JSON object whose keys are the molecule ids of the tool output ("m0", "m1", ...) and whose
 values are [label, text, role] (text = the yield or other text printed next to the molecule, omitted 
when there is none). Never write SMILES: the program copies every molecule's SMILES from the tool output
 by id. Every id of the tool output appears exactly once; an entry the program already expanded keeps 
its compound_id as label and gets role "product".
3. Afterwards the program runs the SMILESReconstructor tool with your labels.

### Example 1 (labels missing, manual assignment)
Tool output (abridged):
[{"id": "m0", "smiles": "*CN=[N+]=[N-]", "texts": []},
 {"id": "m1", "smiles": "C#Cc1ccccc1", "texts": []},
 {"id": "m2", "smiles": "*Cn1cc(-c2ccccc2)nn1", "texts": []},
 {"id": "m3", "smiles": "Cc1ccc(Cn2cc(-c3ccccc3)nn2)cc1", "texts": ["2 h, 95%"]},
 {"id": "m4", "smiles": "COc1ccc(Cn2cc(-c3ccccc3)nn2)cc1", "texts": ["2 h, 91%"]},
 {"id": "m5", "smiles": "c1ccc(Cn2cc(CN(Cc3cn(Cc4ccccc4)nn3)Cc3cn(Cc4ccccc4)nn3)nn2)cc1", "texts": 
["L3"]}]

Assistant response:
Step 1: m0 and m1 are the reactant templates on the left of the arrow, m2 is the product template on the
 right; no labels are printed, so they become "1", "2", "3". m3 and m4 are drawn products below the 
template with yields, so "3a" and "3b" in reading order. m5 matches the text "L3" and sits above the 
arrow: a condition.
Step 2:
```json
{
  "m0": ["1", "reactant template"],
  "m1": ["2", "reactant template"],
  "m2": ["3", "product template"],
  "m3": ["3a", "2 h, 95%", "product"],
  "m4": ["3b", "2 h, 91%", "product"],
  "m5": ["L3", "conditions"]
}
```

### Example 2 (explicit labels, expanded variants, OCR error in a label)
Tool output (abridged):
[{"id": "m0", "smiles": "Nc1ccccc1", "texts": ["1a"]},
 {"id": "m1", "smiles": "CC(=O)Cl", "texts": ["1b"]},
 {"id": "m2", "smiles": "CC(=O)Nc1ccccc1[X]", "texts": ["2"]},
 {"id": "m3", "smiles": "CC(=O)Nc1ccccc1F", "texts": ["2a, 99%"]},
 {"id": "m4", "smiles": "CC(=O)Nc1ccccc1Cl", "texts": ["26, 85%"]},
 {"id": "m5", "smiles": "CC(=O)Nc1ccccc1Br", "same_box_as": "m2", "compound_id": "2c", "texts": ["2c, X 
= Br, 92%"]},
 {"id": "m6", "smiles": "[Pd]", "texts": ["Pd/C"]}]

Assistant response:
Step 1: m0 and m1 carry the printed labels 1a and 1b (reactant templates); m2 is the product template 
"2". m3 is "2a". m4's text reads "26", which breaks the sequence 2a, 2b: it is the OCR misread of "2b". 
m5 was expanded by the program from m2 with compound_id 2c, so it keeps "2c" and is a product. m6 is the
 catalyst Pd/C.
Step 2:
```json
{
  "m0": ["1a", "reactant template"],
  "m1": ["1b", "reactant template"],
  "m2": ["2", "product template"],
  "m3": ["2a", "2a, 99%", "product"],
  "m4": ["2b", "26, 85%", "product"],
  "m5": ["2c", "2c, X = Br, 92%", "product"],
  "m6": ["Pd/C", "conditions"]
}
```
\end{verbatim}
}

\subsection*{Text-based R-group Substitution Agent}
Sent with the graphic carrying the visual prompt; the model obtains the catalog of template label atoms, the condition text and the recognized molecules by calling the tools.
{\scriptsize
\begin{verbatim}
You are the text-based R-group substitution agent. The figure shows a reaction scheme template and a 
table whose rows give the R groups (and often the conditions and yields) of the individual reactions. 
Your task is to read the table and report, per row, the printed value of every variable of the template;
 the program builds the reactions from your answer.

Call "get_full_reaction" first. Its result is a catalog, not a symbol list: the template's reactant and 
product molecules with an id each, their label atoms (atom_id, symbol, variable = true for a placeholder
 such as [R], [R1], [Ar], [X] or *), the condition texts, and the list of recognised molecules with ids.
 The image you receive carries a program overlay: a thin blue outline around every recognised molecule 
with a numeric tag just outside it; tag N is the molecule with id "mN". The outline and the tag are not 
part of the figure.

Do:
1. Template corrections. Compare the label atoms with the drawing and report OCR misreads of label 
atoms, as template_corrections entries {atom_id, expected_symbol (copied from the catalog), 
corrected_symbol}: a placeholder ([Pt] drawn as R1 -> "[R1]"; * drawn as R -> "[R]") or an abbreviation 
/ formula written as text ([Ce2] printed as CS2 -> "[CS2]"; [1s] printed as Ts -> "[Ts]"). Never change 
a correctly read atom and never add or remove atoms.
2. Table rows. One entry per table row, in table order: reaction_id "<row>_1" ("1_1", "2_1", ...), 
bindings = an object mapping each variable NAME of the template (R, R1, R2, Ar, X ...) to the value 
printed in that row, exactly as printed (CO2Me, 4-BrC6H4, H, 2-furyl; never SMILES, never brackets), 
conditions = the row's own conditions as [{"role": "reagent"|"solvent"|"temperature"|"time"|"catalyst", 
"text": "..."}] (use the template's conditions when the row prints none), additional_info = [{"Yield": 
"88%", "R": "CO2Me", ...}] with whatever else the row prints (yield, ee, dr). A row whose table lists no
 R groups (a condition-screening table) has an empty bindings object.
3. Output one JSON object: {"template_corrections": [...], "rows": [...]}. Nothing else.

### Example
Catalog (abridged):
{"reactants": [{"id": "m0", "smiles": "*C(=O)c1ccc([Pt])cc1", "label_atoms": [{"atom_id": "m0:a000", 
"symbol": "*", "variable": true}, {"atom_id": "m0:a006", "symbol": "[Pt]", "variable": false}]},
               {"id": "m2", "smiles": "[R2]N", "label_atoms": [{"atom_id": "m2:a000", "symbol": "[R2]", 
"variable": true}]}],
 "products": [{"id": "m5", "smiles": "[R]C(=O)... ", "label_atoms": [{"atom_id": "m5:a000", "symbol": 
"[R]", "variable": true}, {"atom_id": "m5:a009", "symbol": "[R1]", "variable": true}, {"atom_id": 
"m5:a011", "symbol": "[R2]", "variable": true}]}],
 "conditions": [{"text": ["THF, r.t."]}]}
Table in the image: row 1: R = CO2Me, R1 = H, R2 = OMe, 88%; row 2: R = CO2Et, R1 = 4-BrC6H4, R2 = Cl, 
75%, 100  degC.

Answer:
```json
{
  "template_corrections": [
    {"atom_id": "m0:a000", "expected_symbol": "*", "corrected_symbol": "[R]"},
    {"atom_id": "m0:a006", "expected_symbol": "[Pt]", "corrected_symbol": "[R1]"}
  ],
  "rows": [
    {"reaction_id": "1_1", "bindings": {"R": "CO2Me", "R1": "H", "R2": "OMe"},
     "conditions": [{"role": "solvent", "text": "THF"}, {"role": "temperature", "text": "r.t."}],
     "additional_info": [{"Yield": "88%", "R": "CO2Me", "R1": "H", "R2": "OMe"}]},
    {"reaction_id": "2_1", "bindings": {"R": "CO2Et", "R1": "4-BrC6H4", "R2": "Cl"},
     "conditions": [{"role": "solvent", "text": "THF"}, {"role": "temperature", "text": "100  degC"}],
     "additional_info": [{"Yield": "75%", "R": "CO2Et", "R1": "4-BrC6H4", "R2": "Cl"}]}
  ]
}
```
\end{verbatim}
}

\subsection*{Condition Interpretation Agent}
Sent with the graphic; the model obtains the outputs of RxnConInterpreter and TesseractOCR by calling these tools.
{\scriptsize
\begin{verbatim}
You are the condition interpretation agent. your task is to extract the reaction condition texts in this
 reaction image, interpret the condition texts then output full condition list.
In this reaction graphic, there is a chemical reaction scheme template with multiple product, fragment 
molecular sub-images or tables and the conditions for every detailed reaction.
Your task is to:
1 .first call TesseractOCR tool to extract the condition text.
2. Then call RxnConInterpreter <"get_reaction_c" function> tool to interpret all the condition texts by 
identifying their condition roles initially.
3. Check the results and correct any errors. Place all the text except for "time", "temperature", 
"reagent", "solvent" and "yield" into the "add_info" category.
4. Read the output from the previous agents and align the interpreted reaction condition texts with the 
Reaction SMILES List, then output the final condition list.
5. your output should be in the JSON format. We will provide <examples> in the end to help you output.

<examples>:
### Example 1
**Assistant Response:**
Here is my step-by-step analysis:

**Step 1: Text Extraction & Interpretation**
* **Thought:** I need to extract text from the arrows and labels in the image using OCR, then interpret 
their chemical roles (reagent, solvent, etc.).
* **Observation:**
    * *Raw Text Found:* "5 mol% Cu(OTf)2", "6 mol% L1", "CH2Cl2", "0  degC, 24 h", "45-92%", "92%", 
"95:5 er". * *Interpreter Output:* Identified Cu(OTf)2 and L1 as reagents, CH2Cl2 as solvent, 0  degC as
 temperature, 24 h as time.

**Step 2: Error Correction & Categorization**
* **Thought:** I need to check the categorization.
    * "95:5 er" is a stereochemical outcome, not a reagent or condition. According to the rules, it must
 be moved to **"add_info"**. * "45 - 92%" is a range yield for the general template. * "92%" is a 
specific yield for product 4a.
* **Actions:** Reclassify the "er" values to `add_info`.

**Step 3: Alignment with Reaction List**
* **Thought:**
    * The "General Conditions" (catalyst, ligand, solvent, temp, time) apply to the 
**reaction_template**. * The specific values (Yield 92%, 95:5 er) are physically located next to 
molecule **4a**. * The specific values (Yield 88%, 93:7 er) are next to molecule **4b**.
* **Actions:** Construct the final JSON mapping conditions to their specific reaction IDs.

**Final JSON Output:**
```json{
'reaction_template': {'conditions': {"role":"reagent","text":"5 mol% 
Cu(OTf)2","smiles":"O=S(=O)([O-])C(F)(F)F.O=S(=O)([O-])C(F)(F)F.[Cu+2]"},{"role":"reagent","text":"6 
mol% L1","smiles":"CC(C)(C1=N[C@@H](c2ccccc2)CO1)C1=N[C@@H](c2ccccc2)CO1","label":"L1"},{"role":"solvent
","text":"CH2Cl2","smiles":"ClCCl"},{"role":"temperature","text":"0  degC"},{"role":"time","text":"24 
h"},{"role":"yield","text":"45 - 92%"}},
'reactions':{
'4a': {'conditions': {"role":"yield","text":"92%"},{"role":"add_info","text":"95:5 er"}},
'4b': {'conditions': {"role":"yield","text":"88%"},{"role":"add_info","text":"93:7 er"}},
'4c': ...
}
\end{verbatim}
}

\subsection*{Text Extraction Agent}
Sent with the graphic; the graphical results of the other agents are appended. In this prompt, RxnConInterpreter denotes the tool that runs ChemRxnExtractor on the text of the graphic.
{\scriptsize
\begin{verbatim}
System message:
You are the Text Extraction Agent. Your task is to extract text descriptions from chemical reaction 
images (or process direct text input), identify chemical entities and reactions within that text, and 
output a structured annotation.

User message:

In this input reaction graphic, there is a chemical reaction scheme template with multiple 
product/fragment molecular sub-images or tables, conditions, and a text description underneath (or you 
will receive direct text input).
Your task is to:

1. **Entity Extraction**: First call the `MolNER`("NER_from_text_in_image") tool to extract chemical 
identities (names, formulas, abbreviations) from the text description.
2. **Reaction Extraction**: Call the `RxnConInterpreter`("extract_reactions_from_text_in_image") tool to
 identify reaction types and relationships in the text description.
3. **Correction & Alignment**:
   - Check the tool results and correct any errors.
   - Align the identified text entities with the graphical output (e.g., mapping "precatalyst G32" in 
text to the molecule G32 in the image).
4. **Output**: Your final output must be in JSON format. Use specific tags like `[reactant]`, 
`[reagent]`, `[ABBREVIATION]`, `[SMILES:...]` to annotate the original text.

### Example 1
**User Input:**
Text Input: "A copper-catalyzed cycloaddition of benzyl azides and phenylacetylene was reported. Organic
 azides were prepared from the corresponding benzyl bromides. The tris(triazolylmethyl)amine ligand 
TBTA, labelled L3 in the scheme, accelerated the reaction."

**Assistant Response:**
Here is my step-by-step analysis:

**Step 1: Entity Extraction (MolNER)**
* **Observation:**
    * "benzyl azides" -> Chemical (Plural/Class)
    * "phenylacetylene" -> Chemical (Systematic Name)
    * "Organic azides" -> Chemical (Class)
    * "benzyl bromides" -> Chemical (Reactant precursor)
    * "TBTA" -> Abbreviation
    * "L3" -> Chemical (Identifier/Reagent)
* **Action:** Tag these entities with their specific roles and types.

**Step 2: Reaction & Structure Interpretation (RxnConInterpreter)**
* **Observation:**
    * "L3" is identified as the ligand drawn in the scheme.
    * **Action:** Retrieve/Associate the SMILES for L3 from the tool/image context: 
`c1ccc(Cn2cc(CN(Cc3cn(Cc4ccccc4)nn3)Cc3cn(Cc4ccccc4)nn3)nn2)cc1`.
    * "benzyl azides" and "phenylacetylene" are participating reactants in an azide-alkyne 
cycloaddition.

**Step 3: Text Alignment & Annotation**
* **Thought:** I need to inject the extracted tags and SMILES directly into the original text string to 
create the final description.
* **Action:** Construct the annotated string.

**Final JSON Output:**
```json
{
  "annotated_text": "A copper-catalyzed cycloaddition of benzyl azides [reactant][MULTIPLE] and 
phenylacetylene [reactant][SYSTEMATIC] was reported. Organic azides [FAMILY] were prepared from the 
corresponding benzyl bromides [reactant][MULTIPLE]. The tris(triazolylmethyl)amine ligand TBTA 
[ABBREVIATION], labelled L3 
[reagent][IDENTIFIERS][SMILES:c1ccc(Cn2cc(CN(Cc3cn(Cc4ccccc4)nn3)Cc3cn(Cc4ccccc4)nn3)nn2)cc1] in the 
scheme, accelerated the reaction."
}
```
\end{verbatim}
}

\subsection*{Confidence Scorer: language-model checker}
Sent with the graphic and the extracted reactions.
{\scriptsize
\begin{verbatim}
You check the reactions a program extracted from the chemistry figure attached. You answer fixed 
questions about them; you do not correct anything.

Extracted reactions (JSON). Each has an id (R1, R2, ...), reactants and products as SMILES with their 
printed labels, and conditions.
{reactions}

For every reaction id, answer each question with "yes", "no" or "unsure":
- drawn: the figure shows this reaction (the main scheme, one entry of a substrate scope, or one row of 
a table). Put the label or entry you matched in "entry", or "" when there is none.
- complete: every reactant and product the figure gives for that reaction is present, and nothing the 
figure does not give.
- structures: the reactant and product SMILES match the structures the figure draws or defines for that 
reaction (substituents, their positions, rings, charges and counter-ions).
- conditions: the catalysts and reagents the figure gives for that reaction (drawn, or named by label) 
appear among its conditions.
- resolved: every variable group of that reaction (R, Ar, X, n) is replaced by the value the figure 
gives; no "*" is left where the figure defines the group.
- duplicate: another extracted reaction is the same reaction.
- stereo_drawn: the figure draws at least one wedged or hashed bond on a product of this reaction (on 
the product itself, or on the template product it is built from), or states its absolute or relative 
configuration. Do not judge whether the SMILES gets the configuration right.
- counterion_missing: the figure gives a counter-ion or salt for a reactant, product, catalyst or 
reagent of this reaction (drawn next to the structure, in its label, in a caption or in a footnote such 
as "BF4 counterions omitted for clarity"), and that counter-ion is absent from the corresponding SMILES.
And write "problem": the most important thing wrong with this reaction compared with the figure, in a 
few words, or "" when you see nothing wrong.
Then, for the whole figure:
- expected_reactions: how many distinct reactions the figure defines (scheme arrows, scope entries and 
table rows that are reactions).
- missing: the labels or entries of reactions the figure defines that no extracted reaction covers.
Answer "unsure" when the figure does not let you decide; do not guess.

Respond with one JSON object and nothing else:
{"reactions": [{"id": "R1", "entry": "3a", "drawn": "yes", "complete": "yes", "structures": "yes", 
"conditions": "unsure", "resolved": "yes", "duplicate": "no", "stereo_drawn": "no", 
"counterion_missing": "no", "problem": ""}], "expected_reactions": 1, "missing": []}
\end{verbatim}
}

\subsection*{Main Text Agent: figure roles}
Text only, for up to 25 figures per call; \texttt{\{figures\}} is replaced by the caption, footnotes, printed text and citing sentences of each figure. A role whose quote does not occur verbatim in these texts gives way to a rule on the caption or printed text when one applies.
{\scriptsize
\begin{verbatim}
System message:
You classify the figures of chemistry articles.

User message:
You label what each figure of a chemistry article shows, from its caption, footnotes, printed text and 
the sentences that cite it. Do not guess from chemical knowledge.

Roles:
- reaction_scheme: a reaction or a synthetic route
- substrate_scope: one reaction applied to many substrates (products with yields)
- optimization: the same reaction under different conditions (entries)
- mechanism: a proposed or plausible mechanism, catalytic cycle or intermediates
- control_experiment: control or mechanistic experiments that were run
- synthetic_application: products used further (derivatization, transformations, total synthesis)
- scale_up: gram-scale or large-scale runs
- compound_structures: structures only (bioactive compounds, catalysts, ligands), no reaction
- spectra_or_data: spectra, chromatograms, plots, crystal structures, data tables that are not reactions
- toc_graphic: the graphical abstract
- other

For every figure give "role", "reactions_are" ("performed": reactions that were run; "proposed": 
hypothetical steps such as a mechanism; "none"), "evidence": a verbatim quote (3 to 30 words) from that 
figure's texts below, and "source": the id of the text the quote comes from.

Figures:
{figures}

Respond with one JSON object and nothing else:
{"figures": [{"figure_id": "main_p2_1", "role": "substrate_scope", "reactions_are": "performed", 
"evidence": "...", "source": "caption"}]}
\end{verbatim}
}

\subsection*{Main Text Agent: figure roles from the image}
Sent with the image of a figure that has no caption, footnotes or printed text.
{\scriptsize
\begin{verbatim}
System message:
You classify the figures of chemistry articles.

User message:
The image is one region cut from a page of a chemistry article or its supporting information. It has no 
caption or printed text to go by. Say what it shows.

Roles: reaction_scheme, substrate_scope, optimization, mechanism, control_experiment, 
synthetic_application, scale_up, compound_structures (structures only, no reaction), spectra_or_data 
(NMR, MS or IR spectra, chromatograms, plots, crystal structures, even when a structure is drawn beside 
them), toc_graphic, other.

Respond with one JSON object and nothing else:
{"role": "spectra_or_data", "reactions_are": "none", "description": "1H and 13C NMR spectra of 3b"}
\end{verbatim}
}

\subsection*{Main Text Agent: conditions and yields from the text}
Text only, one call per figure; the placeholders are replaced by the name and page of the figure, its caption, footnotes and citing paragraphs, and the reactions extracted from it. Items whose quote does not occur verbatim in the cited text are discarded.
{\scriptsize
\begin{verbatim}
System message:
You attach text evidence to extracted chemical reactions.

User message:
You add information that the article text states about reactions already extracted from one figure of 
the same article. You never invent, infer or use chemical knowledge.

Figure: {name} (page {page})

Text sources, each with an id:
{sources}

Reactions extracted from this figure (JSON):
{reactions}

Tasks:
1. "figure_conditions": conditions that the caption or a footnote states for all reactions of this 
figure (for example "Reaction conditions: 1a (0.2 mmol), 2a (0.3 mmol), CuI (10 mol %), DMF (2 mL), 80  
degC, 12 h"). Give one item per reagent, catalyst, solvent, temperature, time or atmosphere. Leave out 
the substrates and products themselves.
2. "reactions": for each reaction id, the items the text states for that specific reaction (the text 
names its product or reactant label, or its table entry) that the reaction does not already contain:
   - "conditions": role reagent, catalyst, solvent, additive, ligand, base, temperature, time, 
atmosphere, light or other;
   - "additional_info": role yield, ee, er, dr, conversion, selectivity or note.

Rules:
- Every item has "role", "text", "evidence" and "source". "evidence" is an exact, verbatim quote (5 to 
40 words) copied from the source named by "source". Items whose evidence is not literally in that source
 are discarded.
- "text" is the value as printed (for example "CuI (10 mol %)", "80  degC", "92% ee"), and it must 
appear inside the evidence.
- Skip anything the reaction already contains (the same reagent, temperature or yield, even if written 
differently).
- Do not write SMILES. If nothing applies, return empty lists.

Respond with one JSON object and nothing else:
{"figure_conditions": [{"role": "solvent", "text": "DMF", "evidence": "...", "source": "fn_a"}], 
"reactions": [{"reaction_id": "...", "conditions": [], "additional_info": []}]}
\end{verbatim}
}

\subsection*{Main Text Agent: reactions stated in the text}
Text only, one call per text chunk of about 6,000 characters of the article and its supporting information; the placeholders are replaced by the compound labels and names of the article, the reactions extracted from its figures and the text. A reaction is kept only if its quote is found verbatim in the cited paragraph; its structures are taken from the Label Registry or OPSIN, never from the model, and a reaction whose product matches a figure reaction is attached to it as text evidence.
{\scriptsize
\begin{verbatim}
System message:
You extract the reactions stated in chemistry article text.

User message:
You read part of the text of a chemistry article and list the chemical reactions it says were carried 
out. You never invent, infer or use chemical knowledge.

Compounds of this article (label: name when the text names it):
{compounds}

Reactions already extracted from the article's figures (product <- reactants, by label):
{figure_reactions}

Text (each paragraph starts with its id in brackets):
{text}

List every reaction the text states was performed in this work: a procedure (general or for one 
compound), the preparation of a starting material, a transformation of a product, a control experiment. 
Skip reactions that are only cited from other work or only proposed (mechanisms).
For each reaction give:
- "reactants" and "products": lists of {"label": the printed compound label or "", "name": the compound 
name as printed or ""}, each with a label or a name;
- "conditions": [{"role": "reagent|catalyst|solvent|temperature|time|atmosphere|other", "text": "..."}],
 as printed;
- "yield": as printed, or "";
- "evidence": an exact, verbatim quote (5 to 60 words) from the paragraph that states the reaction, and 
"source": that paragraph id;
- "in_figure": true when it is one of the figure reactions listed above.
Never write SMILES. Copy labels and names exactly as printed. If there is none, return an empty list.

Respond with one JSON object and nothing else:
{"reactions": [{"reactants": [{"label": "1a", "name": ""}], "products": [{"label": "3a", "name": ""}], 
"conditions": [], "yield": "85%", "evidence": "...", "source": "main:p12", "in_figure": false}]}
\end{verbatim}
}


\subsubsection*{Ground-truth annotation}
\paragraph*{Annotators and training}
The benchmark was annotated by four chemists trained in chemical and biomolecular engineering, from senior
undergraduate to senior doctoral level (Table~S5). All four use
structure-drawing programs and RDKit-based SMILES tools, and each completed a training session on the
annotation guideline, with practice graphics outside the benchmark and feedback, before annotation began. No
annotator had access to another annotator's records.

\paragraph*{Guideline, output schema and canonicalization}
All annotators followed one written guideline. It defines (i) the unit of annotation: every substrate or product
variant defined by an R-group table, scope panel or textual enumeration is a separate reaction, and each step of a
multistep scheme is a separate reaction whose intermediate is the product of one step and a reactant of the next;
(ii) the fields: reactants and products as SMILES with their printed labels, and conditions as role-tagged entries
(catalyst, reagent, solvent, temperature, time, yield) holding the verbatim text and, for chemical species, a resolved
SMILES; (iii) stereochemistry as drawn, including wedged and hashed stereocenters and double-bond geometry; and
(iv) conventions for ambiguous cases, such as unlabeled variants and substitutions defined in footnotes. The output
schema is the JSON structure produced by ChemEAGLE, so that human and model outputs are directly comparable. For the
agreement analysis, all records passed through a shared canonicalization: SMILES were canonicalized with RDKit,
multi-component species were written in dot-separated form, chemical names in condition fields were resolved through
the same name-resolution chain as the system, and molecules were compared by their International Chemical Identifier keys (InChIKeys). The guideline is released with
the benchmark.

\paragraph*{Agreement and adjudication}
Agreement was computed over all 3,613 reactions from the four complete annotation sets (Table~S6). Because reaction-level judgments are strongly skewed toward agreement, kappa-type
coefficients are unstable here (the prevalence paradox); we therefore report Gwet's AC1~\cite{gwet2008ac1}, which is
designed for high-agreement settings, together with the raw pairwise scores. A reaction entered adjudication if any pair
of annotators differed in any field. These 5.5\% of the reactions were adjudicated by the first author following a
documented protocol: the record was compared field by field with the graphic and with the four annotations, and the
version supported by the graphic was kept. Every reaction was then audited against its graphic, as described in section ``Benchmark and ground truth''.


\subsubsection*{Matching criteria and their scope}
\paragraph*{Hard and soft criteria}
Under both criteria, every reactant and product structure must match the ground truth, and a molecule filed as a
reactant in one record and as a condition in the other is not an error; under the hard criterion, used throughout the
main text, every condition structure that the graphic draws must also match (section ``Evaluation'').
Fig.~S8 shows an output that is correct under the soft criterion but not under the hard
one.

\paragraph*{Stereochemistry in the benchmark}
We surveyed all 2,983 ground-truth reactions of the article subsets and all 630 of the patent subset for the carriers
of stereochemical information. In the article subsets, 1,962 reactions (65.8\%) contain drawn stereocenters or defined
double-bond geometry, which ChemEAGLE extracts as drawn; in the patent subset, 347 reactions (55.1\%) do. Textual
qualifiers were found by a keyword census of the condition fields and compound labels (racemic, rac-, ($\pm$), ee, dr,
er, enantio-, diastereo-, chiral, optically pure, single enantiomer, absolute configuration). In the condition fields of
the article subsets, 125 reactions (4.2\%) carry matching text: 112 are measured selectivities such as ``92\% ee'',
``$>$20:1 dr'' or ``diastereomixture (53/47)'', 10 describe a catalyst (for example ``chiral imidazolinium salt'') and
3 designate a racemic substrate. Measured selectivities and catalyst descriptions are results or reagent descriptions,
not qualifiers that override the drawn configuration, and they are extracted as condition text; the patent subset
contains no ee, dr or er text. Racemic designations printed in the graphic (``rac-'', ``($\pm$)'', ``trans racemic'')
occur in 45 of the 3,613 reactions (1.2\%) from 9 of the 424 graphics (25 reactions in five article graphics and 20 in
four patent graphics). In 24 of these reactions the designated structure is drawn without stereocenters, in 11 the
designation appears only in the condition fields, and in 10 (0.3\% of all reactions) it is attached to a structure drawn
with stereocenters, the only case in which the drawn structure could misrepresent the stereochemical outcome. These
designations are captured as compound labels or condition text together with the structures. ChemEAGLE extracts stereochemistry as drawn; reconciling it with the running text of an article is left to downstream
processing. Under both matching criteria, stereochemistry must match (section ``Evaluation'').


\subsubsection*{Document-level evaluation against Reaxys}
Whole-article extraction was evaluated on the 19 articles from J. Org. Chem. and Org. Lett. of the OpenChemIE
document benchmark~\cite{fan2024openchemie}, whose 2,149 reactions are indexed in Reaxys~\cite{reaxys}. All reactions that an
article yields, from graphics and from the text, were pooled and compared with the Reaxys records of that article. A
Reaxys reaction was matched by an output reaction when both its products and its reactants matched: every Reaxys
product occurred among the output products and every Reaxys reactant among the output reactants. Molecules were compared exactly after canonicalization, as in the benchmark evaluation (section ``Evaluation''),
so each Reaxys product and reactant had to be written exactly. Matching was one to one and
greedy in the order of the Reaxys records. Precision is the fraction of output reactions matched and recall the
fraction of Reaxys reactions matched. Because Reaxys does not index every reaction reported in an
article, unmatched output reactions are not necessarily wrong, and precision measured this way is a lower bound.

For Fig.~2G, each Reaxys reaction was assigned to one of three locations in its article: drawn in a graphic only (1,246
reactions), reported in a graphic and in the text (645) or stated in the text only (258). For Fig.~2F, the document
modules were switched on one after another in the order shown, with all other settings fixed.


\subsubsection*{Web application}
The ChemEAGLE web application (\url{https://app.chemeagle.net/}) lets researchers upload reaction graphics or PDF
documents and extract their reactions without local installation. Two versions are offered: one runs on a model
reached with the user's application programming interface (API) key through an OpenAI-compatible interface, with Gemini 3.7 Flash as the recommended
closed model, and the open-weight version runs on Qwen3.8-27B. Users can upload or paste a screenshot of any graphic, from a single molecule to a complex
reaction scheme, or a whole PDF. For a PDF, the document layer extracts the figures and tables that contain chemistry and
converts the body text to Markdown, and each extracted graphic is processed by the system described above.

Fig.~S18 shows the interface during the extraction of a PDF. The left panel holds the API
settings, the file upload and the results per graphic: extracted graphics appear as thumbnails, each marked with a check
once its extraction is complete. The top right panel embeds the Ketcher editor, and the bottom right panel shows the PDF
page with every detected region as a clickable box next to the parsed body text. Clicking a thumbnail moves the PDF view and the parsed text to its source page and outlines the region; clicking a region selects its thumbnail. The extracted SMILES are listed below
the thumbnails, and selecting one draws the reactants, products and condition structures in Ketcher, with the
conditions (reagents, solvents, atmosphere, temperature, yield) above it. Users can thus check each extraction against
its source in context and correct structures in Ketcher before exporting the data. The application accepts PNG, JPG
and PDF files, runs on a server accessed through any modern browser and, with explicit user consent, collects anonymized
usage data and feedback.


\subsubsection*{Maintenance and sustainability}
ChemEAGLE is maintained as an open-source project at \url{https://github.com/CYF2000127/ChemEagle}. The maintenance plan has five
parts. (1) Versioned releases: the repository follows a quarterly release cycle with tagged releases and changelogs.
(2) Model abstraction: every base-model call goes through one OpenAI-compatible interface, so changing the base model
requires only a configuration change, with no change to agents, tools or prompts; the seven base models of Table~S8 were evaluated through this interface. (3) Dependencies: all Python dependencies are pinned in the release
manifest; when an upstream library introduces breaking changes, the pinned environment preserves reproducibility, and a
tested migration is released before the pinned versions are updated. (4) Web services: the public web application
(section ``Web application'') is hosted on the HPC4 cluster of the Hong Kong University of Science and Technology (HKUST), and an offline mode of name resolution,
which uses only the alias map, the cache, the condensed-formula parser and a local copy of OPSIN, keeps the core
extraction working without external web services. (5) Issues and contributions: reported
issues are addressed as they arise, independently of the release cycle, and extraction errors reported through the web
interface feed the continued improvement of prompts and tools.

\clearpage

\clearpage
\subsection*{Supplementary Text}

\subsubsection*{Audit of the observers}
We audited the two observers in the first of the three runs of the full system on all 424 graphics (Gemini 3.7 Flash;
Fig.~S6). For the Plan Observer, an intervention is a plan whose set of agents it changed. A change counted
as justified when the original plan lacked an agent that the graphic needed or used the wrong one, judged by the kinds of
graphic and the agents responsible for them (Materials and Methods), and as fixed when the part of the graphic that the change
concerned was extracted correctly under the hard criterion in the final output. For the Action Observer, an intervention
is a check that flagged an issue. A flag counted as justified when the issue was present in the agent's output, compared
with the ground truth, and as fixed when the final output no longer contained it, through a re-execution or because the
final assembly step used the note. All 424 graphics were also screened for plan or agent errors that no observer flagged.

The Plan Observer changed 89 of the 424 plans (0.21 per graphic). Of these changes, 78 (87.6\%) were justified, and 66 of
them (84.6\%) led to a correct extraction of the affected part; 35 graphics (8.3\%) kept a plan error that the observer
did not change. The Action Observer ran 1,158 checks and flagged an issue in 354 of them (0.83 per graphic). Of the flags,
320 (90.4\%) were justified and 223 of these (69.7\%) were fixed in the final output; 29 checks led to a re-execution of the
agent. In 51 graphics (12.0\%), an agent error went unflagged. The rates were similar for article and patent graphics
(Table~S3).


\subsubsection*{Complete outputs on Case II and the sources of generalist-agent errors}
Figs.~S9 to S11 show the complete outputs summarized in Fig.~2C for
ChemEAGLE, for Gemini 3.7 Flash writing structures end to end and for Claude Code (Claude Sonnet 5), listed in the same
format.

To find where the errors of the generalist agents arise, we replayed the tool calls of Claude Code (Claude Sonnet 5)
and Codex (GPT-5.6-sol) on all 424 benchmark graphics. A molecule in its output counts as copied when its canonical SMILES equals a structure
that one of its tool calls returned for that graphic (image tools per graphic, name resolution and R-group
reconstruction per run). All other molecules count as written, including structures the agent edited after a tool call
and structures it obtained by calling Graph2SMILES on a graph it had modified. A molecule counts as correct when it
equals any ground-truth molecule of the same graphic. Of the incorrect products and reactants, 71.2\% and 76.9\% had
been written by Claude Code and 72.7\% and 79.5\% by Codex (Table~S9). Both agents copied
fewer than 60\% of their products and reactants from a tool output, against 97--98\% for ChemEAGLE (Fig.~3B).


\subsubsection*{Procedure and costs of the human workflow study}
The four annotators (Table~S5) each processed the complete benchmark in two workflows:
verification and correction of ChemEAGLE's output through the web interface (section ``Web application''),
and manual annotation from scratch. In the verification workflow, the interface draws every extracted reaction with the
embedded Ketcher editor next to the source graphic. When a structure is wrong, the annotator corrects it in Ketcher and
copies the corrected SMILES into the record; condition and label fields are edited as text. Verification and
correction times were logged separately for each graphic. Each annotator did the verification pass before the manual
pass, so any learning effect favors the manual workflow. The run time of automated extraction is the mean of three
runs.

The same annotators then reviewed only the reactions scored medium or low by the Confidence Scorer, accepting
high-confidence reactions unchanged, with time logged in the same way. These reactions made up 35\% of the output.

Costs combine the measured time with a labor rate of US\$30 per hour and the cost of automated extraction (US\$0.1106 per
graphic, the cost of the model calls of ChemEAGLE with Gemini 3.7 Flash; Fig.~3A). The rate lies between the US Bureau of Labor Statistics May 2024
median wages~\cite{bls2025oews} for chemical technicians (US\$27.78 per hour, occupation 19-4031) and chemists (about US\$40 per hour,
occupation 19-2031), and slightly below the entry-level postdoctoral stipend of the National Institutes of Health (NIH) National Research Service Award
for fiscal year 2025~\cite{nih2025nrsa} (US\$62,232 per year, about US\$31 per hour at 2,000 working hours). A
higher rate only widens the advantage of automated extraction, which stays cheaper than manual annotation for any labor
rate above US\$0.28 per hour, even when every output is reviewed (US\$0.1106 divided by the 0.397 h saved per graphic). Table~S10 gives the time,
accuracy and cost of each workflow and annotator, and Table~S11 the split of the
time spent reviewing every output.


\subsubsection*{Text Extraction Agent on text-only tasks}
The Text Extraction Agent and its tools were also evaluated on two text-only tasks, with Gemini 3.7 Flash
as base model, and compared with specialist models and with the four MLLMs of Fig.~S14 used alone. For
chemical named entity recognition (Fig.~S15), the Planner calls the Text Extraction Agent and
the Data Structure Agent, and MolNER tags and classifies chemical mentions, which the model then corrects. On the
CHEMDNER test set~\cite{krallinger2015chemdner} (3,000 abstracts with 25,351 mentions), ChemEAGLE reached an F1 score of
95.7\%, compared with 87.6\% for MolNER, 82.3\% for ChemBERTa~\cite{chithrananda2020chemberta}, 80.7\% for
BioBERT~\cite{lee2020biobert} and 91.3--94.4\% for the four MLLMs.

For text-based reaction extraction (Fig.~S16), ChemRxnExtractor identifies products and
the roles of the other mentions, and the model corrects and completes them. On 111 annotated reactions in 33 paragraphs
of scientific articles, ChemEAGLE reached an F1 score of 95.9\%, against 91.2\% for the fine-tuned
RxNExtract~\cite{kumari2025enhanced}, 76.7--83.0\% for ChemBERT~\cite{guo2022chemrxnextractor},
ChemRxnExtractor~\cite{guo2022chemrxnextractor}, ReactIE~\cite{zhong2023reactie} and Para2RXNRole~\cite{zhang2024finetuning},
and 87.9--92.4\% for the four MLLMs. The MLLMs performed much better on these text tasks than on reaction graphics
(Fig.~2A).


\subsubsection*{Audit of name-to-structure resolution}
We audited every name-to-structure resolution in the benchmark ground truth (190 unique names with 1,724 occurrences,
after excluding 761 compound labels that belong to visual coreference). Each name was resolved by the algorithmic tiers of the chain
described in the Materials and Methods, without its alias map, including rules that split multi-component names (concentrations, ratios, hydrates), and the
result was compared with the ground truth by InChIKey skeleton blocks. The algorithmic tiers resolved 87.5\% of the
occurrences correctly (1,509 of 1,724). The 182 occurrences (10.6\%) they could not resolve concentrate in 40 reagent
shorthands, almost all organometallic or salt formulas such as Mg(O\textit{t}Bu)\textsubscript{2}, KO\textit{t}Bu and
\textit{n}Bu\textsubscript{4}NPF\textsubscript{6}, which the alias map covers. A further 1.9\% resolved to
structures that disagree with the ground truth; these were adjudicated as defects of the ground-truth records rather
than resolution errors. Free-text descriptions such as ``base, e.g.\ sodium hydride'' are passed to the language-model
fallback.

To characterize this fallback in isolation, the 57 hardest names (46 that the algorithmic tiers cannot resolve and 11
short abbreviations with synonym collisions in web databases) were posed to the seven base models of Table~S8 in single calls that mirror the fallback (Table~S4). Outputs were validated with RDKit and
compared with the ground truth by InChIKey. Sixteen names with defective ground truth or content that cannot be
answered in principle (R-group wildcards, OCR corruptions) were excluded, leaving 41. Chemical validity ranged from
90.2 to 95.1\%, and every invalid output is intercepted by the RDKit validation before it can be accepted.
The fallback can receive at most the 10.6\% of name occurrences that the algorithmic tiers leave unresolved, and this
test set contains the hardest of them. With Gemini 3.7 Flash (75.6\% correct), its contribution to errors in the
conditions is therefore at most about 2.6\% (10.6\% $\times$ 24.4\%).


\subsubsection*{Robustness to the in-context example and the prompt}
The prompt of each extraction agent contains one worked example, or a pair of short examples used as one (Materials and Methods).
With Gemini 3.7 Flash as base model, on all 424 graphics and with three runs per condition, we varied the example
and the wording of the prompt (Fig.~S13). To test whether performance rests on the choice
of example, we removed it, replaced it with one of two other examples written in the same format (alternative
examples A and B) and gave all three examples together; the Planner and the observers, whose prompts contain no
worked example, were left unchanged. To test the wording, we shuffled the order of the rule blocks (a different random
seed in each run), replaced the role definitions with neutral task statements and removed the blocks of
chemical-notation rules. Each condition's three runs were compared with the three runs of the default configuration
by a two-sided Welch's t-test.

Without the example, hard-match F1 was 74.7 $\pm$ 0.6\%, against 82.2 $\pm$ 0.9\% with it (the three runs of Table~S8; \textit{P} = 0.0007). With the alternative examples it was 82.2\% and 82.1\% (\textit{P} = 0.93 and
0.92), and with all three examples 83.4\% (\textit{P} = 0.33). Shuffling the rule blocks changed F1 by $-$0.08 points
(\textit{P} = 0.90), neutral role definitions by $-$0.04 points (\textit{P} = 0.96) and removing the notation rules by
$-$1.44 points (\textit{P} = 0.10). Apart from removing the example, every change stayed within twice the run-to-run
SD of the default configuration. The
notation rules encode general conventions of chemical drawings and were written on the 20 development graphics outside the benchmark.


\subsubsection*{Further results on reaction prediction models trained on the photochemistry database}
\paragraph*{Photocatalysts and further measures of extraction quality}
The substrate-scope reactions extracted by ChemEAGLE span the common classes of photocatalyst (Fig.~S20A): organic photocatalysts in 29.8\% of the 53,184 reactions, iridium complexes in 24.5\%,
ruthenium complexes in 3.5\%, decatungstate in 1.8\% and other metal complexes in 13.6\%; 8.1\% name a photocatalyst
without drawing it, and 18.7\% use none.

Fig.~S20B compares ChemEAGLE with the end-to-end extraction by the same base model (Gemini 3.7 Flash) by
measures that do not depend on the output format; Fig.~5C of the main text gives the hard-match precision, recall and
F1. On the 198 test tables, the end-to-end extraction wrote the right product for 49.0\% of the tables, the right
number of rows for 19.2\% and the yields in table order for 9.6\%, against 100\%, 97.5\% and 97.0\% for ChemEAGLE. Of
the 154 tables that use a photocatalyst, it wrote every photocatalyst structure of the table, in any role, for 11.0\%,
against 98.7\%. On the 248 test scope graphics, 8.1\% of its reactions contained a structure that could not be parsed,
against 1.2\% for ChemEAGLE; it never gave a photocatalyst role, and it gave the light source for 25.3\% of the
reactions and the temperature for 30.0\%, against 93.9\% and 75.4\%. ChemEAGLE's values on these graphics come from its
hand-checked records; on the training graphics, whose records were not reviewed, they are 99.3\% (structures parsed),
81.9\% (photocatalyst role), 98.1\% (light source) and 78.1\% (temperature), against 92.8\%, 0\%, 23.9\% and 32.4\% for
the end-to-end extraction.

\paragraph*{Selection of the ranking configuration}
The features and hyperparameters of the ranking models were chosen on the training tables by a split in time:
tables up to 2022 and those of unknown year for training, and tables from 2023 and 2024 for validation. In the last
step of this selection, the few best configurations were also evaluated on the 118 hand-checked tables from 2025,
which are part of the 198 test tables. The 80 hand-checked tables from 2026 took no part in the selection. Evaluated
once with the fixed configuration, the ensemble ranked the best condition first in 47.2\% of them (95\% confidence interval, 37.8 to
56.8\%), against 11.3\% for a random order and 51.4\% on the tables from 2025.

\paragraph*{Learning curves of condition ranking}
To compare how efficiently the ranker learns from each source, the tuned LambdaRank was trained on 10\%, 20\%, 35\%, 50\%, 75\%
and 100\% of the training tables, with five random subsets per fraction and the same graphics drawn for both sources,
and evaluated on the 198 test tables (Fig.~S21). With species pooled regardless of
role, 58 ChemEAGLE tables gave a rank correlation of 0.459 between predicted and reported yields, which the end-to-end
records reached only with about 370 tables. The share of tables with the best condition ranked first differed between
the two sources by at most 1.6 percentage points at every size, less than the variation between subsets.

\paragraph*{Product prediction in two other settings}
The main text tests product prediction on articles that the model has not seen. Two easier settings are common in the
literature (Fig.~S22, A and B). In the first, five reactions of each test graphic were added
to training and the remaining 3,514 reactions of 209 graphics were tested (graphics with fewer than eight test
reactions were left out). In the second, all usable reactions were deduplicated and split at random into training,
validation and test sets (80, 10 and 10\%; 4,414 test reactions from 1,734 graphics), so that 99.7\% of the test
reactions had another reaction of the same graphic in training. Fine-tuned on ChemEAGLE's records, ReactionT5v2-forward
reached a top-1 accuracy of 58.1 $\pm$ 1.0\% and 70.5 $\pm$ 0.9\% in the two settings, against 35.4 $\pm$ 0.9\% and
39.1 $\pm$ 0.8\% when fine-tuned on the end-to-end records and 7.1\% and 7.7\% for the pretrained model (mean $\pm$ SD of
three runs). In both settings, the training data from the end-to-end extraction were built by the same procedure as those from
ChemEAGLE, and all test labels were checked by hand.

Removing the 4,439 reactions that the Confidence Scorer rated low, which leaves 29,815 training reactions, lowered
top-1 accuracy on unseen articles from 32.7 $\pm$ 0.8\% to 31.5 $\pm$ 0.8\%, by 1.2 $\pm$ 0.1 percentage points (mean $\pm$ SD
of the paired differences over three runs). Removing as many reactions at random lowered it to 30.1 $\pm$ 1.0\%, by
2.6 $\pm$ 0.1 points (Fig.~S22C). The low-confidence reactions thus added less to
training than the average reaction, but removing them did not improve the model.

\paragraph*{Errors added to the training data}
Table~S12 defines the errors added to the training data in Fig.~5, F and I, of the main
text. Each kind was added alone to a growing share of the training tables or graphics, chosen at random in each run;
the test data were left unchanged.


\clearpage
\setcounter{figure}{0}
\begin{figure}[p]
\centering
\includegraphics[width=\textwidth,height=0.66\textheight,keepaspectratio]{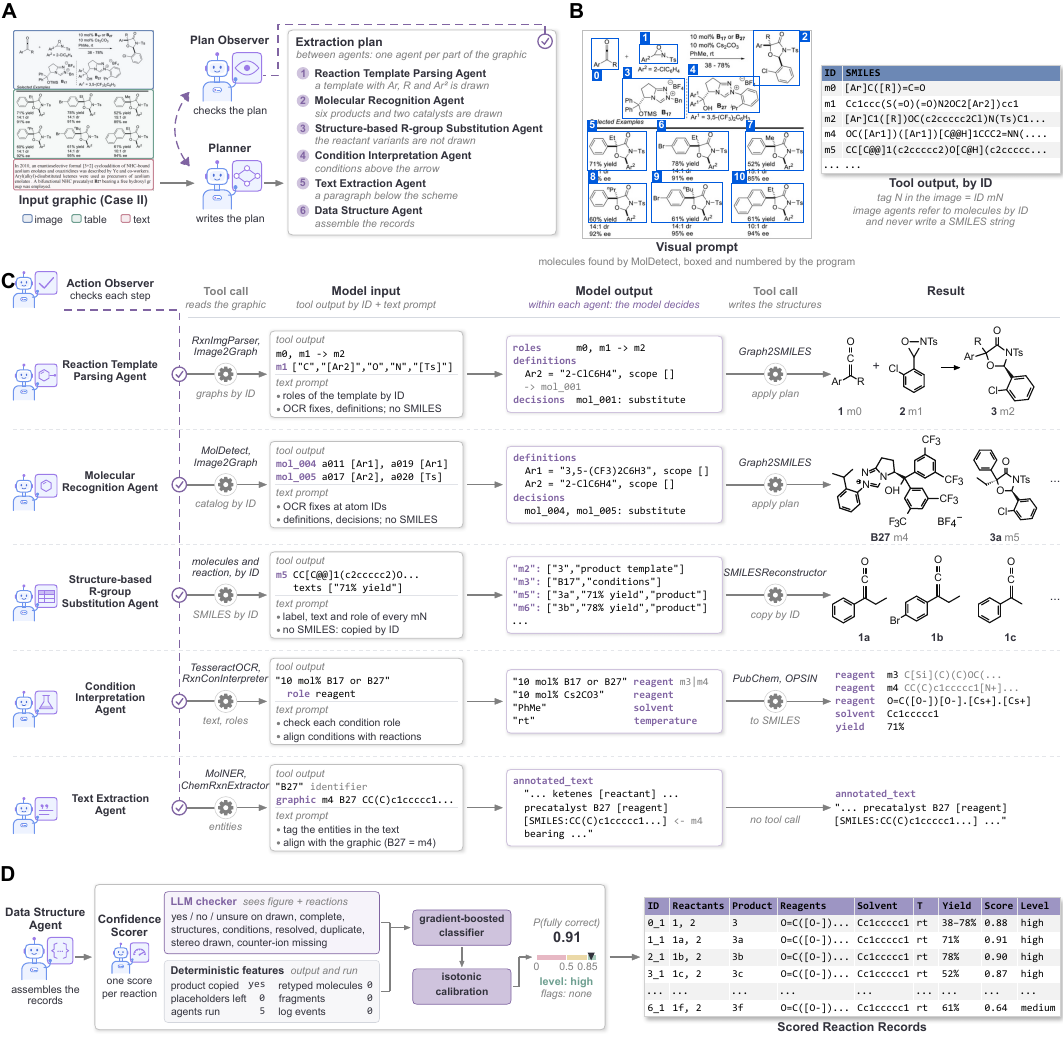}
\caption{\textbf{Detailed extraction workflow on a representative reaction graphic.} (\textbf{A}) Between agents: the Planner assigns one agent to each part of the graphic, and the Plan Observer checks the plan. (\textbf{B}) Visual prompt: the program boxes and numbers the molecules found by MolDetect; tag N is molecule mN of the tool output, and image agents cite these identifiers instead of writing SMILES. (\textbf{C}) Within each agent: a tool call reads the graphic; the model receives the tool output by identifier with a text prompt and returns decisions, such as corrections and definitions for molecules, labels and roles for R-group substitution, or condition roles; and a second tool call writes the structures. The Text Extraction Agent returns annotated text with structures taken from the graphical results, and the Action Observer checks every step. Structures are shown with their compound label and molecule identifier. (\textbf{D}) The Data Structure Agent assembles the records, and the Confidence Scorer combines an optional large language model (LLM) checker with deterministic features in a calibrated classifier, returning for each reaction the probability that it is correct, a level and any flags. Prompts are abridged; model outputs and scores are illustrative.}
\end{figure}

\setcounter{figure}{1}
\begin{figure}[p]
\centering
\includegraphics[width=\textwidth,height=0.78\textheight,keepaspectratio]{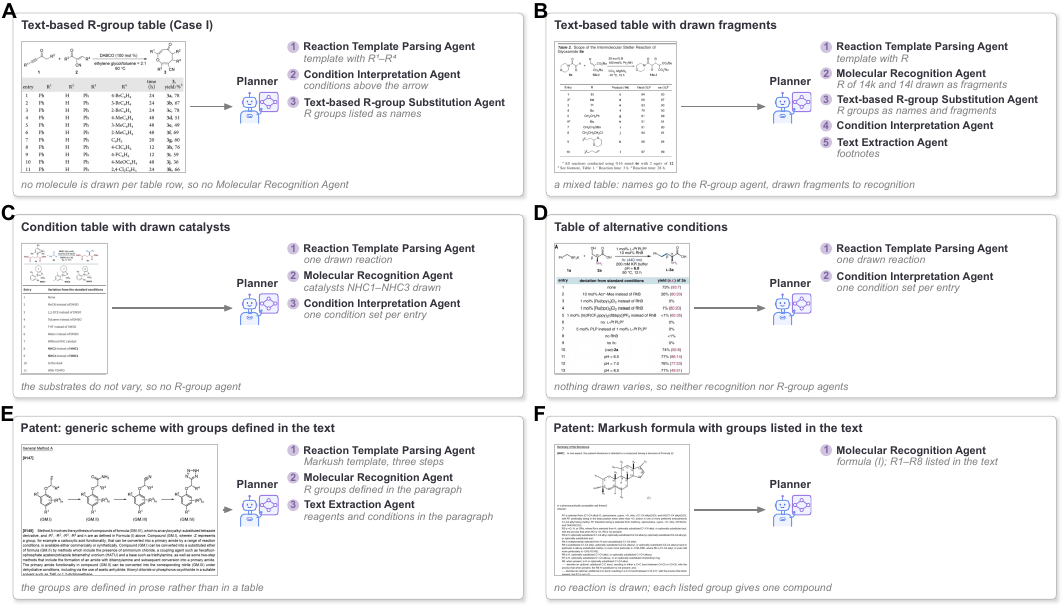}
\caption{\textbf{Plans for six kinds of reaction graphic.} Each panel shows a graphic (left) and its planned agents, in order (right); gray text gives the part that each agent handles. (\textbf{A}) A text-based R-group table (Case I): substituents are names and no molecule is drawn per row, so no Molecular Recognition Agent is planned. (\textbf{B}) A text-based table with two substituents drawn as fragments (products 14k and 14l): the Text-based R-group Substitution Agent reads the names, the Molecular Recognition Agent the fragments and the Text Extraction Agent the footnotes. (\textbf{C}) A condition table with drawn catalysts (NHC1--NHC3), read by the Molecular Recognition Agent; no substrate varies, so no R-group agent is needed. (\textbf{D}) A table of alternative conditions in which nothing drawn varies. (\textbf{E}) A patent page (EP3986873B1, page 3) with a three-step generic scheme; the Molecular Recognition Agent resolves the groups defined below it, and the Text Extraction Agent reads the reagents and conditions. (\textbf{F}) A patent page (EP2935307B1, page 1) with a Markush formula whose listed groups each give one compound, read by the Molecular Recognition Agent. The Data Structure Agent, which follows every plan, is not drawn. All graphics are from the benchmark~\cite{liang2015dabco,liu2008stetter,kim2020deaminative,cheng2023amino}.}
\end{figure}

\setcounter{figure}{2}
\begin{figure}[p]
\centering
\includegraphics[width=\textwidth,height=0.78\textheight,keepaspectratio]{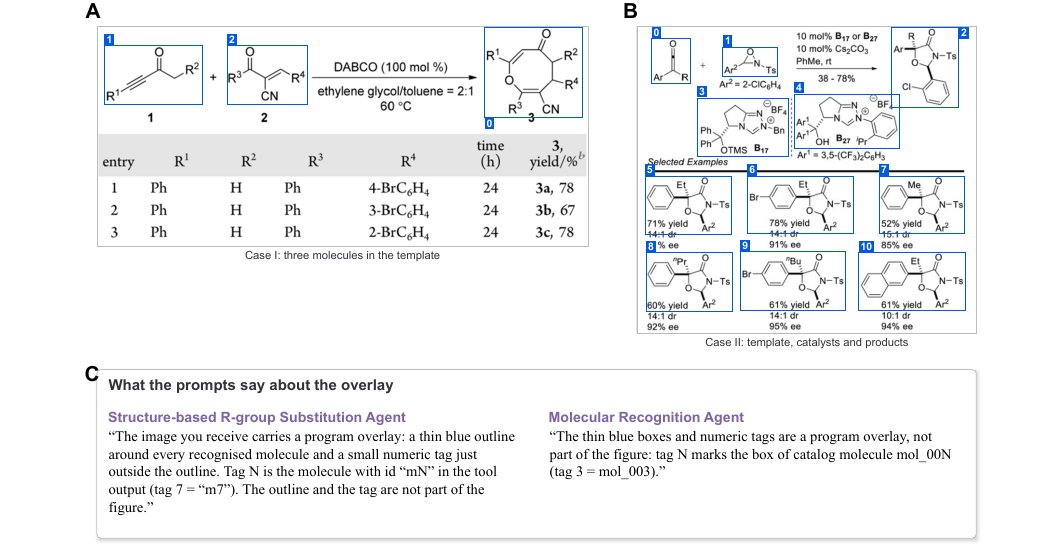}
\caption{\textbf{The visual prompt and how the prompts describe it.} (\textbf{A} and \textbf{B}) The overlay that the program draws on Case I (\textbf{A}; the three molecules of the template) and on Case II (\textbf{B}; template, catalysts B17 and B27 and product variants, 11 molecules) before the graphic reaches the model. Each molecule found by the detection and recognition tools is outlined with a thin blue box and tagged with a number just outside it. Tag N denotes molecule mN in the tool output given to the R-group substitution agents and molecule mol\_N, written with three digits (mol\_004), in the catalog given to the Molecular Recognition and Reaction Template Parsing Agents. Boxes and tags are redrawn as vectors for legibility at print size, at the positions chosen by the program (Materials and Methods). (\textbf{C}) The sentences that state this correspondence in the prompts of the Structure-based R-group Substitution Agent and of the Molecular Recognition Agent, quoted verbatim; the prompt of the Reaction Template Parsing Agent states the same correspondence. The models answer in terms of these identifiers and do not write SMILES strings for recognized molecules.}
\end{figure}

\setcounter{figure}{3}
\begin{figure}[p]
\centering
\includegraphics[width=\linewidth,height=0.78\textheight,keepaspectratio]{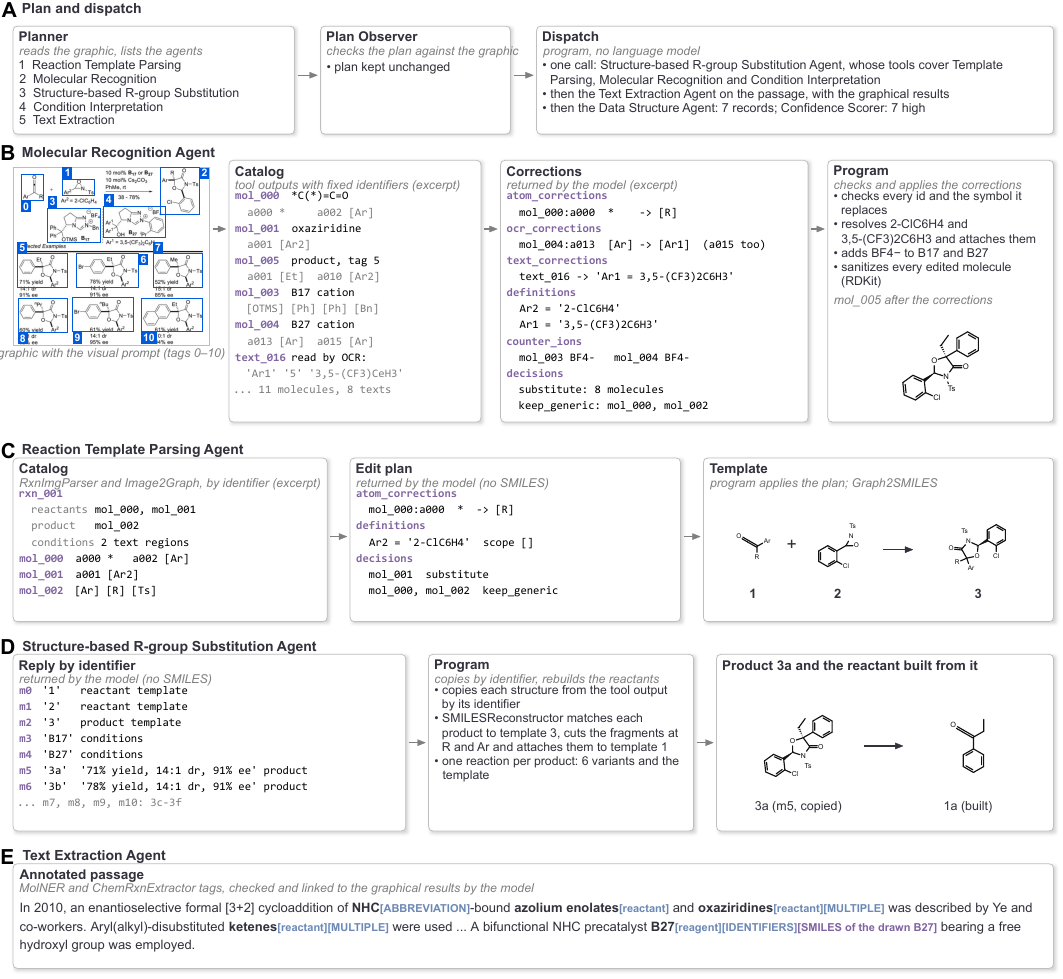}
\caption{\textbf{The agents traced on Case II.} One run of the released code with Gemini 3.7 Flash as base model and
both observers. (\textbf{A}) The plan, the Plan Observer's check and the dispatch. (\textbf{B}) Molecular Recognition
Agent: the graphic with the visual prompt (tag 4 marks molecule \texttt{mol\_004}), excerpts of the catalog and of the corrections returned by the model, and what the program does with them; the
structure is \texttt{mol\_005} after the corrections. (\textbf{C}) Reaction Template Parsing Agent: the catalog of the parsed reaction, the edit plan returned by the model and the resulting template. (\textbf{D}) Structure-based R-group Substitution
Agent: excerpt of the reply by identifier, the program steps, and product 3a with the reactant 1a that
SMILESReconstructor built from it. (\textbf{E}) The passage annotated by the Text Extraction Agent; tags give entity
classes and reaction roles (blue) and the structure taken from the graphical results (purple, abbreviated here).
Monospace text is quoted from the run, with the molecules numbered as in Fig.~S3, and ``...'' marks omissions. The corrections in \textbf{B} come from one further
call of the agent on the same graphic.}
\label{fig:trace2}
\end{figure}
\setcounter{figure}{4}
\begin{figure}[p]
\centering
\includegraphics[width=\linewidth,height=0.78\textheight,keepaspectratio]{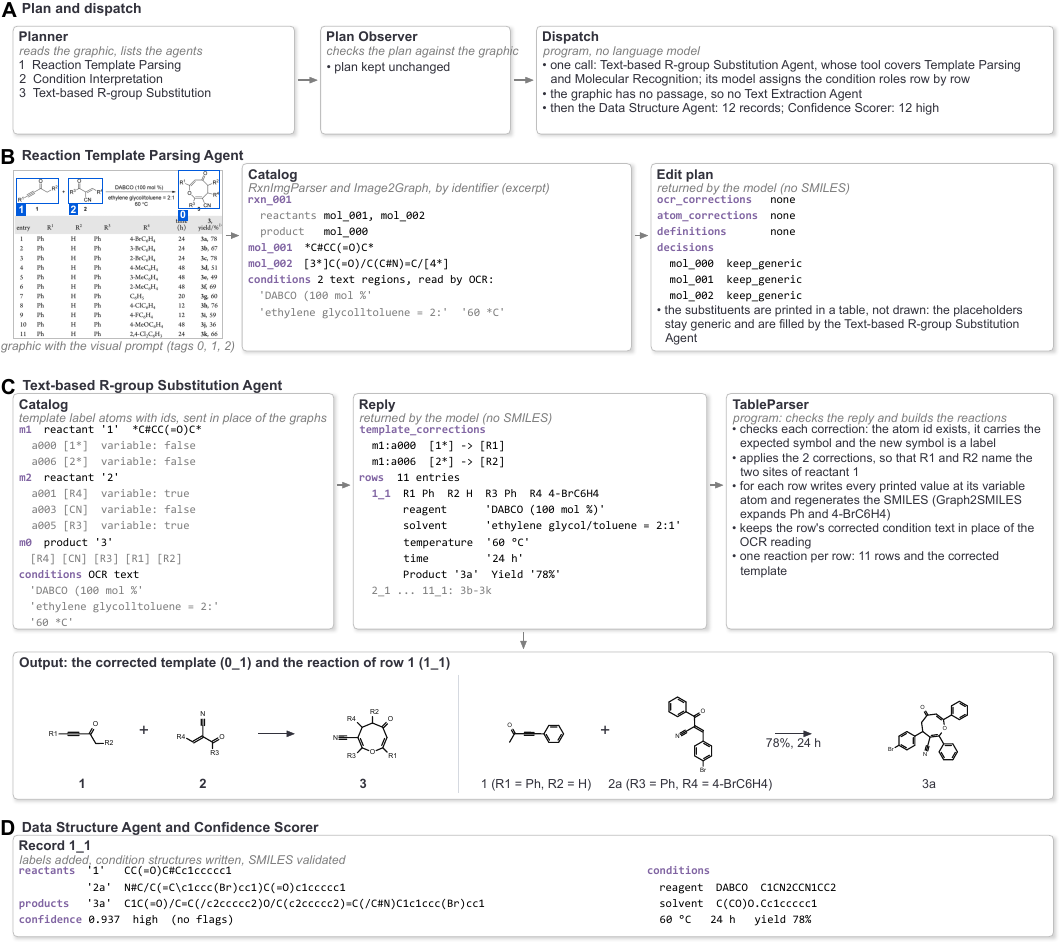}
\caption{\textbf{The agents traced on Case I.} Run as in Fig.~S4. (\textbf{A}) The plan,
the Plan Observer's check and the dispatch; the graphic has no passage, so no Text Extraction Agent is planned.
(\textbf{B}) Reaction Template Parsing Agent: the graphic with the visual prompt, the catalog of the parsed reaction with its condition regions, and the edit plan, which keeps the three template molecules generic. (\textbf{C}) Text-based R-group Substitution Agent: the catalog sent to the model,
an excerpt of its reply (one of eleven rows), the steps of TableParser, and the corrected template with the reaction
built for row 1. (\textbf{D}) The record of row 1 after the Data Structure Agent and the Confidence Scorer.}
\label{fig:trace1}
\end{figure}
\setcounter{figure}{5}
\begin{figure}[p]
\centering
\includegraphics[width=\textwidth,height=0.78\textheight,keepaspectratio]{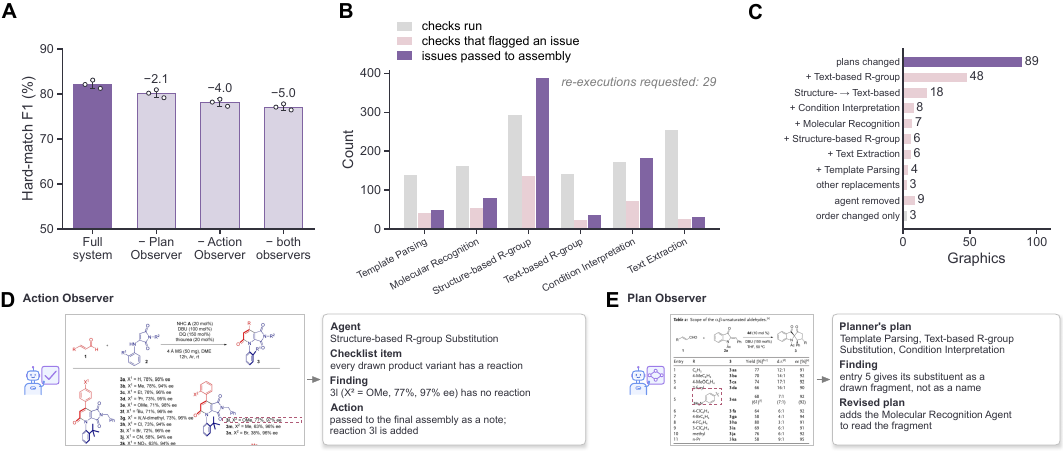}
\caption{\textbf{Contribution and activity of the observers.} (\textbf{A}) Hard-match F1 of the full system and with the Plan Observer, the Action Observer or both switched off (Gemini 3.7 Flash, 424 graphics): means of three runs (bars), SD (error bars), single runs (dots) and difference from the full system (percentage points); P = 0.0497, 0.0048 and 0.0022 (two-sided Welch's t-tests). (\textbf{B}) Action Observer activity per agent in run 1: checks run, checks that flagged an issue and issues passed on as notes; 29 re-executions requested. (\textbf{C}) Plan Observer revisions in run 1: plans whose set of agents changed, by kind of change, and plans whose order alone changed (82 and 94 plans in runs 2 and 3). (\textbf{D}) An Action Observer check on a graphic from~\cite{li2022atroposelective} (boxed): product variant 3l had no reaction, and from this note the final assembly step added it, as in the ground truth. (\textbf{E}) A Plan Observer revision: the Planner assigned this text-based table to the Reaction Template Parsing, Text-based R-group Substitution and Condition Interpretation Agents; the observer found a substituent drawn as a fragment (entry 5, boxed) and added the Molecular Recognition Agent.}
\end{figure}

\setcounter{figure}{6}
\begin{figure}[p]
\centering
\includegraphics[width=\textwidth,height=0.78\textheight,keepaspectratio]{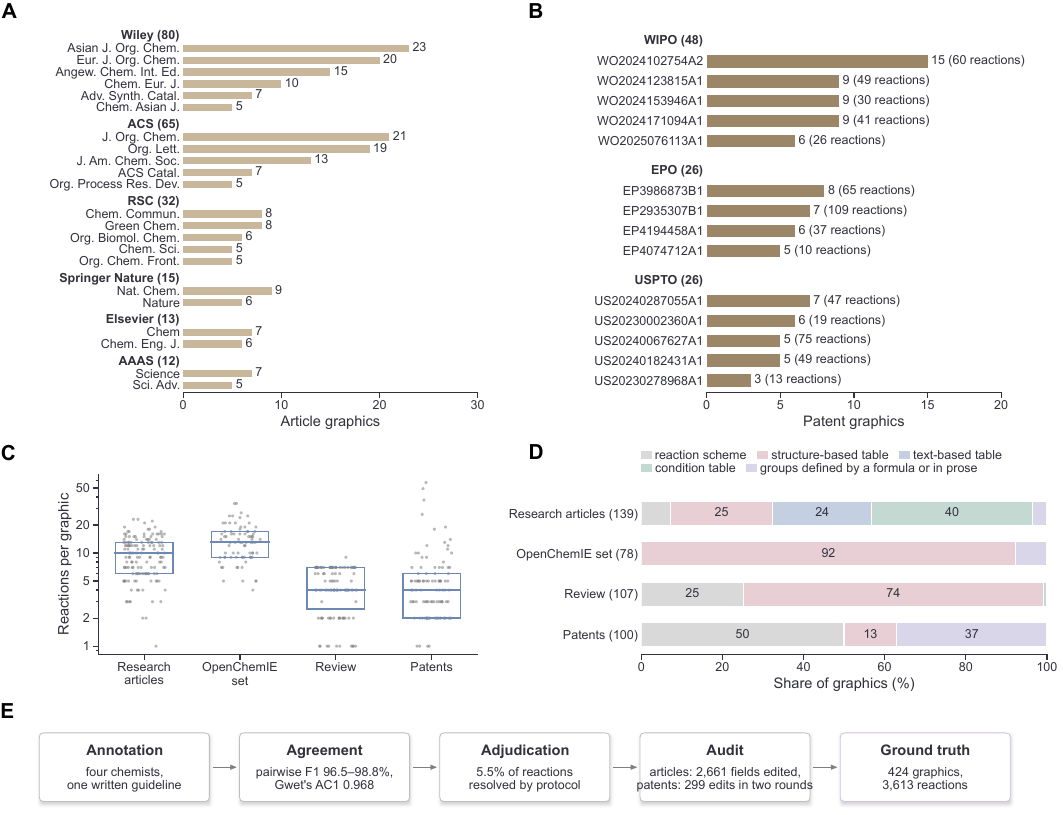}
\caption{\textbf{Composition of the benchmark and construction of the ground truth.} (\textbf{A}) Article graphics per journal, grouped by publisher (totals in parentheses; 217 graphics from 22 journals in the research-article and OpenChemIE subsets; Elsevier includes Cell Press). (\textbf{B}) Patent graphics per patent document, grouped by office, with their reactions (100 graphics, 630 reactions, 14 patents). (\textbf{C}) Reactions per graphic in each subset (log scale): dots are graphics, jittered horizontally; boxes span the interquartile range and lines mark the median (n = 139, 78, 107 and 100). (\textbf{D}) Kinds of graphic (percent of each subset), assigned by inspection to the part that generates the reaction variants: reaction scheme, structure-based table, text-based table (including drawn fragments), condition table, or groups defined by a formula or in prose (including Markush formulas). (\textbf{E}) Construction of the ground truth: independent annotation by four chemists following one guideline; agreement over all 3,613 reactions (pairwise F1 96.5--98.8\%, Gwet's AC1 0.968); adjudication of the 5.5\% of reactions on which they differed; an audit of every reaction against its graphic (2,661 fields changed in the article subsets, 299 entries in the patent subset); and the final ground truth of 424 graphics with 3,613 reactions.}
\end{figure}

\setcounter{figure}{7}
\begin{figure}[p]
\centering
\includegraphics[width=\linewidth,height=0.78\textheight,keepaspectratio]{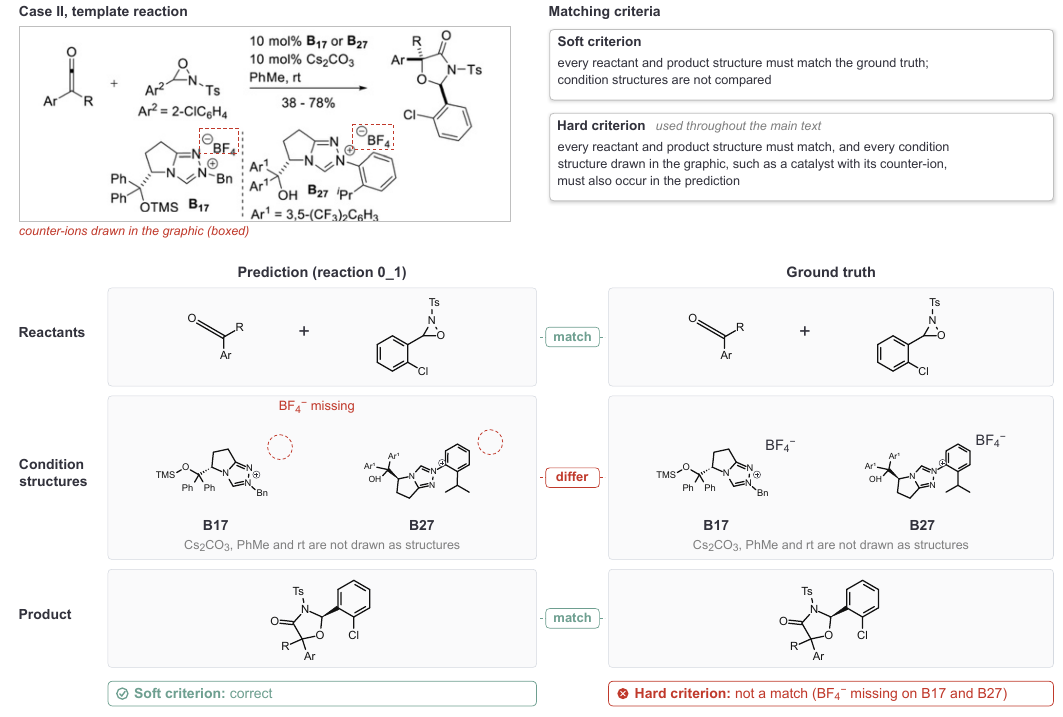}
\caption{\textbf{Hard and soft matching on the template reaction of Case II.} Top left, the template reaction of the
graphic; the tetrafluoroborate counter-ions of the catalysts B17 and B27 are boxed in red. Top right, the two criteria.
Bottom, a prediction and the ground truth compared row by row, with structures drawn from their SMILES (Ts, Ph, Bn,
TMS and Ar\textsuperscript{1} abbreviated as in the graphic). The reactants and the product match, so the reaction is
correct under the soft criterion. The prediction writes B17 and B27 without their counter-ions (red dashed circles);
because every condition structure drawn in the graphic must match under the hard criterion, the reaction is not a hard
match.}
\label{fig:hardsoft}
\end{figure}
\setcounter{figure}{8}
\begin{figure}[p]
\scriptsize\renewcommand{\arraystretch}{1.15}
\begin{tabularx}{\linewidth}{@{}l>{\raggedright\arraybackslash}p{1.9cm}>{\raggedright\arraybackslash}X>{\raggedright\arraybackslash}X>{\raggedright\arraybackslash}p{2.3cm}@{}}
\toprule
ID & Labels & Reactants & Products & Yield; dr; ee \\
\midrule
0\_1 & 1, 2 $\to$ 3 & \texttt{*C(\allowbreak{}*)\allowbreak{}=\allowbreak{}C=\allowbreak{}O} + \texttt{Cc1ccc(\allowbreak{}S(\allowbreak{}=\allowbreak{}O)\allowbreak{}(\allowbreak{}=\allowbreak{}O)\allowbreak{}N2OC2c2ccccc2Cl)\allowbreak{}cc1} & \texttt{*C1(\allowbreak{}*)\allowbreak{}O[C@H]\allowbreak{}(\allowbreak{}c2ccccc2Cl)\allowbreak{}N(\allowbreak{}S(\allowbreak{}=\allowbreak{}O)\allowbreak{}(\allowbreak{}=\allowbreak{}O)\allowbreak{}c2ccc(\allowbreak{}C)\allowbreak{}cc2)\allowbreak{}C1=\allowbreak{}O} & 38 - 78\% \\
1\_1 & 1a, 2 $\to$ 3a & \texttt{CCC(\allowbreak{}=\allowbreak{}C=\allowbreak{}O)\allowbreak{}c1ccccc1} + \texttt{Cc1ccc(\allowbreak{}S(\allowbreak{}=\allowbreak{}O)\allowbreak{}(\allowbreak{}=\allowbreak{}O)\allowbreak{}N2OC2c2ccccc2Cl)\allowbreak{}cc1} & \texttt{CC[C@@]\allowbreak{}1(\allowbreak{}c2ccccc2)\allowbreak{}O[C@H]\allowbreak{}(\allowbreak{}c2ccccc2Cl)\allowbreak{}N(\allowbreak{}S(\allowbreak{}=\allowbreak{}O)\allowbreak{}(\allowbreak{}=\allowbreak{}O)\allowbreak{}c2ccc(\allowbreak{}C)\allowbreak{}cc2)\allowbreak{}C1=\allowbreak{}O} & 71\%; 14:1 dr; 91\% ee \\
2\_1 & 1b, 2 $\to$ 3b & \texttt{CCC(\allowbreak{}=\allowbreak{}C=\allowbreak{}O)\allowbreak{}c1ccc(\allowbreak{}Br)\allowbreak{}cc1} + \texttt{Cc1ccc(\allowbreak{}S(\allowbreak{}=\allowbreak{}O)\allowbreak{}(\allowbreak{}=\allowbreak{}O)\allowbreak{}N2OC2c2ccccc2Cl)\allowbreak{}cc1} & \texttt{CC[C@@]\allowbreak{}1(\allowbreak{}c2ccc(\allowbreak{}Br)\allowbreak{}cc2)\allowbreak{}O[C@H]\allowbreak{}(\allowbreak{}c2ccccc2Cl)\allowbreak{}N(\allowbreak{}S(\allowbreak{}=\allowbreak{}O)\allowbreak{}(\allowbreak{}=\allowbreak{}O)\allowbreak{}c2ccc(\allowbreak{}C)\allowbreak{}cc2)\allowbreak{}C1=\allowbreak{}O} & 78\%; 14:1 dr; 91\% ee \\
3\_1 & 1c, 2 $\to$ 3c & \texttt{CC(\allowbreak{}=\allowbreak{}C=\allowbreak{}O)\allowbreak{}c1ccccc1} + \texttt{Cc1ccc(\allowbreak{}S(\allowbreak{}=\allowbreak{}O)\allowbreak{}(\allowbreak{}=\allowbreak{}O)\allowbreak{}N2OC2c2ccccc2Cl)\allowbreak{}cc1} & \texttt{Cc1ccc(\allowbreak{}S(\allowbreak{}=\allowbreak{}O)\allowbreak{}(\allowbreak{}=\allowbreak{}O)\allowbreak{}N2C(\allowbreak{}=\allowbreak{}O)\allowbreak{}[C@]\allowbreak{}(\allowbreak{}C)\allowbreak{}(\allowbreak{}c3ccccc3)\allowbreak{}O[C@@H]\allowbreak{}2c2ccccc2Cl)\allowbreak{}cc1} & 52\%; 15:1 dr; 85\% ee \\
4\_1 & 1d, 2 $\to$ 3d & \texttt{CCCC(\allowbreak{}=\allowbreak{}C=\allowbreak{}O)\allowbreak{}c1ccccc1} + \texttt{Cc1ccc(\allowbreak{}S(\allowbreak{}=\allowbreak{}O)\allowbreak{}(\allowbreak{}=\allowbreak{}O)\allowbreak{}N2OC2c2ccccc2Cl)\allowbreak{}cc1} & \texttt{CCC[C@@]\allowbreak{}1(\allowbreak{}c2ccccc2)\allowbreak{}O[C@H]\allowbreak{}(\allowbreak{}c2ccccc2Cl)\allowbreak{}N(\allowbreak{}S(\allowbreak{}=\allowbreak{}O)\allowbreak{}(\allowbreak{}=\allowbreak{}O)\allowbreak{}c2ccc(\allowbreak{}C)\allowbreak{}cc2)\allowbreak{}C1=\allowbreak{}O} & 60\%; 14:1 dr; 92\% ee \\
5\_1 & 1e, 2 $\to$ 3e & \texttt{CCCCC(\allowbreak{}=\allowbreak{}C=\allowbreak{}O)\allowbreak{}c1ccc(\allowbreak{}Br)\allowbreak{}cc1} + \texttt{Cc1ccc(\allowbreak{}S(\allowbreak{}=\allowbreak{}O)\allowbreak{}(\allowbreak{}=\allowbreak{}O)\allowbreak{}N2OC2c2ccccc2Cl)\allowbreak{}cc1} & \texttt{CCCC[C@@]\allowbreak{}1(\allowbreak{}c2ccc(\allowbreak{}Br)\allowbreak{}cc2)\allowbreak{}O[C@H]\allowbreak{}(\allowbreak{}c2ccccc2Cl)\allowbreak{}N(\allowbreak{}S(\allowbreak{}=\allowbreak{}O)\allowbreak{}(\allowbreak{}=\allowbreak{}O)\allowbreak{}c2ccc(\allowbreak{}C)\allowbreak{}cc2)\allowbreak{}C1=\allowbreak{}O} & 61\%; 14:1 dr; 95\% ee \\
6\_1 & 1f, 2 $\to$ 3f & \texttt{CCC(\allowbreak{}=\allowbreak{}C=\allowbreak{}O)\allowbreak{}c1ccc2ccccc2c1} + \texttt{Cc1ccc(\allowbreak{}S(\allowbreak{}=\allowbreak{}O)\allowbreak{}(\allowbreak{}=\allowbreak{}O)\allowbreak{}N2OC2c2ccccc2Cl)\allowbreak{}cc1} & \texttt{CC[C@@]\allowbreak{}1(\allowbreak{}c2ccc3ccccc3c2)\allowbreak{}O[C@H]\allowbreak{}(\allowbreak{}c2ccccc2Cl)\allowbreak{}N(\allowbreak{}S(\allowbreak{}=\allowbreak{}O)\allowbreak{}(\allowbreak{}=\allowbreak{}O)\allowbreak{}c2ccc(\allowbreak{}C)\allowbreak{}cc2)\allowbreak{}C1=\allowbreak{}O} & 61\%; 10:1 dr; 94\% ee \\
\bottomrule
\end{tabularx}
\par\vspace{6pt}
\begin{tabularx}{\linewidth}{@{}l>{\raggedright\arraybackslash}p{2.8cm}l>{\raggedright\arraybackslash}X@{}}
\toprule
Role & Text & Label & SMILES \\
\midrule
reagent &  & B17 & \texttt{C[Si]\allowbreak{}(\allowbreak{}C)\allowbreak{}(\allowbreak{}C)\allowbreak{}OC(\allowbreak{}c1ccccc1)\allowbreak{}(\allowbreak{}c1ccccc1)\allowbreak{}[C@@H]\allowbreak{}1CCc2n[n+]\allowbreak{}(\allowbreak{}Cc3ccccc3)\allowbreak{}cn21.\allowbreak{}F[B-]\allowbreak{}(\allowbreak{}F)\allowbreak{}(\allowbreak{}F)\allowbreak{}F} \\
reagent &  & B27 & \texttt{CC(\allowbreak{}C)\allowbreak{}c1ccccc1-[n+]\allowbreak{}1cn2c(\allowbreak{}n1)\allowbreak{}CC[C@H]\allowbreak{}2C(\allowbreak{}O)\allowbreak{}(\allowbreak{}c1cc(\allowbreak{}C(\allowbreak{}F)\allowbreak{}(\allowbreak{}F)\allowbreak{}F)\allowbreak{}cc(\allowbreak{}C(\allowbreak{}F)\allowbreak{}(\allowbreak{}F)\allowbreak{}F)\allowbreak{}c1)\allowbreak{}c1cc(\allowbreak{}C(\allowbreak{}F)\allowbreak{}(\allowbreak{}F)\allowbreak{}F)\allowbreak{}cc(\allowbreak{}C(\allowbreak{}F)\allowbreak{}(\allowbreak{}F)\allowbreak{}F)\allowbreak{}c1.\allowbreak{}F[B-]\allowbreak{}(\allowbreak{}F)\allowbreak{}(\allowbreak{}F)\allowbreak{}F} \\
reagent & 10 mol\% Cs2CO3 &  & \texttt{[Cs+]\allowbreak{}.\allowbreak{}[Cs+]\allowbreak{}.\allowbreak{}[O-]\allowbreak{}C(\allowbreak{}[O-]\allowbreak{})\allowbreak{}=\allowbreak{}O} \\
solvent & PhMe &  & \texttt{Cc1ccccc1} \\
temperature & rt &  &  \\
reagent & 10 mol\% B17 or B27 & B17 & \texttt{C[Si]\allowbreak{}(\allowbreak{}C)\allowbreak{}(\allowbreak{}C)\allowbreak{}OC(\allowbreak{}c1ccccc1)\allowbreak{}(\allowbreak{}c1ccccc1)\allowbreak{}[C@@H]\allowbreak{}1CCc2n[n+]\allowbreak{}(\allowbreak{}Cc3ccccc3)\allowbreak{}cn21.\allowbreak{}F[B-]\allowbreak{}(\allowbreak{}F)\allowbreak{}(\allowbreak{}F)\allowbreak{}F} \\
reagent & 10 mol\% B17 or B27 & B27 & \texttt{CC(\allowbreak{}C)\allowbreak{}c1ccccc1-[n+]\allowbreak{}1cn2c(\allowbreak{}n1)\allowbreak{}CC[C@H]\allowbreak{}2C(\allowbreak{}O)\allowbreak{}(\allowbreak{}c1cc(\allowbreak{}C(\allowbreak{}F)\allowbreak{}(\allowbreak{}F)\allowbreak{}F)\allowbreak{}cc(\allowbreak{}C(\allowbreak{}F)\allowbreak{}(\allowbreak{}F)\allowbreak{}F)\allowbreak{}c1)\allowbreak{}c1cc(\allowbreak{}C(\allowbreak{}F)\allowbreak{}(\allowbreak{}F)\allowbreak{}F)\allowbreak{}cc(\allowbreak{}C(\allowbreak{}F)\allowbreak{}(\allowbreak{}F)\allowbreak{}F)\allowbreak{}c1.\allowbreak{}F[B-]\allowbreak{}(\allowbreak{}F)\allowbreak{}(\allowbreak{}F)\allowbreak{}F} \\
\bottomrule
\end{tabularx}
\par\vspace{6pt}
\begin{tabularx}{\linewidth}{@{}>{\raggedright\arraybackslash}X@{}}
\toprule
Annotated text (Text Extraction Agent) \\
\midrule
\texttt{In 2010, an enantioselective formal [3+2]\allowbreak{} cycloaddition of NHC-bound azolium enolates [reactant]\allowbreak{}[MULTIPLE]\allowbreak{} and oxaziridines [reactant]\allowbreak{}[SYSTEMATIC]\allowbreak{} was described by Ye and co-workers.\allowbreak{} Aryl(\allowbreak{}alkyl)\allowbreak{}-disubstituted ketenes were used as precursors of azolium enolates [reactant]\allowbreak{}.\allowbreak{} A bifunctional NHC [ABBREVIATION]\allowbreak{} precatalyst B27 [reagent]\allowbreak{} [IDENTIFIERS]\allowbreak{}[SMILES:CCCC(\allowbreak{}C=\allowbreak{}CC=\allowbreak{}C1)\allowbreak{}=\allowbreak{}C1[N+]\allowbreak{}2=\allowbreak{}CN3[C@H]\allowbreak{}(\allowbreak{}C(\allowbreak{}C1=\allowbreak{}CC(\allowbreak{}=\allowbreak{}CC(\allowbreak{}=\allowbreak{}C1C(\allowbreak{}F)\allowbreak{}(\allowbreak{}F)\allowbreak{}F)\allowbreak{}C(\allowbreak{}F)\allowbreak{}(\allowbreak{}F)\allowbreak{}F)\allowbreak{})\allowbreak{}(\allowbreak{}C1=\allowbreak{}CC(\allowbreak{}=\allowbreak{}CC(\allowbreak{}=\allowbreak{}C1C(\allowbreak{}F)\allowbreak{}(\allowbreak{}F)\allowbreak{}F)\allowbreak{}C(\allowbreak{}F)\allowbreak{}(\allowbreak{}F)\allowbreak{}F)\allowbreak{})\allowbreak{}O)\allowbreak{}CCC3=\allowbreak{}N2.\allowbreak{}F[B-]\allowbreak{}(\allowbreak{}F)\allowbreak{}(\allowbreak{}F)\allowbreak{}F]\allowbreak{} bearing a free hydroxyl group was employed.\allowbreak{}} \\
\bottomrule
\end{tabularx}

\caption{\textbf{Complete output of ChemEAGLE on Case II.} Top, each reaction with its labels, reactant and product
SMILES, yield, diastereomeric ratio and enantiomeric excess. Middle, the condition entries, which are the same in
every reaction (each with its own yield), listed once; the template reaction gives B17 and B27 without the condition
text. Bottom, the passage as annotated by the Text Extraction Agent; the SMILES that the model attached there to B27 is an isomer of the drawn catalyst, whereas the records carry the drawn structure. All seven reactions are present, and every
reactant, product, R-group substitution and condition structure matches the ground truth.}
\label{fig:out-ce}
\end{figure}
\setcounter{figure}{9}
\begin{figure}[p]
\scriptsize\renewcommand{\arraystretch}{1.15}
\begin{tabularx}{\linewidth}{@{}l>{\raggedright\arraybackslash}p{1.9cm}>{\raggedright\arraybackslash}X>{\raggedright\arraybackslash}X>{\raggedright\arraybackslash}p{2.3cm}@{}}
\toprule
ID & Labels & Reactants & Products & Yield; dr; ee \\
\midrule
0\_1 & 1, 2 $\to$ 3 & \texttt{O=\allowbreak{}C=\allowbreak{}C(\allowbreak{}Ar)\allowbreak{}R} + \texttt{O1C(\allowbreak{}N1S(\allowbreak{}=\allowbreak{}O)\allowbreak{}(\allowbreak{}=\allowbreak{}O)\allowbreak{}c2ccccc2)\allowbreak{}Ar2} & \texttt{O=\allowbreak{}C1C(\allowbreak{}OC(\allowbreak{}N1S(\allowbreak{}=\allowbreak{}O)\allowbreak{}(\allowbreak{}=\allowbreak{}O)\allowbreak{}c2ccccc2)\allowbreak{}Ar2)\allowbreak{}(\allowbreak{}R)\allowbreak{}Ar} & 38 - 78\% \\
1\_1 & 1a, 2a $\to$ 3a & \texttt{O=\allowbreak{}C=\allowbreak{}C(\allowbreak{}c1ccccc1)\allowbreak{}CC} + \texttt{O1C(\allowbreak{}N1S(\allowbreak{}=\allowbreak{}O)\allowbreak{}(\allowbreak{}=\allowbreak{}O)\allowbreak{}c2ccccc2)\allowbreak{}c3ccccc3Cl} & \texttt{O=\allowbreak{}C1[C@]\allowbreak{}(\allowbreak{}CC)\allowbreak{}(\allowbreak{}OC(\allowbreak{}[C@H]\allowbreak{}1N1S(\allowbreak{}=\allowbreak{}O)\allowbreak{}(\allowbreak{}=\allowbreak{}O)\allowbreak{}c2ccccc2)\allowbreak{}c3ccccc3Cl)\allowbreak{}c4ccccc4} & 71\% yield; 14:1 dr; 91\% ee \\
2\_1 & 1b, 2a $\to$ 3b & \texttt{O=\allowbreak{}C=\allowbreak{}C(\allowbreak{}c1ccc(\allowbreak{}cc1)\allowbreak{}Br)\allowbreak{}CC} + \texttt{O1C(\allowbreak{}N1S(\allowbreak{}=\allowbreak{}O)\allowbreak{}(\allowbreak{}=\allowbreak{}O)\allowbreak{}c2ccccc2)\allowbreak{}c3ccccc3Cl} & \texttt{O=\allowbreak{}C1[C@]\allowbreak{}(\allowbreak{}CC)\allowbreak{}(\allowbreak{}OC(\allowbreak{}[C@H]\allowbreak{}1N1S(\allowbreak{}=\allowbreak{}O)\allowbreak{}(\allowbreak{}=\allowbreak{}O)\allowbreak{}c2ccccc2)\allowbreak{}c3ccccc3Cl)\allowbreak{}c4ccc(\allowbreak{}cc4)\allowbreak{}Br} & 78\% yield; 14:1 dr; 91\% ee \\
3\_1 & 1c, 2a $\to$ 3c & \texttt{O=\allowbreak{}C=\allowbreak{}C(\allowbreak{}c1ccccc1)\allowbreak{}C} + \texttt{O1C(\allowbreak{}N1S(\allowbreak{}=\allowbreak{}O)\allowbreak{}(\allowbreak{}=\allowbreak{}O)\allowbreak{}c2ccccc2)\allowbreak{}c3ccccc3Cl} & \texttt{O=\allowbreak{}C1[C@]\allowbreak{}(\allowbreak{}C)\allowbreak{}(\allowbreak{}OC(\allowbreak{}[C@H]\allowbreak{}1N1S(\allowbreak{}=\allowbreak{}O)\allowbreak{}(\allowbreak{}=\allowbreak{}O)\allowbreak{}c2ccccc2)\allowbreak{}c3ccccc3Cl)\allowbreak{}c4ccccc4} & 52\% yield; 15:1 dr; 85\% ee \\
4\_1 & 1d, 2a $\to$ 3d & \texttt{O=\allowbreak{}C=\allowbreak{}C(\allowbreak{}c1ccccc1)\allowbreak{}CCC} + \texttt{O1C(\allowbreak{}N1S(\allowbreak{}=\allowbreak{}O)\allowbreak{}(\allowbreak{}=\allowbreak{}O)\allowbreak{}c2ccccc2)\allowbreak{}c3ccccc3Cl} & \texttt{O=\allowbreak{}C1[C@]\allowbreak{}(\allowbreak{}CCC)\allowbreak{}(\allowbreak{}OC(\allowbreak{}[C@H]\allowbreak{}1N1S(\allowbreak{}=\allowbreak{}O)\allowbreak{}(\allowbreak{}=\allowbreak{}O)\allowbreak{}c2ccccc2)\allowbreak{}c3ccccc3Cl)\allowbreak{}c4ccccc4} & 60\% yield; 14:1 dr; 92\% ee \\
5\_1 & 1e, 2a $\to$ 3e & \texttt{O=\allowbreak{}C=\allowbreak{}C(\allowbreak{}c1ccc(\allowbreak{}cc1)\allowbreak{}Br)\allowbreak{}CCCC} + \texttt{O1C(\allowbreak{}N1S(\allowbreak{}=\allowbreak{}O)\allowbreak{}(\allowbreak{}=\allowbreak{}O)\allowbreak{}c2ccccc2)\allowbreak{}c3ccccc3Cl} & \texttt{O=\allowbreak{}C1[C@]\allowbreak{}(\allowbreak{}CCCC)\allowbreak{}(\allowbreak{}OC(\allowbreak{}[C@H]\allowbreak{}1N1S(\allowbreak{}=\allowbreak{}O)\allowbreak{}(\allowbreak{}=\allowbreak{}O)\allowbreak{}c2ccccc2)\allowbreak{}c3ccccc3Cl)\allowbreak{}c4ccc(\allowbreak{}cc4)\allowbreak{}Br} & 61\% yield; 14:1 dr; 95\% ee \\
6\_1 & 1f, 2a $\to$ 3f & \texttt{O=\allowbreak{}C=\allowbreak{}C(\allowbreak{}c1ccc2ccccc2c1)\allowbreak{}CC} + \texttt{O1C(\allowbreak{}N1S(\allowbreak{}=\allowbreak{}O)\allowbreak{}(\allowbreak{}=\allowbreak{}O)\allowbreak{}c2ccccc2)\allowbreak{}c3ccccc3Cl} & \texttt{O=\allowbreak{}C1[C@]\allowbreak{}(\allowbreak{}CC)\allowbreak{}(\allowbreak{}OC(\allowbreak{}[C@H]\allowbreak{}1N1S(\allowbreak{}=\allowbreak{}O)\allowbreak{}(\allowbreak{}=\allowbreak{}O)\allowbreak{}c2ccccc2)\allowbreak{}c3ccccc3Cl)\allowbreak{}c4ccc5ccccc5c4} & 61\% yield; 10:1 dr; 94\% ee \\
\bottomrule
\end{tabularx}
\par\vspace{6pt}
\begin{tabularx}{\linewidth}{@{}l>{\raggedright\arraybackslash}p{2.8cm}l>{\raggedright\arraybackslash}X@{}}
\toprule
Role & Text & Label & SMILES \\
\midrule
reagent & 10 mol\% B17 or B27 & B17 & \texttt{CN1CCCC1=\allowbreak{}NC2=\allowbreak{}CN(\allowbreak{}C=\allowbreak{}[N+]\allowbreak{}2[C@]\allowbreak{}3(\allowbreak{}C4=\allowbreak{}CC=\allowbreak{}CC=\allowbreak{}C4)\allowbreak{}C5=\allowbreak{}CC=\allowbreak{}CC=\allowbreak{}C5OTMS)\allowbreak{}C6=\allowbreak{}CC=\allowbreak{}CC=\allowbreak{}C6.\allowbreak{}[BF4-]\allowbreak{}} \\
reagent & 10 mol\% B17 or B27 & B27 & \texttt{CN1CCCC1=\allowbreak{}NC2=\allowbreak{}CN(\allowbreak{}C=\allowbreak{}[N+]\allowbreak{}2C3=\allowbreak{}CC=\allowbreak{}CC=\allowbreak{}C3[C@H]\allowbreak{}(\allowbreak{}C)\allowbreak{}C)\allowbreak{}C4=\allowbreak{}C(\allowbreak{}C=\allowbreak{}CC=\allowbreak{}C4)\allowbreak{}C(\allowbreak{}C(\allowbreak{}F)\allowbreak{}(\allowbreak{}F)\allowbreak{}F)\allowbreak{}(\allowbreak{}C(\allowbreak{}F)\allowbreak{}(\allowbreak{}F)\allowbreak{}F)\allowbreak{}C5=\allowbreak{}CC(\allowbreak{}=\allowbreak{}CC(\allowbreak{}=\allowbreak{}C5)\allowbreak{}C(\allowbreak{}F)\allowbreak{}(\allowbreak{}F)\allowbreak{}F)\allowbreak{}C(\allowbreak{}F)\allowbreak{}(\allowbreak{}F)\allowbreak{}F.\allowbreak{}[BF4-]\allowbreak{}} \\
reagent & 10 mol\% Cs2CO3 & Cs2CO3 & \texttt{C(\allowbreak{}=\allowbreak{}O)\allowbreak{}(\allowbreak{}[O-]\allowbreak{})\allowbreak{}[O-]\allowbreak{}.\allowbreak{}[Cs+]\allowbreak{}.\allowbreak{}[Cs+]\allowbreak{}} \\
solvent & PhMe &  & \texttt{Cc1ccccc1} \\
temperature & rt &  &  \\
reagent & 10 mol\% B17 or B27 & B17 & \texttt{CN1CCCC1=\allowbreak{}NC2=\allowbreak{}CN(\allowbreak{}C=\allowbreak{}[N+]\allowbreak{}2[C@]\allowbreak{}3(\allowbreak{}C4=\allowbreak{}CC=\allowbreak{}CC=\allowbreak{}C4)\allowbreak{}C5=\allowbreak{}CC=\allowbreak{}CC=\allowbreak{}C4OTMS)\allowbreak{}C6=\allowbreak{}CC=\allowbreak{}CC=\allowbreak{}C6.\allowbreak{}[BF4-]\allowbreak{}} \\
\bottomrule
\end{tabularx}

\caption{\textbf{Complete output of Gemini 3.7 Flash used end to end on Case II.} Top, each returned reaction with its
labels, reactant and product SMILES as returned, yield and additional information. Bottom, the condition entries,
which the model repeated in every reaction (each with its own yield), listed once; the two SMILES for B17 differ in a
ring-closure digit. The complete JSON is released with the data. All seven reactions are present with the correct
yields, diastereomeric ratios and enantiomeric excesses, and the six ketenes are correct. The oxaziridine carries a benzenesulfonyl group in place of Ts, none of the
six product SMILES can be parsed (a ring bond opened at the nitrogen is never closed), the template writes Ar and R as
atoms, and the SMILES of the catalysts B17 and B27 are invalid. No reaction matches the ground truth.}
\label{fig:out-gem}
\end{figure}
\setcounter{figure}{10}
\begin{figure}[p]
\scriptsize\renewcommand{\arraystretch}{1.15}
\begin{tabularx}{\linewidth}{@{}l>{\raggedright\arraybackslash}p{1.9cm}>{\raggedright\arraybackslash}X>{\raggedright\arraybackslash}X>{\raggedright\arraybackslash}p{2.3cm}@{}}
\toprule
ID & Labels & Reactants & Products & Yield; dr; ee \\
\midrule
0\_1 & --, -- $\to$ -- & \texttt{O=\allowbreak{}C=\allowbreak{}C(\allowbreak{}*)\allowbreak{}*} + \texttt{O1C(\allowbreak{}N1S(\allowbreak{}=\allowbreak{}O)\allowbreak{}(\allowbreak{}=\allowbreak{}O)\allowbreak{}c1ccc(\allowbreak{}C)\allowbreak{}cc1)\allowbreak{}Ar2} & \texttt{Clc1ccccc1[C@H]\allowbreak{}1O[C@]\allowbreak{}(\allowbreak{}Ar)\allowbreak{}(\allowbreak{}R)\allowbreak{}C(\allowbreak{}=\allowbreak{}O)\allowbreak{}N1S(\allowbreak{}=\allowbreak{}O)\allowbreak{}(\allowbreak{}=\allowbreak{}O)\allowbreak{}c1ccc(\allowbreak{}C)\allowbreak{}cc1} & 38 - 78\% \\
1\_1 & --, -- $\to$ -- & \texttt{O=\allowbreak{}C=\allowbreak{}C(\allowbreak{}*)\allowbreak{}*} + \texttt{O1C(\allowbreak{}c2ccccc2Cl)\allowbreak{}N1S(\allowbreak{}=\allowbreak{}O)\allowbreak{}(\allowbreak{}=\allowbreak{}O)\allowbreak{}c1ccc(\allowbreak{}C)\allowbreak{}cc1} & \texttt{CC[C@@]\allowbreak{}1(\allowbreak{}c2ccccc2)\allowbreak{}O[C@H]\allowbreak{}(\allowbreak{}c2ccccc2Cl)\allowbreak{}N(\allowbreak{}S(\allowbreak{}=\allowbreak{}O)\allowbreak{}(\allowbreak{}=\allowbreak{}O)\allowbreak{}c2ccc(\allowbreak{}C)\allowbreak{}cc2)\allowbreak{}C1=\allowbreak{}O} & 71\% yield; 14:1 dr; 91\% ee \\
1\_2 & --, -- $\to$ -- & \texttt{O=\allowbreak{}C=\allowbreak{}C(\allowbreak{}*)\allowbreak{}*} + \texttt{O1C(\allowbreak{}c2ccccc2Cl)\allowbreak{}N1S(\allowbreak{}=\allowbreak{}O)\allowbreak{}(\allowbreak{}=\allowbreak{}O)\allowbreak{}c1ccc(\allowbreak{}C)\allowbreak{}cc1} & \texttt{CC[C@@]\allowbreak{}1(\allowbreak{}c2ccc(\allowbreak{}Br)\allowbreak{}cc2)\allowbreak{}O[C@H]\allowbreak{}(\allowbreak{}c2ccccc2Cl)\allowbreak{}N(\allowbreak{}S(\allowbreak{}=\allowbreak{}O)\allowbreak{}(\allowbreak{}=\allowbreak{}O)\allowbreak{}c2ccc(\allowbreak{}C)\allowbreak{}cc2)\allowbreak{}C1=\allowbreak{}O} & 78\% yield; 14:1 dr; 91\% ee \\
1\_3 & --, -- $\to$ -- & \texttt{O=\allowbreak{}C=\allowbreak{}C(\allowbreak{}*)\allowbreak{}*} + \texttt{O1C(\allowbreak{}c2ccccc2Cl)\allowbreak{}N1S(\allowbreak{}=\allowbreak{}O)\allowbreak{}(\allowbreak{}=\allowbreak{}O)\allowbreak{}c1ccc(\allowbreak{}C)\allowbreak{}cc1} & \texttt{C[C@@]\allowbreak{}1(\allowbreak{}c2ccccc2)\allowbreak{}O[C@@H]\allowbreak{}(\allowbreak{}c2ccccc2Cl)\allowbreak{}N(\allowbreak{}S(\allowbreak{}=\allowbreak{}O)\allowbreak{}(\allowbreak{}=\allowbreak{}O)\allowbreak{}c2ccc(\allowbreak{}C)\allowbreak{}cc2)\allowbreak{}C1=\allowbreak{}O} & 52\% yield; 15:1 dr; 85\% ee \\
2\_1 & --, -- $\to$ -- & \texttt{O=\allowbreak{}C=\allowbreak{}C(\allowbreak{}*)\allowbreak{}*} + \texttt{O1C(\allowbreak{}c2ccccc2Cl)\allowbreak{}N1S(\allowbreak{}=\allowbreak{}O)\allowbreak{}(\allowbreak{}=\allowbreak{}O)\allowbreak{}c1ccc(\allowbreak{}C)\allowbreak{}cc1} & \texttt{CCC[C@@]\allowbreak{}1(\allowbreak{}c2ccccc2)\allowbreak{}O[C@@H]\allowbreak{}(\allowbreak{}c2ccccc2Cl)\allowbreak{}N(\allowbreak{}S(\allowbreak{}=\allowbreak{}O)\allowbreak{}(\allowbreak{}=\allowbreak{}O)\allowbreak{}c2ccc(\allowbreak{}C)\allowbreak{}cc2)\allowbreak{}C1=\allowbreak{}O} & 60\% yield; 14:1 dr; 92\% ee \\
2\_2 & --, -- $\to$ -- & \texttt{O=\allowbreak{}C=\allowbreak{}C(\allowbreak{}*)\allowbreak{}*} + \texttt{O1C(\allowbreak{}c2ccccc2Cl)\allowbreak{}N1S(\allowbreak{}=\allowbreak{}O)\allowbreak{}(\allowbreak{}=\allowbreak{}O)\allowbreak{}c1ccc(\allowbreak{}C)\allowbreak{}cc1} & \texttt{CCCC[C@@]\allowbreak{}1(\allowbreak{}c2ccc(\allowbreak{}Br)\allowbreak{}cc2)\allowbreak{}O[C@H]\allowbreak{}(\allowbreak{}c2ccccc2Cl)\allowbreak{}N(\allowbreak{}S(\allowbreak{}=\allowbreak{}O)\allowbreak{}(\allowbreak{}=\allowbreak{}O)\allowbreak{}c2ccc(\allowbreak{}C)\allowbreak{}cc2)\allowbreak{}C1=\allowbreak{}O} & 61\% yield; 14:1 dr; 95\% ee \\
2\_3 & --, -- $\to$ -- & \texttt{O=\allowbreak{}C=\allowbreak{}C(\allowbreak{}*)\allowbreak{}*} + \texttt{O1C(\allowbreak{}c2ccccc2Cl)\allowbreak{}N1S(\allowbreak{}=\allowbreak{}O)\allowbreak{}(\allowbreak{}=\allowbreak{}O)\allowbreak{}c1ccc(\allowbreak{}C)\allowbreak{}cc1} & \texttt{CC[C@@]\allowbreak{}1(\allowbreak{}c2ccc3ccccc3c2)\allowbreak{}O[C@@H]\allowbreak{}(\allowbreak{}c2ccccc2Cl)\allowbreak{}N(\allowbreak{}S(\allowbreak{}=\allowbreak{}O)\allowbreak{}(\allowbreak{}=\allowbreak{}O)\allowbreak{}c2ccc(\allowbreak{}C)\allowbreak{}cc2)\allowbreak{}C1=\allowbreak{}O} & 61\% yield; 10:1 dr; 94\% ee \\
\bottomrule
\end{tabularx}
\par\vspace{6pt}
\begin{tabularx}{\linewidth}{@{}l>{\raggedright\arraybackslash}p{2.8cm}l>{\raggedright\arraybackslash}X@{}}
\toprule
Role & Text & Label & SMILES \\
\midrule
reagent & 10 mol\% B17 & B17 & \texttt{C[C@]\allowbreak{}(\allowbreak{}c1ccccc1)\allowbreak{}(\allowbreak{}c1ccccc1)\allowbreak{}O[Si]\allowbreak{}(\allowbreak{}C)\allowbreak{}(\allowbreak{}C)\allowbreak{}C} \\
reagent & 10 mol\% B27 & B27 & \texttt{CC(\allowbreak{}C)\allowbreak{}c1ccccc1N1C=\allowbreak{}N[N+]\allowbreak{}2=\allowbreak{}C1CCC[C@@H]\allowbreak{}2C(\allowbreak{}O)\allowbreak{}(\allowbreak{}c1cc(\allowbreak{}C(\allowbreak{}F)\allowbreak{}(\allowbreak{}F)\allowbreak{}F)\allowbreak{}cc(\allowbreak{}C(\allowbreak{}F)\allowbreak{}(\allowbreak{}F)\allowbreak{}F)\allowbreak{}c1)\allowbreak{}c1cc(\allowbreak{}C(\allowbreak{}F)\allowbreak{}(\allowbreak{}F)\allowbreak{}F)\allowbreak{}cc(\allowbreak{}C(\allowbreak{}F)\allowbreak{}(\allowbreak{}F)\allowbreak{}F)\allowbreak{}c1} \\
reagent & 10 mol\% Cs2CO3 &  & \texttt{[Cs+]\allowbreak{}.\allowbreak{}[Cs+]\allowbreak{}.\allowbreak{}[O-]\allowbreak{}C(\allowbreak{}=\allowbreak{}O)\allowbreak{}[O-]\allowbreak{}} \\
solvent & PhMe &  & \texttt{Cc1ccccc1} \\
temperature & rt &  &  \\
\bottomrule
\end{tabularx}

\caption{\textbf{Complete output of Claude Code (Claude Sonnet 5) on Case II.} Layout as in Fig.~S10; the agent gave no compound labels (--). All seven reactions are present with the correct yields,
diastereomeric ratios and enantiomeric excesses, and the oxaziridine is correct. The ketene of every variant reaction
is left as the template \texttt{O=C=C(*)*}. Three of the six products (1\_1, 1\_2 and 2\_2) are exact; the other three
carry the inverted configuration at C2 (2\textit{S}). The template writes Ar and R as atoms, B17 is written as a
fragment of the catalyst, and B27 has a six-membered ring in place of the five-membered one and lacks its
tetrafluoroborate. No reaction matches the ground truth.}
\label{fig:out-cc}
\end{figure}
\setcounter{figure}{11}
\begin{figure}[p]
\centering
\includegraphics[width=\textwidth,height=0.78\textheight,keepaspectratio]{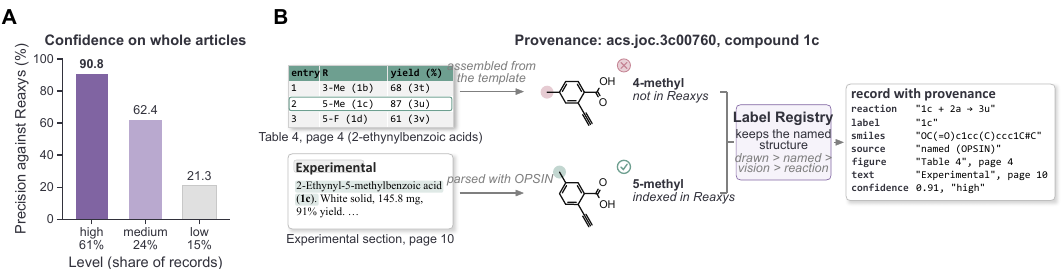}
\caption{\textbf{Confidence and provenance on whole articles.} (\textbf{A}) Share of records in each confidence level and their precision against Reaxys (19 articles, Gemini 3.7 Flash). (\textbf{B}) Provenance of compound 1c~\cite{zhang2023phthalides}: assembling the structure from the drawn template places the methyl group at C4, whereas the name in the experimental section, parsed with OPSIN, gives the 5-methyl isomer indexed in Reaxys; the Label Registry keeps the named structure, and the record stores its source. Precision against Reaxys is a lower bound, because Reaxys does not index every reaction in an article; in (B), the reaction and the score are illustrative.}
\end{figure}

\setcounter{figure}{12}
\begin{figure}[p]
\centering
\includegraphics[width=\linewidth,height=0.78\textheight,keepaspectratio]{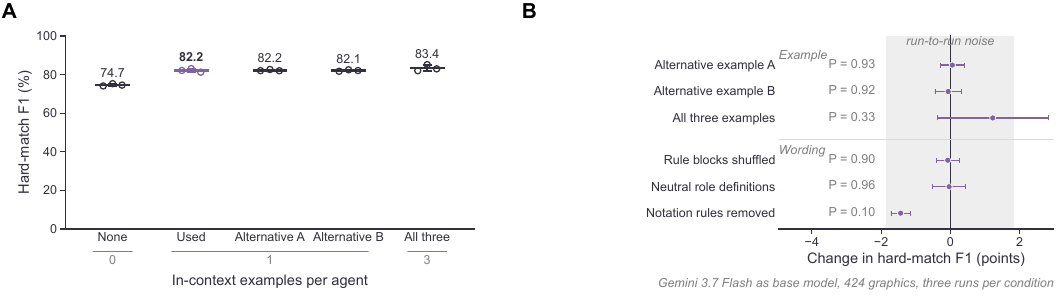}
\caption{\textbf{Robustness of ChemEAGLE to the in-context example and the prompt.} (\textbf{A}) Hard-match F1 without
the example, with the example used, with each of two alternative examples and with all
three examples; open circles show the three runs, and bars the mean and SD. (\textbf{B}) Change in hard-match F1 from
the default configuration (82.2 $\pm$ 0.9\%) for the alternative examples, all three examples and three changes
to the wording, mean and SD of three runs. The shaded band spans twice the run-to-run SD of the default.
\textit{P} values are from two-sided Welch's t-tests of each condition's three runs against those of the default.
Gemini 3.7 Flash as base model, 424 graphics.}
\label{fig:prompt}
\end{figure}
\setcounter{figure}{13}
\begin{figure}[p]
\centering
\includegraphics[width=\textwidth,height=0.78\textheight,keepaspectratio]{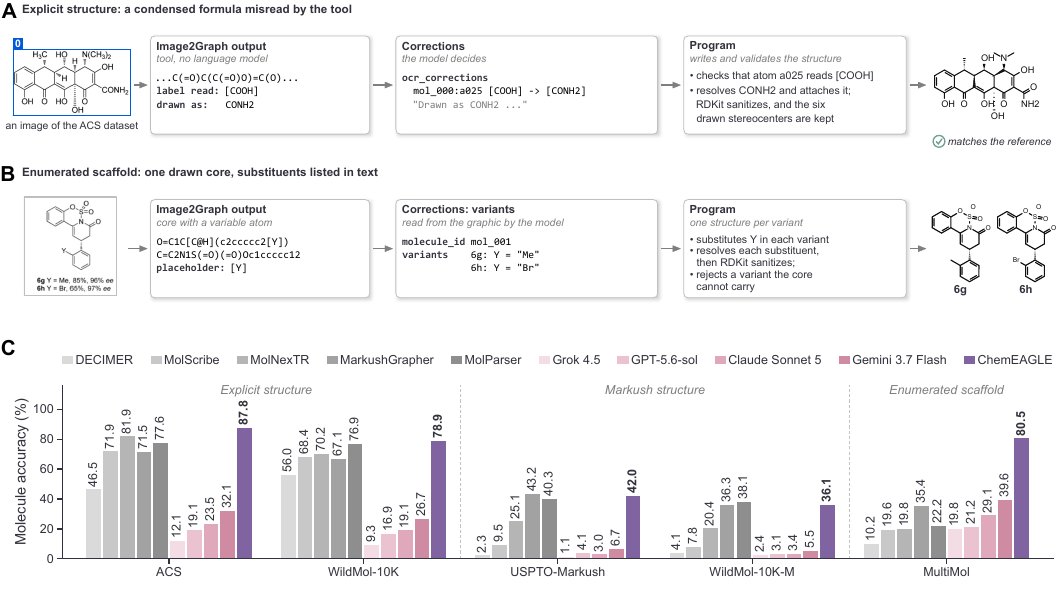}
\caption{\textbf{Molecular recognition with corrections by the model.} (\textbf{A}) A condensed formula misread by the tool, in an image of the ACS dataset: Image2Graph reads the carboxamide CONH$_2$ as COOH, a valid but wrong structure. The model relabels that atom CONH$_2$; the program checks the atom, resolves the label and sanitizes the result with RDKit, keeping the six stereocenters, and the structure matches the reference. (\textbf{B}) An enumerated scaffold, the graphic of Fig. 3G: a drawn core with a variable atom Y and substituents listed below it. Image2Graph returns the core with Y as a placeholder atom, the model lists the variants 6g (Y = Me) and 6h (Y = Br), and the program writes one structure per variant and rejects any variant that the core cannot carry. (\textbf{C}) Molecule accuracy on five datasets (ACS, WildMol-10K, USPTO-Markush, WildMol-10K-M and MultiMol, a set of 50 images built for this work) for specialist recognition models (DECIMER, MolScribe, MolNexTR, MarkushGrapher and MolParser), four MLLMs writing SMILES (Grok 4.5, GPT-5.6-sol, Claude Sonnet 5 and Gemini 3.7 Flash) and ChemEAGLE with Gemini 3.7 Flash (as in Fig. 3F); correct means an exact match of canonical SMILES.}
\end{figure}

\setcounter{figure}{14}
\begin{figure}[p]
\centering
\includegraphics[width=\linewidth,height=0.78\textheight,keepaspectratio]{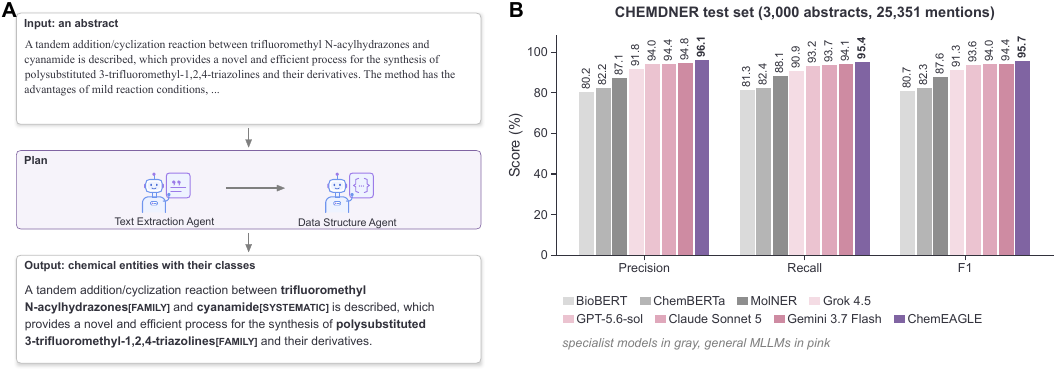}
\caption{\textbf{Chemical named entity recognition.} (\textbf{A}) An abstract, the plan (Text Extraction Agent, then
Data Structure Agent) and the output, with each entity in bold followed by its class. (\textbf{B}) Precision, recall
and F1 on the CHEMDNER test set (3,000 abstracts, 25,351 mentions) for specialist models (gray), the four MLLMs of
Fig.~S14 used alone (pink) and ChemEAGLE (purple; Gemini 3.7 Flash as base model).}
\label{fig:ner}
\end{figure}
\setcounter{figure}{15}
\begin{figure}[p]
\centering
\includegraphics[width=\linewidth,height=0.78\textheight,keepaspectratio]{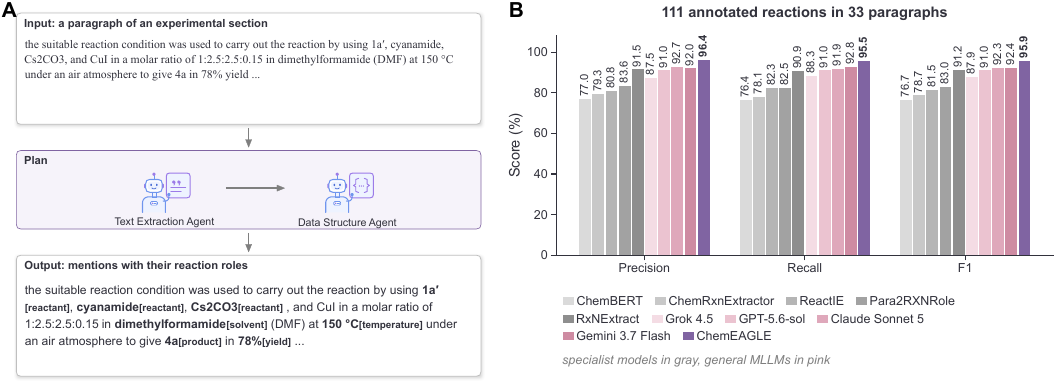}
\caption{\textbf{Text-based reaction extraction.} (\textbf{A}) A paragraph of an experimental section, the plan and the
output, with each mention in bold followed by its reaction role. (\textbf{B}) Precision, recall and F1 on 111 reactions
in 33 paragraphs for sequence-labeling and fine-tuned models (gray), the four MLLMs of Fig.~S14 used alone
(pink) and ChemEAGLE (purple; Gemini 3.7 Flash as base model).}
\label{fig:textrxn}
\end{figure}
\setcounter{figure}{16}
\begin{figure}[p]
\centering
\includegraphics[width=\textwidth,height=0.78\textheight,keepaspectratio]{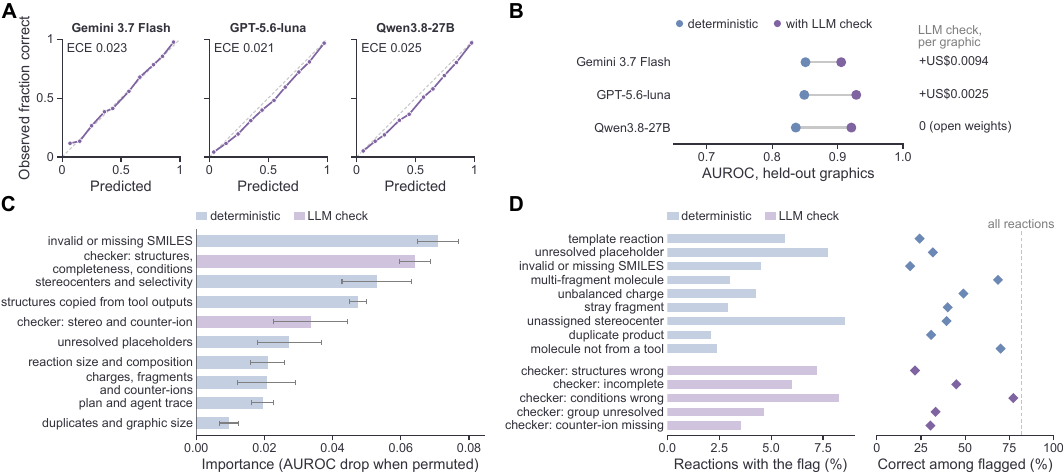}
\caption{\textbf{Details of the Confidence Scorer.} (\textbf{A}) Calibration for the three base models of Fig. 4C (Gemini 3.7 Flash, GPT-5.6-luna and Qwen3.8-27B): observed fraction correct against predicted probability in ten equal-width bins, with the expected calibration error (ECE); dashed line, perfect calibration. (\textbf{B}) Area under the receiver operating characteristic curve (AUROC) with deterministic features only and with the check by a large language model (LLM) added, and the cost of the check per graphic. (\textbf{C}) Importance of ten feature families: decrease in AUROC when all features of a family are permuted together within held-out graphics, so that correlated features do not divide their importance; mean over five folds and SD. Deterministic features and answers of the LLM checker are colored separately. (\textbf{D}) The 14 flags reported with the score: share of output reactions carrying each flag (left) and fraction correct among them (right); dashed line, fraction correct of all output reactions. Panels \textbf{C} and \textbf{D} use Gemini 3.7 Flash. Scores are out-of-fold predictions of fivefold cross-validation grouped by graphic, on the output reactions of three runs on all 424 graphics (10,959, 10,497 and 10,824 reactions).}
\end{figure}

\setcounter{figure}{17}
\begin{figure}[p]
\centering
\includegraphics[width=\linewidth,height=0.78\textheight,keepaspectratio]{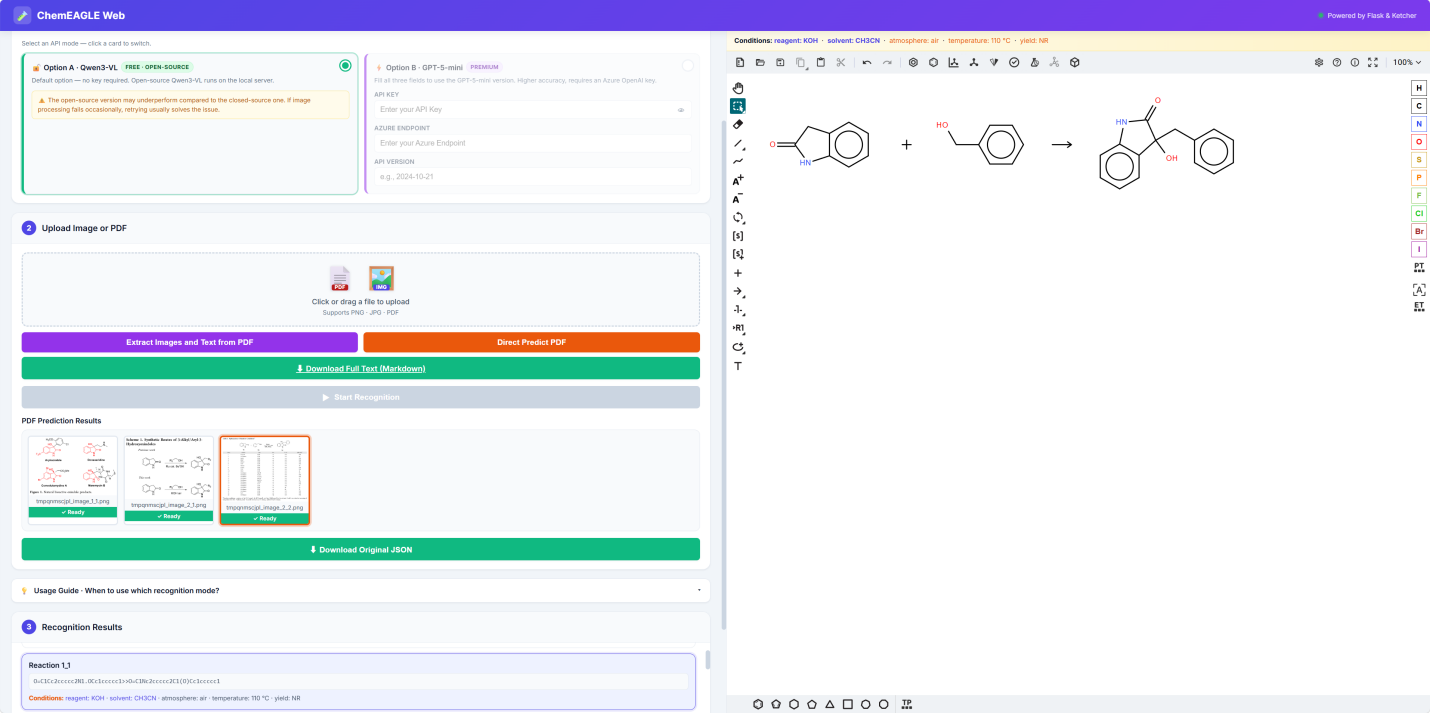}
\caption{\textbf{The ChemEAGLE web application during the extraction of a PDF.} Left, upload, settings, thumbnails of the
extracted graphics (the selected one in orange) and the extracted reactions. Top right, the Ketcher editor with the
selected reaction and its conditions. Bottom right, the PDF page with the detected regions next to the parsed text.}
\label{fig:webapp}
\end{figure}
\setcounter{figure}{18}
\begin{figure}[p]
\centering
\includegraphics[width=\textwidth,height=0.78\textheight,keepaspectratio]{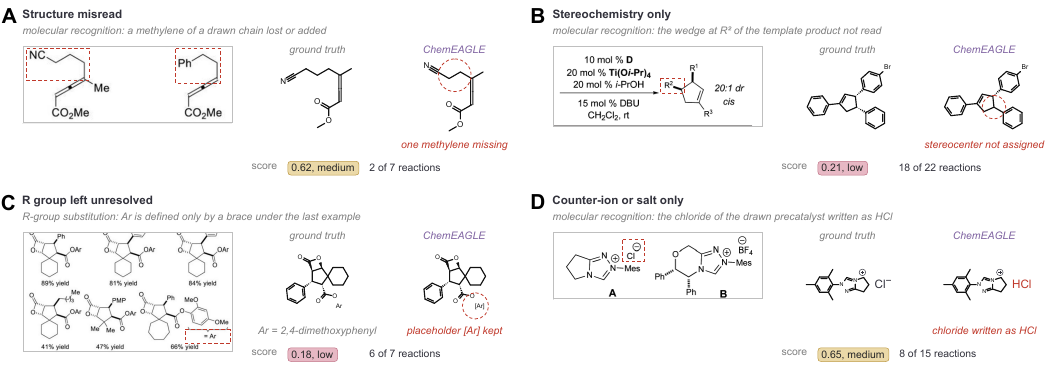}
\caption{\textbf{Typical failures of ChemEAGLE.} One example for each of four error types, from the Gemini 3.7 Flash run of Fig. 4E: part of the graphic with the source of the error boxed, the ground-truth structure, ChemEAGLE's structure with the error circled, the score and level of the reaction shown, and the number of reactions of the graphic affected. (\textbf{A}) Structure misread: one methylene of a drawn chain is lost (shown), and one is added in a second product; both leave valid molecules; scored medium. (\textbf{B}) Stereochemistry only: the wedge at R$^2$ of the template product is not read, so 18 of the 22 products lack this stereocenter; scored low. (\textbf{C}) R group left unresolved: Ar is defined only by a brace under the last example and stays a placeholder in 6 of the 7 reactions; scored low. (\textbf{D}) Counter-ion or salt only: the chloride of precatalyst A is written as HCl, leaving the salt with a net charge, in 8 of the 15 reactions; scored medium. Scores are out-of-fold predictions of cross-validation grouped by graphic (Materials and Methods); levels are high ($\geq$ 0.85), medium (0.5--0.85) and low (\textless{} 0.5).}
\end{figure}

\setcounter{figure}{19}
\begin{figure}[p]
\centering
\includegraphics[width=\linewidth,height=0.78\textheight,keepaspectratio]{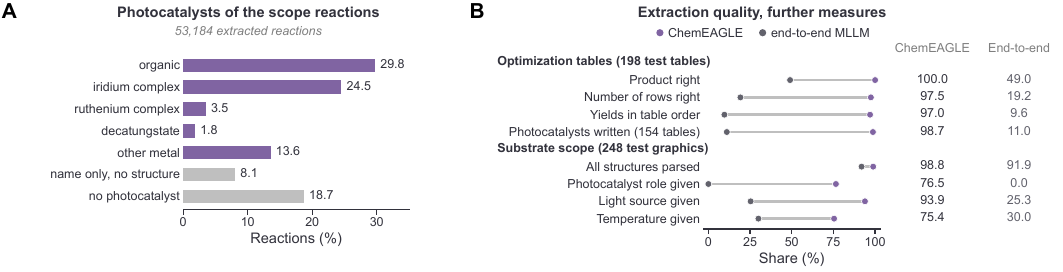}
\caption{\textbf{Photocatalysts of the substrate-scope reactions and further measures of extraction quality.}
(\textbf{A}) Photocatalyst class of the 53,184 substrate-scope reactions extracted by ChemEAGLE, assigned from the
extracted structure (iridium, ruthenium, tungsten or another metal, otherwise organic); name only, a photocatalyst
named without a structure. (\textbf{B}) ChemEAGLE against the end-to-end extraction by the same base model (Gemini 3.7 Flash). Optimization
tables, per table: the most common extracted product equals the product of the hand-checked record; as many rows as the
record; the yields in table order equal the record's and the product is right; every photocatalyst structure of the
record occurs in the extraction, in any role (the 154 tables that use a photocatalyst). Substrate scope, per reaction:
every structure can be parsed; a chemical in the photocatalyst role; a light source or wavelength; a temperature. ChemEAGLE's substrate-scope values are those of its hand-checked records.}
\label{fig:photodata}
\end{figure}
\setcounter{figure}{20}
\begin{figure}[p]
\centering
\includegraphics[width=\linewidth,height=0.78\textheight,keepaspectratio]{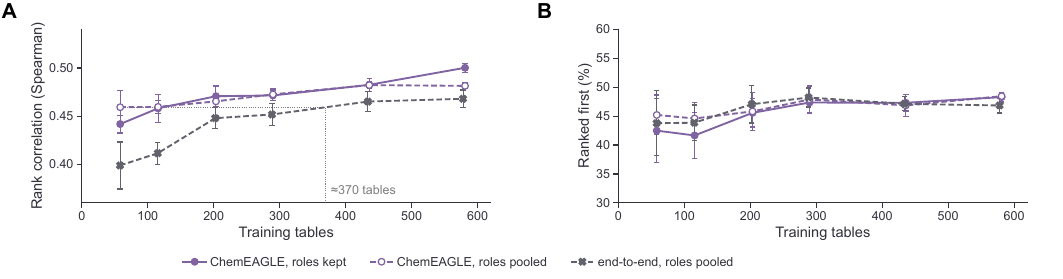}
\caption{\textbf{Learning curves of condition ranking.} The tuned LambdaRank trained on random fractions of the
training tables (up to 2024) and evaluated on the 198 test tables (2025 and 2026). (\textbf{A}) Spearman correlation
between predicted and reported yields within a table, averaged over tables. The dotted line marks the number of
end-to-end tables needed to reach the correlation of 58 ChemEAGLE tables, both with roles pooled (linear interpolation).
(\textbf{B}) Share of tables with the best condition ranked first. Mean and SD of five random subsets per fraction; the
same graphics are drawn for both sources.}
\label{fig:photolearn}
\end{figure}
\setcounter{figure}{21}
\begin{figure}[p]
\centering
\includegraphics[width=\linewidth,height=0.78\textheight,keepaspectratio]{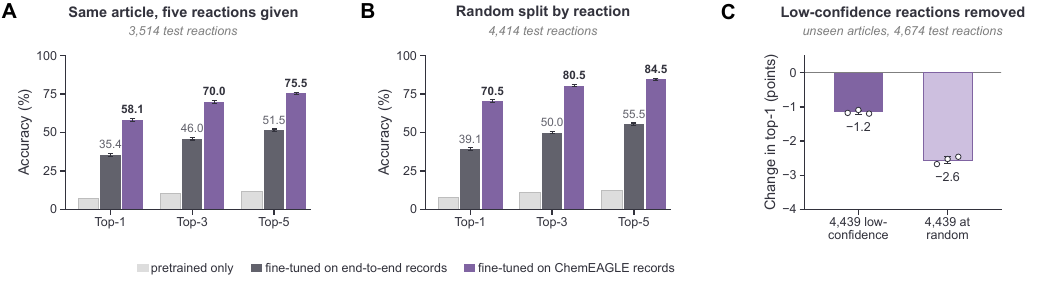}
\caption{\textbf{Product prediction in two other settings and without low-confidence reactions.} ReactionT5v2-forward,
pretrained only or fine-tuned on the end-to-end or ChemEAGLE records of the training graphics. (\textbf{A}) Five
reactions of each test graphic added to training; 3,514 test reactions from 209 graphics. (\textbf{B}) Random split by
reaction; 4,414 test reactions. Top-$k$, the correct product among the first $k$ predictions. Bars show the mean of
three runs and error bars the SD; the pretrained model is deterministic. (\textbf{C}) Change in top-1 accuracy on the
4,674 test reactions of unseen articles when the 4,439 low-confidence reactions, or as many random reactions, are
removed from training; bars show the mean and error bars the SD of the paired differences over three runs, and
open circles the single runs.}
\label{fig:photoforward}
\end{figure}
\clearpage
\clearpage
\setcounter{table}{0}
\begingroup
\footnotesize
\renewcommand{\arraystretch}{1.15}
\setlength{\LTcapwidth}{\linewidth}
\begin{xltabular}{\linewidth}{@{}>{\raggedright\arraybackslash}p{2.6cm}>{\raggedright\arraybackslash\hsize=0.8\hsize}X>{\raggedright\arraybackslash\hsize=0.95\hsize}X>{\raggedright\arraybackslash\hsize=1.25\hsize}X@{}}
\caption{\textbf{What the tool calls and the model do in each agent and in the document layer.} All agents use
the same base model. For components that write no structures, the last column gives the program steps applied to the
model's answer.}
\label{tab:division} \\
\toprule
\textbf{Component} & \textbf{Tool calls: read the graphic} & \textbf{Model returns} & \textbf{Tool calls: write the
structures (or programs)} \\
\midrule
\endfirsthead
\multicolumn{4}{@{}l}{\small\textbf{Table~S1} (continued)} \\[3pt]
\toprule
\textbf{Component} & \textbf{Tool calls: read the graphic} & \textbf{Model returns} & \textbf{Tool calls: write the
structures (or programs)} \\
\midrule
\endhead
\endfoot
\bottomrule
\endlastfoot
\multicolumn{4}{@{}l}{\textit{Agents (Fig.~1B)}} \\
Planner & none & ordered list of the agents to run & parse the list; run an R-group agent that the plan omitted when a recognized molecule carries a variable atom (structure-based if at least five fully specified molecules are drawn, text-based otherwise) \\
Plan Observer & none & revised list of agents with a reason & replace the plan before any agent runs \\
Reaction Template Parsing Agent & RxnImgParser: molecules, condition regions and their roles, multistep schemes & edit plan by identifier: corrections of label atoms, charges and look-alike elements, counter-ions, explicit definitions and a decision for each molecule with a variable & apply the plan to every appearance of each molecule; Graph2SMILES; pass condition regions on \\
Molecular Recognition Agent & MolDetect: molecules and their labels; Image2Graph: atom symbols, coordinates and bonds; Graph2SMILES; OCR & corrections of atom symbols, charges and text; definitions of abbreviations with the molecules they apply to; groups that expand a core into variants; counter-ions; a decision per molecule & validate every entry against the catalog (existing atom, current symbol, placeholder present, valence, rings, RDKit sanitization); one retry with the validator's message; apply the entries that pass and write the structures with Graph2SMILES; remove stray atoms; repair ring bond orders; register labels \\
Structure-based R-group Substitution Agent & molecules recognized by the Molecular Recognition Agent, with identifiers & per identifier: printed label, adjacent text and role & SMILESReconstructor: copy each structure by identifier; match each product variant to the template, cut out the variable fragments and build the reactant variant \\
Text-based R-group Substitution Agent & catalog of template label atoms; condition texts; molecules and drawn fragments recognized by the Molecular Recognition Agent & corrections of misread label atoms; one entry per table row with the printed value of every variable, the row's conditions and yield (never SMILES) & TableParser: resolve each value to a fragment, attach it to the template and build one reaction per row \\
Condition Interpretation Agent & RxnConInterpreter: condition text and roles; TesseractOCR & corrected roles and readings; links between drawn catalysts or reagents and their structures by label & convert names that are not drawn by the name-resolution chain \\
Text Extraction Agent & TesseractOCR; MolNER; ChemRxnExtractor & annotated passage that aligns chemical mentions with the structures found by the other agents & none \\
Action Observer & none & issues per agent (location, category, severity); optionally one re-execution & discard reports that would change a molecule's main body; forward the rest as notes; rerun the agent once when requested \\
Data Structure Agent & none & reaction records in a fixed schema, structures copied from the agents' outputs & validate and repair SMILES; resolve remaining names; propagate drawn condition structures; repair charges and radicals \\
Confidence Scorer & none & optional check: yes, no or unsure answers to fixed questions, a problem description, the expected number of reactions & compute deterministic features, including whether each reactant and product was copied from an agent's tool calls; gradient-boosted classifier with isotonic calibration; report probability, level and flags \\
\midrule
\multicolumn{4}{@{}l}{\textit{Document level (Fig.~1A)}} \\
Parsing layer & pdfplumber: text layer in reading order; VisualHeist: figure and table regions & none & register each figure and table with its name, caption, footnotes, page and box; split the text into paragraphs by section, each with the figures it cites and the compound labels it names \\
Main Text Agent & text layer and figure registry of the parsing layer & the role of each figure, from its caption and citing sentences (from its image when it has no text); per figure, the conditions, yields and selectivities stated in the text; per section of text, the reactions it states were run; each item with a verbatim quote & check every quote against the cited text: unverified items are dropped and unverified roles give way to caption rules; skip spectra and data plots; mark reactions of a mechanism as proposed steps; convert the systematic names of characterization paragraphs with OPSIN; take the structures of text reactions from the Label Registry or OPSIN, never from the model; attach a text reaction whose product matches a figure reaction to it \\
Label Registry & structures drawn beside spectra, read by the recognition tools & none & one structure per compound label, in order of trust: drawn beside the label, named (OPSIN), drawn beside a spectrum, from a labeled molecule of an extracted reaction; within an origin, the structure seen in most figures; a label that names several compounds (two structures in one figure, or three or more in the article) is not used \\
Document assembly & none & none & give molecules that carry only a label the registered structure and replace structures built from a template row; keep a structure the figure draws itself and record a different registered one as a conflict; attach the text evidence; record the figure, page and paragraph of every reaction \\
\end{xltabular}
\endgroup
\setcounter{table}{1}
\begin{table}[htbp]
\centering
\caption{\textbf{Molecule detection on the extended test set.} Average precision on 350 test images.}
\label{tab:moldetect}
\small
\begin{tabular}{@{}lc@{}}
\toprule
Model & Average precision (\%) \\
\midrule
ChemSchematicResolver~\cite{beard2020chemschematicresolver} & 28.0 \\
MolMiner~\cite{xu2022molminer} & 80.9 \\
MolDetect~\cite{fan2024openchemie} & 84.8 \\
MolDetect, retrained (this work) & \textbf{86.5} \\
\bottomrule
\end{tabular}
\end{table}
\setcounter{table}{2}
\begin{table}[htbp]
\centering
\caption{\textbf{Audit of the observers.} First run of the full system on all 424 graphics (Gemini 3.7 Flash).
Interventions are plan changes (Plan Observer) or checks that flagged an issue (Action Observer; out of all checks run).
Justified and unnecessary or false are shares of the interventions; fixed is the share of justified interventions after
which the affected part was correct in the final output; missed is the share of graphics with a plan or agent error that
the observer did not flag. Counts in parentheses.}
\label{tab:observers}
\footnotesize
\setlength{\tabcolsep}{4pt}
\begin{tabular}{@{}llcccccc@{}}
\toprule
Observer & Graphics & Interventions & \makecell{Per\\ graphic} & \makecell{Justified\\ (\%)} & \makecell{Unnecessary\\ or false (\%)} & \makecell{Fixed\\ (\%)} & \makecell{Missed\\ (\%)} \\
\midrule
Plan & All (424) & 89 & 0.21 & 87.6 (78) & 12.4 (11) & 84.6 (66) & 8.3 (35) \\
 & Articles (324) & 70 & 0.22 & 87.1 (61) & 12.9 (9) & 85.2 (52) & 7.4 (24) \\
 & Patents (100) & 19 & 0.19 & 89.5 (17) & 10.5 (2) & 82.4 (14) & 11.0 (11) \\
\midrule
Action & All (424) & 354 / 1,158 & 0.83 & 90.4 (320) & 9.6 (34) & 69.7 (223) & 12.0 (51) \\
 & Articles (324) & 270 / 882 & 0.83 & 90.0 (243) & 10.0 (27) & 68.3 (166) & 12.0 (39) \\
 & Patents (100) & 84 / 276 & 0.84 & 91.7 (77) & 8.3 (7) & 74.0 (57) & 12.0 (12) \\
\bottomrule
\end{tabular}
\end{table}
\setcounter{table}{3}
\begin{table}[htbp]
\centering
\caption{\textbf{Language-model fallback for name resolution on the hardest names.} Validity and
correctness over the 41 names that remain after excluding 16 with defective ground truth or unanswerable content; outputs validated with RDKit
and compared with the ground truth by InChIKey. Invalid outputs are rejected by the pipeline before acceptance.}
\label{tab:fallback}
\small
\begin{tabular}{@{}lcc@{}}
\toprule
Base model & Chemical validity (\%) & Semantic correctness (\%) \\
\midrule
Gemini 3.7 Flash & 92.7 & 75.6 \\
GPT-5.6-sol & 95.1 & 78.0 \\
GPT-5.6-luna & 92.7 & 70.7 \\
Claude Sonnet 5 & 95.1 & 80.5 \\
Grok 4.5 & 95.1 & 75.6 \\
Qwen3.8-Max & 95.1 & 73.2 \\
Qwen3.8-27B & 90.2 & 70.7 \\
\bottomrule
\end{tabular}

\end{table}
\setcounter{table}{4}
\begin{table}[htbp]
\centering
\caption{\textbf{Annotators.} Position, field and approximate experience with the annotation of reaction schemes.}
\label{tab:annotators}
\small
\begin{tabular}{@{}llll@{}}
\toprule
Annotator & Position & Field & Annotation experience \\
\midrule
R1 & Second-year PhD student & Chemical and biomolecular engineering & about 2 years \\
R2 & Second-year PhD student & Chemical and biomolecular engineering & about 1 year \\
R3 & Fourth-year undergraduate & Chemical and biomolecular engineering & less than 1 year \\
R4 & Fourth-year PhD candidate & Chemical and biomolecular engineering & about 2 years \\
\bottomrule
\end{tabular}
\end{table}
\setcounter{table}{5}
\begin{table}[htbp]
\centering
\caption{\textbf{Inter-annotator agreement.} Computed over all 3,613 benchmark reactions from four complete,
independent annotation sets.}
\label{tab:agreement}
\small
\begin{tabular}{@{}lc@{}}
\toprule
Metric & Value \\
\midrule
Pairwise F1, R1 and R2 (\%) & 98.8 \\
\quad R1 and R3 (\%) & 96.8 \\
\quad R1 and R4 (\%) & 98.5 \\
\quad R2 and R3 (\%) & 97.0 \\
\quad R2 and R4 (\%) & 98.2 \\
\quad R3 and R4 (\%) & 96.5 \\
Mean of the six pairs $\pm$ SD (\%) & 97.6 $\pm$ 1.0 \\
Reactions with identical records from all four annotators (\%) & 94.5 \\
Gwet's AC1 & 0.968 \\
Reactions entering adjudication (\%) & 5.5 \\
Records changed by adjudication, \textit{n} (\%) & 22 (0.6) \\
\bottomrule
\end{tabular}
\end{table}
\setcounter{table}{6}
\begin{table}[htbp]
\centering
\caption{\textbf{Results of every system by subset and over all 424 graphics.} Hard-match F1 on each subset (139 research-article, 78 OpenChemIE-set, 107 review and 100 patent graphics), and precision (P), recall (R), F1 and soft-match F1 over all 424 graphics (3,613 reactions). Values are means $\pm$ SD of three runs; for the overall columns, run k on the article graphics was pooled with run k on the patent graphics before scoring. Under the hard criterion, every reactant and product structure and every condition structure drawn in the graphic must match the ground truth; the soft criterion does not require the drawn condition structures, and under both a molecule filed as a reactant on one side and as a condition on the other is not an error (Materials and Methods). ChemEAGLE uses Gemini 3.7 Flash; Codex runs on GPT-5.6-sol and Claude Code on Claude Sonnet 5.}
\footnotesize
\setlength{\tabcolsep}{3pt}
\resizebox{\textwidth}{!}{
\begin{tabular}{@{}lcccccccc@{}}
\toprule
& \multicolumn{4}{c}{Hard-match F1 by subset (\%)} & \multicolumn{4}{c}{All 424 graphics (\%)} \\
\cmidrule(lr){2-5}\cmidrule(l){6-9}
System & Research & OpenChemIE set & Review & Patents & P & R & F1 & Soft F1 \\
\midrule
\multicolumn{9}{@{}l}{\textit{Rule-based pipeline}} \\
OpenChemIE & 27.1~$\pm$~0.1 & 63.7~$\pm$~0.0 & 1.8~$\pm$~0.2 & 12.8~$\pm$~4.2 & 55.0~$\pm$~2.6 & 24.6~$\pm$~0.4 & 34.0~$\pm$~0.9 & 39.7~$\pm$~1.0 \\
\multicolumn{9}{@{}l}{\textit{Single MLLM with tools}} \\
MERMaid & 7.4~$\pm$~0.1 & 10.0~$\pm$~0.2 & 1.8~$\pm$~0.2 & 29.4~$\pm$~6.4 & 33.3~$\pm$~4.3 & 8.2~$\pm$~1.1 & 13.1~$\pm$~1.8 & 19.0~$\pm$~1.9 \\
\multicolumn{9}{@{}l}{\textit{End-to-end MLLM}} \\
Qwen3.8-27B & 1.7~$\pm$~0.0 & 0.1~$\pm$~0.1 & 0.2~$\pm$~0.2 & 0.8~$\pm$~0.2 & 0.9~$\pm$~0.1 & 0.9~$\pm$~0.1 & 0.9~$\pm$~0.1 & 6.7~$\pm$~0.2 \\
GPT-5.6-luna & 1.9~$\pm$~0.3 & 0.1~$\pm$~0.1 & 0.1~$\pm$~0.2 & 0.9~$\pm$~0.7 & 0.9~$\pm$~0.2 & 0.9~$\pm$~0.2 & 0.9~$\pm$~0.2 & 6.6~$\pm$~0.2 \\
Grok 4.5 & 14.9~$\pm$~1.0 & 10.7~$\pm$~0.4 & 12.5~$\pm$~0.6 & 11.6~$\pm$~0.2 & 12.7~$\pm$~0.5 & 12.9~$\pm$~0.5 & 12.8~$\pm$~0.5 & 27.2~$\pm$~0.9 \\
GPT-5.6-sol & 15.6~$\pm$~1.0 & 13.1~$\pm$~1.1 & 12.0~$\pm$~2.2 & 12.0~$\pm$~0.1 & 13.6~$\pm$~0.5 & 13.9~$\pm$~0.6 & 13.7~$\pm$~0.5 & 28.7~$\pm$~0.4 \\
Qwen3.8-Max & 18.3~$\pm$~1.1 & 20.7~$\pm$~0.2 & 21.5~$\pm$~2.4 & 17.4~$\pm$~2.0 & 19.2~$\pm$~0.9 & 19.4~$\pm$~1.0 & 19.3~$\pm$~0.9 & 30.5~$\pm$~0.7 \\
Claude Sonnet 5 & 19.9~$\pm$~2.8 & 35.1~$\pm$~0.4 & 9.3~$\pm$~1.8 & 18.0~$\pm$~2.6 & 22.6~$\pm$~1.5 & 22.7~$\pm$~1.6 & 22.7~$\pm$~1.5 & 31.5~$\pm$~0.8 \\
Gemini 3.7 Flash & 17.1~$\pm$~0.8 & 51.3~$\pm$~0.7 & 11.4~$\pm$~0.5 & 23.7~$\pm$~2.5 & 27.7~$\pm$~0.9 & 27.9~$\pm$~0.7 & 27.8~$\pm$~0.8 & 35.5~$\pm$~1.5 \\
\multicolumn{9}{@{}l}{\textit{Generalist agent}} \\
Codex (GPT-5.6-sol) & 34.6~$\pm$~1.3 & 47.5~$\pm$~0.4 & 26.7~$\pm$~4.1 & 32.5~$\pm$~2.6 & 45.4~$\pm$~1.0 & 31.3~$\pm$~1.0 & 37.1~$\pm$~1.0 & 48.1~$\pm$~0.7 \\
Claude Code (Claude Sonnet 5) & 41.9~$\pm$~2.5 & 61.9~$\pm$~1.5 & 39.3~$\pm$~1.6 & 43.0~$\pm$~2.9 & 57.1~$\pm$~2.1 & 41.0~$\pm$~1.5 & 47.7~$\pm$~1.7 & 61.4~$\pm$~0.4 \\
\midrule
\textbf{ChemEAGLE (Gemini 3.7 Flash)} & \textbf{75.5~$\pm$~0.8} & \textbf{93.2~$\pm$~0.2} & \textbf{79.5~$\pm$~3.1} & \textbf{80.0~$\pm$~1.5} & \textbf{81.7~$\pm$~1.3} & \textbf{82.6~$\pm$~0.9} & \textbf{82.2~$\pm$~0.9} & \textbf{85.5~$\pm$~0.8} \\
\bottomrule
\end{tabular}}
\end{table}

\setcounter{table}{7}
\begin{table}[htbp]
\centering
\caption{\textbf{Every base model alone, driving a generalist agent and inside ChemEAGLE.} For each base model: hard-match F1 when used end to end (alone) and when driving a generalist agent given ChemEAGLE's toolset (Codex on GPT-5.6-sol, Claude Code on Claude Sonnet 5); precision (P), recall (R), F1 and soft-match F1 of ChemEAGLE built on it; hard-match F1 without the visual prompt for the three base models of the ablation (Fig. 3C); and the cost of one ChemEAGLE run per graphic in US dollars, with the Qwen models counted as free. Values are means $\pm$ SD of three runs on all 424 graphics; dashes mark configurations that were not run.}
\footnotesize
\setlength{\tabcolsep}{3pt}
\resizebox{\textwidth}{!}{
\begin{tabular}{@{}lcccccccc@{}}
\toprule
& Alone & Generalist agent & \multicolumn{5}{c}{Inside ChemEAGLE (\%)} & Cost \\
\cmidrule(lr){4-8}
Base model & F1 (\%) & F1 (\%) & P & R & F1 & Soft F1 & F1, no visual prompt & (US\$ per graphic) \\
\midrule
GPT-5.6-sol & 13.7~$\pm$~0.5 & 37.1~$\pm$~1.0 (Codex) & 78.3~$\pm$~0.6 & 77.3~$\pm$~0.9 & 77.8~$\pm$~0.7 & 81.1~$\pm$~0.7 & -- & 0.60 \\
Claude Sonnet 5 & 22.7~$\pm$~1.5 & 47.7~$\pm$~1.7 (Claude Code) & 80.4~$\pm$~0.7 & 79.9~$\pm$~0.4 & 80.2~$\pm$~0.5 & 82.3~$\pm$~0.0 & -- & 0.30 \\
Grok 4.5 & 12.8~$\pm$~0.5 & -- & 77.8~$\pm$~0.8 & 76.4~$\pm$~0.8 & 77.1~$\pm$~0.8 & 80.0~$\pm$~0.7 & -- & 0.24 \\
Gemini 3.7 Flash & 27.8~$\pm$~0.8 & -- & 81.7~$\pm$~1.3 & 82.6~$\pm$~0.9 & 82.2~$\pm$~0.9 & 85.5~$\pm$~0.8 & 79.0~$\pm$~0.4 & 0.11 \\
GPT-5.6-luna & 0.9~$\pm$~0.2 & -- & 75.4~$\pm$~1.6 & 73.0~$\pm$~0.9 & 74.2~$\pm$~1.2 & 76.2~$\pm$~1.2 & 67.3~$\pm$~0.5 & 0.03 \\
Qwen3.8-Max & 19.3~$\pm$~0.9 & -- & 79.4~$\pm$~0.9 & 78.6~$\pm$~0.4 & 79.0~$\pm$~0.7 & 81.9~$\pm$~0.2 & -- & 0* \\
Qwen3.8-27B & 0.9~$\pm$~0.1 & -- & 75.6~$\pm$~0.5 & 75.5~$\pm$~0.2 & 75.5~$\pm$~0.3 & 78.1~$\pm$~1.1 & 67.3~$\pm$~0.5 & 0* \\
\bottomrule
\end{tabular}}
\par\smallskip{\raggedright *Open-weight model, run on our own hardware.\par}
\end{table}

\setcounter{table}{8}
\begin{table}[htbp]
\centering
\caption{\textbf{Copied and written molecules in the output of the generalist agents.} Claude Code with Claude
Sonnet 5 and Codex with GPT-5.6-sol on all 424 benchmark graphics, one run each. Copied, identical to a structure returned by
the agent's own tool calls; written, all other molecules. Correct, identical to a ground-truth molecule of the same
graphic (stereocenters compared). Errors written, share of the incorrect molecules that the agent wrote.}
\label{tab:agentwritten}
\small
\begin{tabular}{@{}lcccc@{}}
\toprule
Molecules & Copied (\%) & Copied, correct (\%) & Written, correct (\%) & Errors written (\%) \\
\midrule
\multicolumn{5}{@{}l}{\textit{Claude Code (Claude Sonnet 5)}} \\
All & 58.5 & 91.9 & 88.3 & 50.6 \\
Products & 58.5 & 95.8 & 85.3 & 71.2 \\
Reactants & 46.3 & 97.7 & 93.4 & 76.9 \\
Condition structures & 76.1 & 84.8 & 75.4 & 33.7 \\
\midrule
\multicolumn{5}{@{}l}{\textit{Codex (GPT-5.6-sol)}} \\
All & 52.8 & 90.0 & 86.2 & 55.2 \\
Products & 51.6 & 93.8 & 82.3 & 72.7 \\
Reactants & 39.7 & 96.7 & 91.5 & 79.5 \\
Condition structures & 72.3 & 83.4 & 73.3 & 38.1 \\
\bottomrule
\end{tabular}
\end{table}
\setcounter{table}{9}
\begin{table}[htbp]
\centering
\caption{\textbf{Time, accuracy and cost of the extraction workflows.} All 424 graphics. Time is the mean per graphic;
medians (ranges) were 0.520 (0.25--1.80) h for manual annotation, 0.150 (0.08--0.45) h for reviewing every output and
0.022 (0.012--0.040) h for automated extraction. Accuracy is the hard-match accuracy of the final records, mean
$\pm$ SD over the four annotators (over three runs for automated extraction). Cost at US\$30 per hour plus US\$0.1106
per graphic for automated extraction. R1--R4, time (h per graphic) / accuracy (\%) of each annotator. The last row gives ChemEAGLE with Gemini 3.7
Flash without review, whose accuracy is its hard-match precision (Fig.~2A).}
\label{tab:workflow}
\footnotesize
\setlength{\tabcolsep}{4pt}
\begin{tabular}{@{}lccccccc@{}}
\toprule
Workflow & Time (h) & Accuracy (\%) & \makecell{Cost (US\$\\ per graphic)} & R1 & R2 & R3 & R4 \\
\midrule
\makecell[l]{Manual\\ annotation} & 0.572 & 98.42 $\pm$ 0.88 & 17.16 & 0.550 / 98.6 & 0.518 / 99.5 & 0.655 / 97.4 & 0.565 / 98.2 \\
\makecell[l]{Review every\\ output} & 0.175 & 99.35 $\pm$ 0.19 & 5.36 & 0.165 / 99.5 & 0.150 / 99.5 & 0.215 / 99.1 & 0.170 / 99.3 \\
\makecell[l]{Review medium\\ and low only} & 0.068 & 96.55 $\pm$ 0.35 & 2.16 & 0.066 / 96.5 & 0.065 / 97.0 & 0.074 / 96.1 & 0.068 / 96.6 \\
\makecell[l]{ChemEAGLE\\ alone} & 0.025 & 81.7 $\pm$ 1.3 & 0.11 & & & & \\
\bottomrule
\end{tabular}
\end{table}
\setcounter{table}{10}
\begin{table}[htbp]
\centering
\caption{\textbf{Split of the time spent reviewing every output.} Verification and correction were logged separately
through the web interface. Graphics that needed stereochemistry corrections (about 22\% of the benchmark) took longer
to correct.}
\label{tab:timesplit}
\small
\begin{tabular}{@{}lcc@{}}
\toprule
Component (h per graphic) & Without stereochemistry corrections & With stereochemistry corrections \\
\midrule
Automated run & 0.025 & 0.025 \\
Verification & 0.065 & 0.095 \\
Correction & 0.060 & 0.145 \\
Total & 0.150 & 0.265 \\
\bottomrule
\end{tabular}
\end{table}
\setcounter{table}{11}
\begin{table}[htbp]
\centering
\caption{\textbf{Errors added to the training data.} Names as in Fig.~5F (condition ranking) and Fig.~5I (product
prediction). Condition ranking: 10, 25, 50, 75 and 100\% of the training tables, three runs; product prediction: 25,
50, 75 and 100\% of the training scopes, three runs. The end-to-end rows replace ChemEAGLE's records of a graphic with
the end-to-end extraction of the same graphic; in condition ranking, species are pooled regardless of role for both
sources, so that only the content differs.}
\label{tab:photoerrors}
\footnotesize
\begin{tabularx}{\linewidth}{@{}>{\raggedright\arraybackslash}p{3.2cm}>{\raggedright\arraybackslash}X>{\raggedright\arraybackslash}X@{}}
\toprule
Error (Fig.~5F; Fig.~5I) & Optimization tables (Fig.~5F) & Substrate scopes (Fig.~5I) \\
\midrule
structures & Reactants and products replaced by the end-to-end structures of the same graphic where they differ,
otherwise one heavy atom of the product changed & One heavy atom of each product changed (C to N, N to C, O to S, S to
O, others to C) \\
row shift; products on wrong entry & Yields moved by one row within a block of at least two rows & The products of
about a third of the entries moved by one entry \\
merged rows; substituents not applied & About a third of the rows given the conditions of the first row, their yields
kept & About a third of the entries given the reactants of the first entry, their products kept \\
roles & The chemicals of the table pooled into one list, without photocatalyst, catalyst and ligand roles &
Photocatalysts, catalysts, ligands and reagents written as reactants \\
end-to-end & The rows of a graphic replaced by the end-to-end extraction & The reactions of a graphic replaced by the
end-to-end extraction \\
\bottomrule
\end{tabularx}
\end{table}
\clearpage
\paragraph{Caption for Data S1.}
\textbf{Values plotted in the figures.} One worksheet per figure (main and supplementary), with one table per panel giving every value drawn (bar heights, error bars, points of curves and their single runs) and, where tests were made, the P values.

\end{document}